\documentclass[pdflatex,sn-mathphys]{sn-jnl}% Math and Physical Sciences Reference Style
\usepackage{algpseudocode}
\usepackage{latexsym}
\usepackage{graphicx}
\usepackage{mathptmx}
\usepackage{amsmath}
\usepackage{amsfonts}
\usepackage{amssymb}
\usepackage{amsbsy}
\usepackage{amsthm}
\usepackage{subfig}

\usepackage{multicol}
\usepackage{wrapfig}
\usepackage{lipsum}
\usepackage{caption}
\usepackage{multirow}
\usepackage{color}
\usepackage{booktabs}
\usepackage{adjustbox}
\usepackage{flushend}
\usepackage[a-1b]{pdfx}   % for PDF/A-1b
\usepackage{balance}

\algtext*{EndWhile}
\algtext*{EndIf}
\algtext*{EndFunction}
\algtext*{EndFor}
\algrenewcommand\algorithmicrequire{\textbf{Input:}}
\algrenewcommand\algorithmicensure{\textbf{Output:}}

\newcommand{\gee}{\mathcal{G}}
\newcommand{\dee}{\mathcal{D}}
\newcommand{\tee}{\mathcal{T}}
\newcommand{\ef}{\mathcal{F}}

\usepackage{newfloat}
\usepackage{listings}
\floatstyle{ruled}
\newfloat{listing}{tb}{lst}{}
\floatname{listing}{Listing}
\usepackage{bibentry}
 
\newcommand{\stitle}[1]{\noindent{\bf #1}}

\newcommand{\neww}[1]{\textcolor{black}{#1}}

\newcommand{\indep}{\perp\!}
\newcommand{\rep}{R_\mathcal{F}}

\jyear{2021}%

\theoremstyle{thmstyleone}%
\newtheorem{theorem}{Theorem}%  meant for continuous numbers
\theoremstyle{thmstyletwo}%

\theoremstyle{thmstylethree}%

\newtheorem{lemma}[theorem]{Lemma}

\title[]{Finding Representative Group Fairness Metrics Using Correlation Estimations}

\begin{document}
%%=============================================================%%
%% Prefix	-> \pfx{Dr}
%% GivenName	-> \fnm{Joergen W.}
%% Particle	-> \spfx{van der} -> surname prefix
%% FamilyName	-> \sur{Ploeg}
%% Suffix	-> \sfx{IV}
%% NatureName	-> \tanm{Poet Laureate} -> Title after name
%% Degrees	-> \dgr{MSc, PhD}
%% \author*[1,2]{\pfx{Dr} \fnm{Joergen W.} \spfx{van der} \sur{Ploeg} \sfx{IV} \tanm{Poet Laureate} 
%%                 \dgr{MSc, PhD}}\email{iauthor@gmail.com}
%%=============================================================%%

\author*[1]{\fnm{Hadis} \sur{Anahideh}}\email{hadis@uic.edu}

\author[1]{\fnm{Nazanin} \sur{Nezami}}\email{nnezam2@uic.edu}
% \equalcont{These authors contributed equally to this work.}
\author[2]{\fnm{Abolfazl} \sur{Asudeh}}\email{asudeh@uic.edu}

% \author[1,2]{\fnm{Third} \sur{Author}}\email{iiiauthor@gmail.com}

\affil[1]{\orgdiv{Mechanical and Industrial Engineering Department}, \orgname{University of Illinois Chicago}, \orgaddress{\street{842 W. Taylor St}, \city{Chicago}, \postcode{60607}, \state{IL}, \country{USA}}}
\affil[2]{\orgdiv{Computer Science Department}, \orgname{University of Illinois Chicago}, \orgaddress{\street{851 S. Morgan St}, \city{Chicago}, \postcode{60607}, \state{IL}, \country{USA}}}

% \affil[2]{\orgdiv{Department}, \orgname{Organization}, \orgaddress{\street{Street}, \city{City}, \postcode{10587}, \state{State}, \country{Country}}}

% \affil[3]{\orgdiv{Department}, \orgname{Organization}, \orgaddress{\street{Street}, \city{City}, \postcode{610101}, \state{State}, \country{Country}}}

%%==================================%%
%% sample for unstructured abstract %%
%%==================================%%

\abstract{
It is of critical importance to be aware of the historical discrimination embedded in the data and to consider a fairness measure to reduce bias throughout the predictive modeling pipeline. Given various notions of fairness defined in the literature, investigating the correlation and interaction among metrics is vital for addressing the unfairness. Practitioners and data scientists should be able to comprehend each metric and examine their impact on one another given the context, use case, and regulations. Exploring the combinatorial space of different metrics for such examination is burdensome. To alleviate the burden of selecting fairness notions for consideration, we propose a framework that estimates the correlation among fairness notions. Our framework consequently identifies a set of diverse and semantically distinct metrics as representative for a given context. We propose a Monte-Carlo sampling technique for computing the correlations between fairness metrics by indirect and efficient perturbation in the model space. Using the estimated correlations, we then find a subset of representative metrics. The paper proposes a generic method that can generalize to any arbitrary set of fairness metrics. We showcase the validity of the proposal using comprehensive experiments on real-world benchmark datasets.}


%%\pacs[JEL Classification]{D8, H51}

%%\pacs[MSC Classification]{35A01, 65L10, 65L12, 65L20, 65L70}

\maketitle

\section{Introduction}\label{sec:intro}
% \textbf{Motivation}

% \abol{let's make sure to use "reducing unfairness" instead of "mitigating unfairness"}

Machine learning (ML) has become one the most applicable and influential tools to support critical decision makings such as college admission, job hiring, loan decisions, criminal risk assessment, etc. \citep{makhlouf2021applicability}. Widespread applications of ML-based predictive modeling have induced growing concerns regarding social inequities and unfairness in decision-making processes. With fairness being critical in practicing responsible machine learning, \emph{fairness-aware learning} has been the primary goal in many recent machine learning developments.

{Fairness-aware learning} can be achieved by pre-processing, in-processing, or post-processing intervention strategies~\citep{friedler2019comparative}. 
Pre-processing strategies involve the fairness measure in the data preparation step to mitigate the potential bias in the input data and produce fair outcomes~\citep{kamiran2012data,feldman2015certifying,calmon2017optimized}. In-process approaches~\citep{agarwal2018reductions,celis2019classification,zafar2015fairness} incorporate fairness in the design of the algorithm to generate a fair outcome. Post-process methods \citep{hardt2016equality,kamiran2010discrimination}, manipulate the model outcome to mitigate unfairness.

% \hadis{two level: experts mask out irrelevant notions and then data scientist have to choose among the rest which are highly correlated or contradictory. Prior studies showed that}

Fairness is an abstract term with many definitions.
The literature on 
% As a result of previous efforts, 
\emph{Fairness in ML} encompasses more than 21 fairness metrics~\citep{narayanan2018translation,verma2018fairness}. % that could be utilized to mitigate the bias in a given model. 
\neww{
On the other hand, impossibility theorems prove that, assuming the existence of bias in the underlying data, it is not possible to {\em fully} mitigate all unfairnesses\citep{kleinberg2016inherent,garg2020fairness,bakalar2021fairness}.
As a result, it is of interest to {\em reduce unfairnesses} of a model for the set of relevant measures~\citep{zhang2021fairrover}.
While systems such as \cite{omnifair} are capable of reducing model unfairness for multiple definitions, 
their time complexity is exponential to the number of fairness constraints; hence those are not practical a set of fairness metrics that is not small~\cite{zhang2021fairrover}.
Although selecting a small subset of relevant fairness metrics is important, 
selecting such a set is not straightforward since
it is not clear how the fairness metrics interact and whether addressing one will benefit or harm the other.
}

\neww{
% In particular, different fairness metrics can be proper for different context, requiring the domain knowledge which machine learning practitioners may not have.
Besides, different fairness metrics may be appropriate for different contexts, requiring domain expertise that machine learning practitioners may lack.
% We argue that domain experts also may not have a comprehensive domain knowledge to properly identify all relevant fairness metrics, or to know the implications of applying specific fairness metrics.
We argue that domain experts may lack the comprehensive domain knowledge required to properly identify all relevant fairness metrics or understand the implications of applying specific fairness metrics.
% For example, consider an education researcher with some knowledge about the type of disparities influencing the college attainment. First, it may still not be easy for them to identify and formally define appropriate notions to capture disparities.
Consider an education researcher who is familiar with the types of disparities that influence college attainment. First, identifying and properly defining appropriate notions to capture disparities may still be difficult for them.
% Second, considering their limited knowledge, they may miss to specify some relevant measures. 
Second, given their limited scope of knowledge, they may overlook some important metrics. 
Third, domain specialists are unable to determine the impact of resolving identified disparities on the developed model and other aspects of fairness that were overlooked.
% Third, they cannot know the impact of resolving the identified disparities on the generated model and the other aspects of fairness they did not consider.
% For domain experts or machine learning practitioners, choosing the proper fairness notion could depend on the context. For example, the choice of fairness may depend on the domain expert knowledge (e.g., education researchers) about the type of disparities influencing the outcome of interest (e.g., college attainment) and its corresponding independent variables in the specific application domain. However, formalizing and evaluating the appropriate notion is often inaccessible to practitioners due to the limited awareness of the applicability of fairness notions within a given problem. 
% Furthermore, the practitioner's comprehension or expectation of the fairness aspects within a certain context is enough to merely mask out irrelevant notions before addressing the unfairness.
}

\neww{
While it is not easy for domain experts to specify all relevant measures, they
should be able to help masking out the irrelevant measures.
Even after removing irrelevant metrics, data scientists responsible for model building and unfairness reduction may still need to choose or prioritize among many metrics due to the abundance of fairness notions in the literature.}
%\hadis{is it because of this sentence that reviewers thought we are trying to mitigate all notions? should we revise?}
% Also, if the practitioners want to address significant amount of unfairness overall, addressing all is impossible.
More importantly, even if chosen properly, there is no indication that the notions of fairness chosen overlap or are related to one another, which is critical to know when unfairness need to be addressed using the available mitigation strategies.
% to support the fact that to what extend the selected fairness metrics overlaps or correlate with each other.
% enables to describe the overall unfairness and is adequate to address it within a given problem.
% \hadis{what challenge?}. \abol{I don't understand the prev. sentence}
\neww{According to our hypothesis, if one can specify the correlations between the relevant fairness metrics, they could indeed group the metrics that are positively correlated in such a way that reducing unfairness for one metric in the group also reduces unfairness for the other metrics in the group. 
% Our idea is that 
% if one can specify the correlation between the relevant fairness metrics, they can use correlation values to group the positively correlated metrics such that reducing unfairness for one of the metrics in the group reduces unfairness for other metrics in the group as well.
As a result, they can now focus on the group representatives. Since some of the representatives may be negatively correlated, reducing unfairness for one group, however, may cause a rise in unfairness in the other. Nevertheless, only focusing on the group representatives, the practitioner is able to {\em balance} unfairness among different representatives such that the model is relatively fair for all group representatives. As we previously explained, if the number of metrics is constrained, some algorithms can discover a balanced reduction among all. Alternately, by limiting the range of fairness metrics, the practitioner can achieve a workable balance in the reduction of unfairness by performing a scenario analysis on the extent of the reduction (the allowed unfairness gap threshold) using mitigation strategies for each distinct representative metric.}
% due to the correlations and trade-offs between metrics,  improving algorithms to satisfy a given notion of fairness, may induce unfairness on some other aspects.} 
% In fact, the potential intractability (correlation and contradiction) of various notions thwart the selection process, hinder the unfairness reduction of predictive modeling, and consequently affect the trust of practitioners in ML applicability}.

\neww{Although recent studies have attempted to answer questions such as how to measure fairness and reduce algorithmic bias, little is known about the sufficiency of various fairness notions and their correlations. The correlation between different fairness notions has been shown in~\cite{friedler2019comparative}, where the authors demonstrated an empirical analysis of the effectiveness of the state-of-the-art unfairness mitigation techniques and how addressing one notion affects others. However, only a subset of existing metrics is considered, and there is no recommendation on representative notions.
% that, if reduced, cover a large portion of overall unfairness. %\hadis{is this statement precise? are we mitigating the overall unfairness? "that if mitigated would hurt others less" is this better?}
% The trade-offs between fairness notions have also been studied in the literature~\citep{kleinberg2016inherent,garg2020fairness,bakalar2021fairness}. The existence of such a trade-off exacerbates the difficulty of the fairness metrics selection task, increases the chance of impulsiveness, and may introduce additional bias as a result of a mitigating negatively correlated metrics.
}
% wrong choice.

\neww{Specifying a small set of appropriate fairness metrics to reduce within a specific context is a challenging endeavor that requires a rigorous exploration of the combinatorial space of fairness metrics and their interactions.
Therefore, in this paper, we aim to elaborate on the sufficiency of different fairness metrics in a given problem by considering their potential overlaps. We develop {\em an automatic tool} to guide practitioners in identifying the representative metrics and subsequently use the off-the-shelf tools to reduce unfairness in light of estimated correlations. To evaluate our findings, we conduct comprehensive experiments using real-world benchmark datasets, multiple types of classifiers, and a comprehensive set of fairness metrics. Our experiment results verify the effectiveness of our approach in finding representative metrics.}

% \hadis{we can discuss that even with the knowledge of which notions is useful within a context there are many metrics that can measure a certain aspect of unfairness, which is hard to choose. Also, if the practitioners want to address significant amount of fairness overall, addressing all is impossible.} \hadis{we can add a discussion here on how correlation and contradiction of notions make the selection harder for bias mitigation and in general addressing unfairness in practice.}

% Abol: the following paragraph was repeatative and not needed
% \newww{
% Specifying the representative fairness metrics to consider practical implications for data scientists and machine learning practitioners. It could also be usefully deployed in systems where data and models are stored. Having only a subset of all metrics is a feasible way to evaluate the fairness of a model/data. 
% We propose {\em an automatic framework} to detect the interactions of fairness metrics and a small subset of representative metrics of other metrics subsequently, which if addressed will cover a large fraction of overall unfairness \hadis{same comment: "if addressed will less hurt others or less introduce unfairness from other perspectives"}}
In summary, we make the following contributions:

\begin{itemize}
    \item \neww{We propose a framework to detect the potential correlations across fairness metrics that is essential to be aware of in order to audit and mitigate unfairness.}
    \item \neww{By utilizing correlations between metrics, we pose the challenge of identifying a small subset of metrics that properly represent other fairness metrics in a given context (specified by training data and model type).
    Identifying this subset is essential for unfairness reduction because existing solutions are only practical for a limited set of fairness metrics.
    } 
    % \item We propose the problem of using the correlations between different fairness metrics, for a given context (specified by training data and a model type), to find a small subset of metrics that represent other fairness metrics. To the best of our knowledge, we are the first to propose this problem.
    \item We design a sampling-based Monte-Carlo method to estimate the correlations between the fairness metrics.
    \item We develop an efficient approach for sampling models with different fairness metric values, which enables estimating the correlation between the fairness metrics.
    % \item We adapt and extend the existing work \citep{friedler2019comparative,kleinberg2016inherent,bakalar2021fairness} in order to specify the small subset of representative metrics, using the correlations between fairness metrics.
    \item We show, both theoretically and experimentally, that unfairness reduction on a representative metric approximately reduces unfairness on the ones it represents.
    % \item \newww{To evaluate our findings, we conduct comprehensive experiments using real-world benchmark datasets, multiple types of classifiers, and a comprehensive set of fairness metrics. Our experiment results verify the effectiveness of our approach in finding representative metrics.} 
\end{itemize}

% \textbf{Paper Organization}

In the following, we first provide a brief explanation of the fairness metrics in Section~\ref{sec:back}, then propose our technical solution to identify the representative metrics in Section~\ref{sec:alg}, and finally show empirical results to support the effectiveness of the proposal in Section~\ref{sec:exp}.

\section{Fairness Model}\label{sec:back}

% In the last few years an incredible number of definitions have been proposed, formalizing different perspectives from which to assess and monitor fairness in decision making processes. A popular tutorial presented at the Conference on Fairness, Accountability, and Transparency in 2018 was titled “21 fairness definitions and their politics” (Narayanan, 2018). The number has grown since then.
Consider a training dataset $D$ consisting of $n$ data points denoted by the vectors $(X,S,Y)$. Let $X$ be the set of non-sensitive attributes of dimension $d$, $S$ be the set of sensitive attributes specifying the sensitive groups (e.g. {\tt \small gender=female}), and $Y$ be the true response variable (label). In a classification setting $Y={1,\dots, K}$ with $K$ being the total number of distinct classes.
% In this paper, we focus on binary classification problems where $Y=\{0,1\}$. However, the proposed framework can be extended to consider multi-label classification problems. 
Let $h$ be the classifier function where $h: (X,S)\longrightarrow Y$. Let $\hat{Y}$ denote the predicted labels in a given test problem. 
Let $\mathcal{F}$ denotes the set of $m$ group fairness metrics, $\mathcal{F}=\{f_1, f_2,\dots,f_m\}$. Since fairness metrics are defined for each sensitive group, let $f_j^s$ be the fairness metric $j$ defined for sensitive group $s$.

% Table~\ref{tab:confusion} represents the confusion matrix in a given binary classification problem where positive class corresponds to the privileged protected group. 
% Let $A$ represent the sensitive/protected attribute.
Most of the existing fairness notions are defined based on the joint distribution of the $S$, $Y$ and $\hat{Y}$ variables, and fall into one of three well-known categories of \emph{Independence}($f \indep S$), \emph{Separation}($f \indep S \vert Y$), \emph{Sufficiency}($Y \indep S \vert f$) \cite{barocas2017fairness}. Various fairness metrics have been proposed in the literature, each based on one of the aforementioned categories. %\hadis{describe notions briefly in words e.g. demo parity required equal positive predictions for all subrgoups.}
% Let $N$ be the total number of observations. 
% The aforementioned four notions will then enable us to define and measure many fairness measures. 
Following the literature on defining the fairness metrics and for the ease of explanation let us consider a binary classification $h$ and a binary sensitive attribute $S$ \footnote{
We use single binary sensitive and label attribute for the simplicity of explanation. The techniques proposed in this paper, however, are not limited to these cases. 
Besides, our techniques are agnostic to the choice of fairness metrics and machine learning tasks. 
% Extensions to higher numbers and carnality of the sensitive attributes and class labels are straightforward as we shall explain in the discussion section
}. For the purpose of our analysis, we assume $S=1$ and $S=0$ to represent the \emph{Privileged} and \emph{Unprivileged} sensitive groups, respectively. The fairness metrics can be derived by expanding the confusion matrix on the outcome of $h$ split according to each sensitive attribute value \citep{kim2020fact}.  Let $TP$ (True Positive), $FN$ (False Negative), $FP$ (False Positive), and $TN$ (True Negative) be the elements of a confusion matrix. Given the binary sensitive attribute $S=\{0,1\}$, a split of the confusion matrix on the Privileged group is denoted by $TP_1$, $FN_1$, $FP_1$, and $TN_1$, and the total number of observations for the Privileged group is denoted by $N_1$. Table~\ref{tab:notions} demonstrates major fairness metrics that we consider in this paper. For instance, Statistical parity (i.e. $\vert P(\hat{Y}=1 \vert S=1)-P(\hat{Y}=1 \vert S=0) \vert $) would be equivalent to $\frac{TP_1+FP_1}{N_1}=\frac{TP_0+FP_0}{N_0}$ which measures the positive prediction outcome ($\hat{Y}=1$) among different sensitive groups without considering their true $Y$ label. Similarly, Equalized odds (i.e. $ \vert P(\hat{Y}=1 \vert Y=y,S=1)-P(\hat{Y}=1 \vert Y=y,S=0) \vert,  \forall {y \in \{0,1\}}$) can be expressed as $\frac{FP_0}{FP_0+TN_0}=\frac{FP_1}{FP_1+TN_1}$ and $\frac{TP_0}{TP_0+FN_0}=\frac{TP_1}{TP_1+FN_1}$ which emphasizes on positive prediction outcome and measures false positive and true positive rates among sensitive groups. 
% \nazanin{Moreover, we shall define other metrics, derived from the confusion matrix, to be used in calculating unfairness gaps which includes Error rate difference $ERD=\frac{(FP+FN)}{(N_{1}+N_{0})}$,
% False discovery rate 
% $FDR=\frac{FP}{(TP+FP)}$, and 
% False Omission rate $FOR=\frac{FN}{(TN+FN)}$.} 
\neww{For the detailed definition of other notions, please refer to the prior works \citep{narayanan2018translation,verma2018fairness,makhlouf2021applicability}.}
%FPR=\frac{FP}{FP+TN}
%TPR=\frac{TP}{TP+FN}
%FNR=\frac{FN}{FN+TP}
% \hadis{we have more than this though based on differences and ratios, we need to explain.}
Note that when $K>2$, the fairness metrics can be defined upon multiple confusion matrices split according to a combination of class labels. 

\neww{Although in this paper we focus on a subset of fairness notions provided in Table~\ref{tab:notions} to show the empirical results, our proposed framework is a generic method that can incorporate other notions and can be generalized to arbitrary fairness metrics. Note that in this paper we mainly focus on group-based fairness notions and we do not consider individual fairness aspects. Also, the causal fairness is out of the scope of this project.}

\begin{table}[!tb]
\begin{adjustbox}{width=\columnwidth}
  \centering
  \small
  %\caption{Add caption}
    \begin{tabular}{|c|l|r|}
    \toprule
    \textbf{Fairness Notion } & \multicolumn{1}{c|}{\textbf{Label }} & \multicolumn{1}{c|}{\textbf{Formulation}} \\
    \midrule
    Equalized Odds & f1    & \multicolumn{1}{l|}{$\frac{1}{2}* ( \vert \frac{FP_0}{FP_0+TN_0}-\frac{FP_1}{FP_1+TN_1} \vert + \vert \frac{TP_0}{TP_0+FN_0}-\frac{TP_1}{TP_1+FN_1} \vert)$} \\
    \midrule
    Error difference & f2    & \multicolumn{1}{l|}{$\frac{FP_0+FN_0}{N_{1}+N_{0}}-\frac{FP_1+FN
    _1}{N_{1}+N_{0}}$} \\
    \midrule
    Error ratio & f3    & \multicolumn{1}{l|}{$\frac{\frac{FP_0+FN_0}{N_{1}+N_{0}}}{\frac{FP_1+FN_0}{N_{1}+N_{0}}}$} \\
    \midrule
    Discovery difference  & f4    & \multicolumn{1}{l|}{$\frac{FP_0}{TP_0+FP_0}-\frac{FP_1}{TP_1+FP_1}$} \\
    \midrule
    Discovery ratio & f5    & \multicolumn{1}{l|}{$\frac{\frac{FP_0}{TP_0+FP_0}}{\frac{FP_1}{TP_1+FP_1}}$} \\
    \midrule
    Predictive Equality  & f6    &  \multicolumn{1}{l|}{$\frac{FP_0}{FP_0+TN_0}-\frac{FP_1}{FP_1+TN_1}$}\\
    \midrule
    FPR ratio & f7    & \multicolumn{1}{l|}{$\frac{\frac{FP_0}{FP_0+TN_0}}{\frac{FP_1}{FP_1+TN_1}}$} \\
    \midrule
    False Omission rate (FOR) difference  & f8    & \multicolumn{1}{l|}{$ \frac{FN_0}{TN_0+FN_0}-\frac{FN_1}{TN_1+FN_1}$} \\
    \midrule
    False Omission rate (FOR) ratio  & f9   & \multicolumn{1}{l|}{$\frac{\frac{FN_0}{TN_0+FN_0}}{\frac{FN_1}{TN_1+FN_1}}$} \\
    \midrule
    Disparate Impact & f10   & \multicolumn{1}{l|}{$\frac{\frac{TP_0+FP_0}{N_0}}{\frac{TP_1+FP_1}{N_1}}$} \\
    \midrule
    Statistical Parity  & f11   & \multicolumn{1}{l|}{$\frac{TP_0+FP_0}{N_0}-\frac{TP_1+FP_1}{N_1}$}\\
    \midrule
    Equal Opportunity  & f12   & \multicolumn{1}{l|}
    {$\frac{TP_0}{TP_0+FN_0}-\frac{TP_1}{TP_1+FN_1}$}\\
    \midrule
    FNR difference  & f13   & \multicolumn{1}{l|}{$\frac{FN_0}{FN_0+TP_0}-\frac{FN_1}{FN_1+TP_1}$} \\
    \midrule
    FNR ratio  & f14   & \multicolumn{1}{l|}{$\frac{\frac{FN_0}{FN_0+TP_0}}{\frac{FN_1}{FN_1+TP_1}}$} \\
    \midrule
    Average odd difference  & f15   & \multicolumn{1}{l|}{$\frac{1}{2}* (\frac{FP_0}{FP_0+TN_0}-\frac{FP_1}{FP_1+TN_1}+\frac{TP_0}{TP_0+FN_0}-\frac{TP_1}{TP_1+FN_1})$} \\
    \midrule
    Predictive Parity & f16   & \multicolumn{1}{l|}
    {$\frac{TP_0}{TP_0+FP_0}-\frac{TP_1}{TP_1+FP_1}$}\\
    \bottomrule
    \end{tabular}%
     \end{adjustbox}
% \vspace{-1mm}
  \caption{Fairness notions}\label{tab:notions}%
%   \vspace{-10mm}
\end{table}%

% \hadis{Explain the challenge here}
Following the discussion on various fairness metrics, \neww{we next present our framework for estimating the correlations between metrics and identifying a small subset of representative metrics in a given context.}

% \textbf{Logistic Regression} (also known as logit) is a popular machine learning classifier in binary classification problems. Although logistic regression is primarily used with dichotomous dependent variables, the model can be extended to be multivariable problems with 3 or more categories \cite{wright1995logistic}. Let $\mathcal{C}: (X,S) \rightarrow y$ be the logit classifier in a given binary classification problem where $y=\{0,1\}$. The logit function is then defined as: 
% \begin{equation}
% \label{eq:logit}
%     \mathbb{C}(X,S)=\frac{1}{1+e^{-X-S}}
% \end{equation}

\section{Identifying Representative Fairness Metrics}\label{sec:alg}
Fairness is an abstract concept with many definitions from different perspectives and in various contexts.
% This variety of definitions, coupled with the impossibility theorems and the trade-off between different definitions make it overwhelmingly complicated for ordinary users and data scientists to select a subset of those to consider.
% After providing the preliminaries and the set of well-known fairness metrics
{Besides, the impossibility theorems prove it is not possible to fully satisfy fairness metrics that are mutually exclusive~\citep{friedler2016possibility}.
As a result, partial mitigation of unfairness is often considered in practice~\citep{omnifair}.
}
Still, the variety of definitions, coupled with the correlation and trade-off between them~\citep{kleinberg2016inherent}, make it overwhelmingly complicated for ordinary users and data scientists \neww{to evaluate the fairness of the predictive outcome.}
% select a subset of those to consider.

\neww{Therefore, in this section, we aim to determine the correlations between metrics and identify a subset of representatives to facilitate the fairness evaluation and acknowledge the implications of unfairness mitigation for a given context.}
% choice of fairness for a given context. 
In particular, 
we use the correlation\footnote{{Note that data and model details cause unfairness. The metrics measure unfairness; hence the correlations between them are not causation.}} between the fairness metrics to identify their similarities\neww{\footnote{The correlation measures the strength and the direction of the relationship between two variables. Note that our proposed framework is independent of the choice of similarity and the correlation can be replaced by other metrics such as Kullback-Leibler divergence \citep{thomas2006elements}.}}. 
That is, we say a fairness measure $f_i$ represents a measure $f_j$ if 
%That is, $\forall f_i\in \mathcal{F}$, there exists a fairness metric $ f_j\in \rep$ such that the correlation between $f_i$ and $f_j$ is ``high''.
$f_i$ and $f_j$ are highly correlated.
Given a universe $\mathcal{F}$ of fairness metrics of interest, we seek to find a subset $\rep$, with a significantly smaller size representing all metrics in $\mathcal{F}$.

% \hadis{Need to clarify in the paper that we are not talking about achieving zero unfairness (impossibility theorem), instead one can consider multiple notions and to try to decrease the unfairness gap for each. If we want we can also clarify that there is no causation between the measures. The data induces the unfairness, and not necessarily one notion causes the other.}
% \hadis{do we want to emphasize on the causation? notions do not cause each other and the unfairness is in the data.}

To this end, we first need to be able to determine the correlations between metrics of fairness for a given context.
Estimating these correlations is a major challenge we shall resolve in this section.
While the general ideas proposed in this section are not limited to a specific ML task, 
% without loss of generality, 
in this paper
we focus on classification for developing our techniques. %\hadis{explain more for regression settings}
We note that given a classifier,
one can audit it and compute its (un)fairness with regard to different metrics.
But single fairness values do not provide enough information to compute correlations. 
On the other hand,
% In other words, it is not clear how trying to resolve unfairness on one metric will impact the other metrics. \hadis{this is studied in \cite{friedler2019comparative}}
% Besides, existing systems for achieving fairness does not seem to provide a promising approach for finding the correlations \hadis{correlations? this is irrelevant to i ii }: 
% (i) existing fairML learning approaches (including the benchmark methods in AIF360~\cite{aif360}) are designed for a subset of fairness metric; (ii) the ones, including~\cite{omnifair}, that cover different metrics are {\em inefficient} when multiple metrics are considered simultaneously -- due to the exponentially large search space they consider.
% In summary, 
the FairML approaches are designed to build fair models, as opposed to finding the correlation between different fairness metrics.

%Correlation calculation requires vectors of values for each metric. These values can be derived constructing multiple classifiers and subsequently different decision boundaries. There exist an infinite pool of possible decision boundaries that can be constructed on a dataset. Hence, choosing appropriate subset of models in a large exploration space for correlation estimation is a challenging task. We first develop an efficient Monte-Carlo sampling technique to overcome this challenge for correlation estimation. Discovering a representative subset of metrics based on the calculated correlation is a critical task, since not only the correlation magnitudes are important for representations but also their connection is the key for trade-offs.

% In this paper, we propose a systematical framework that discovers the underlying correlation and trade-offs between fairness metrics and identifies a subset representing all. 

Therefore, in the following, we design a Monte-Carlo method~\cite{montecarlo, hickernell2013guaranteed} for estimating the underlying correlation and trade-offs between fairness metrics.
Monte-Carlo methods turn out to be both efficient and accurate for such approximations.

After identifying the correlations, we use them to find the set $\rep$ of the representative metrics in Section~\ref{sec:corr}.

% In particular, the proposed frameworks consists of two major steps: (a) correlation calculation through a Monte-Carlo sampling method, and (b) identifying representative subset of metrics considering the correlation values.
% In this section, we provide the details of our proposed automated tool which outputs the selected metrics of fairness in any given problem. In brief, 
% \emph{Choice of Fairness} procedure consists of three major steps: \emph{Bootstrap Data Representation}, \emph{Fairness Table}, and \emph{Fairness Metric Selection}. We shall elaborate on each step in the following sections.  

\subsection{Estimating Between Fairness Correlations}
Monte-Carlo methods use repeated sampling and the central limit theorem for solving deterministic problems~\citep{durrett2010probability}.
At a high level, the Monte-Carlo methods work as follows:
first, they generate a large enough set of random samples; then they use these inputs to estimate aggregate results.
We use Monte-Carlo methods to estimate correlations between the fairness metrics.
The major challenge towards developing the Monte-Carlo method is being able to generate a large pool of samples. 
Every sample is a classifier that provides different values for every fairness metric.
We use a sampling oracle that upon calling it, returns the fairness values for a sampled classifier. We shall provide the details of the oracle in the next subsection.

Correlation is a measure of linear association between two variables $f_i$ and $f_j$. When both variables are random it is referred to as Coefficient of Correlation $r$. The correlation model most widely employed to calculate $r$ is the normal correlation model. The normal correlation model for the case of two variables is based on the bivariate normal distribution~\citep{neter1996applied}.
Having enough repeated sampling ($N>30$), we can assume the variables $f_i$ and $f_j$ follows the Normal distribution (central limit theorem) with means $\mu_i$ and $\mu_j$, and standard deviations of $\sigma_i$ and $\sigma_j$, respectively. In a bivariate normal model, the parameter $\rho_{ij}$ provides information about
the degree of the linear relationship between the two variables $f_i$ and $f_j$, which is calculated as $\rho_{ij}=\frac{\sigma_{ij}}{\sigma_i\sigma_j}$. The correlation coefficient takes values between -1 and 1. Note that if two variables are independent $\sigma_{ij}=0$ and subsequently $\rho_{ij}=0$. When $\rho_{ij}>0$, $f_1$ tends to be large when $f_2$ is large, or small when $f_2$ is small. In contrast, when $\rho_{ij}<0$ (i.e. two variables are negatively correlated), $f_1$ tends to be large when $f_2$ is small, or vice versa. $\rho_{ij}=1$ indicates a perfect direct linear relation and -1 denotes a perfect inverse relation. Since $\sigma_i$ is unknown, a point estimator of $\rho$ is required. The estimator is often called the \emph{Pearson Correlation Coefficient}.

To estimate the correlations, we use the sampling oracle to sample $N$ classifiers $h_k, \forall k=1,\dots,N$, and to calculate fairness values for each.
Let $f_{i,k}$ be fairness metric $i$ of classifier $h_k$, thus the \emph{Pearson Correlation Coefficient} is defined as follow $\forall i, j = 1, \dots,m$:
\begin{align} \label{eq:corr}
% \forall i, j = 1,& \dots,m\\
\nonumber    r_{ij}=&\frac{\sum_{k=1}^K (f_{i,k}^s-\Bar{f}_i^s) (f_{j,k}^s-\Bar{f}_{j})}{\sum_{k=1}^K (f_{i,k}^s-\Bar{f}_i)^2 \sum_{k=1}^K (f_{j,k}^s-\Bar{f}_{j})^2}
\end{align}

\neww{
The set of $N$ samples gives us an unbiased estimation of the correlation between the fairness metrics.
Still, the estimation variance can be high, affecting our identification of representative fairness metrics.
To resolve this issue, We design a two-level approach, where the lower level uses $N$ samples to provide an estimation of the correlation values, and the upper level repeats the correlation estimation process $L$ times (we use the rule of thumb number $L=30$ in our experiments) to reduce the estimation error.
In particular,
}
% In order to reduce the variance of our estimation, 
% we repeat the estimation $L$ times (we use the rule of thumb number $L=30$ in our experiments),
% \hadis{why? what is the difference between N and L is not clear}), 
\neww{
Let the estimated correlation between a pair of fairness metrics $f_i$ and $f_j$ at iteration $\ell$ be $r^\ell_{ij}$.
}
% every iteration $\ell$ returns the correlation estimation $r^\ell_{ij}, \forall i, j = 1, \dots,m$.
Then, the correlation $r^*_{ij}$ is computed as the average of estimations in each round. That is,

% \vspace{-3mm}
\begin{align}
    \forall i, j = 1, \dots,m:~~r^*_{ij}=\frac{1}{L}\sum_{\ell=1}^L r^\ell_{ij}
\end{align}

Using the central limit theorem, $\rho_{ij}$ follows the Normal distribution $\mathcal{N}\big(\rho_{ij},\frac{ \sigma_{ij}}{\sqrt{L}}\big)$.
Given a confidence level $\alpha$, the confidence error $e$ identifies the range $[r^*_{ij}-e, r^*_{ij}+e]$ where 
$$
p(r^*_{ij}-e\leq \rho_{ij}\leq r^*_{ij}+e)= 1-\alpha
$$
Using the Z-table, 
while using the sample variance $s^2_{ij}$ to estimate $\sigma^2_{ij}$, the confidence error is computed as
% $    e = Z(1-\frac{\alpha}{2})\sigma_{ij}$.

% \vspace{-4mm}
\begin{align}
    e = Z(1-\frac{\alpha}{2})\frac{s_{ij}}{\sqrt{L}}
\end{align}
The pseudocode of our correlation estimator is provided in Algorithm~\ref{alg:montecarlo}.
\neww{At every estimation iteration, the algorithm splits the data into training and test-set. It then groups the data according to their label and demographic group.
Next, calling the sampling oracle $N$ times, it finds $N$ samples of fairness values. To make sure it only considers accurate models, the algorithm rejects the samples corresponding to the models with low accuracy.
The algorithm then uses the samples of fairness values to estimate the (Pearson) correlations between each pair of fairness metrics.
Finally, after generating $L$ estimations of the correlation values, the algorithm aggregates the estimations and computes the estimation errors.
}

\begin{algorithm}[!tb]
\caption{{\sc CorrEstimate}}\label{alg:montecarlo}
\begin{small}
\begin{algorithmic}[1]
\Require {Data set $\dee$, Fairness notions of interest $\ef=\{f_1\cdots f_m\}$}
\Ensure {$O$, target data set}
\For{$\ell\gets 1$ to $L$} {\small \tt // number of iterations}
    \State $(\tee,$test-set$)\gets${\bf split}$(\dee)${\small \tt // $\tee$ is training set}
    %\State \hadis{add the filtering if for accuracy}
    \State $\langle\gee_{0}, \gee_{1},\gee_{2},\gee_{3}\rangle\gets \langle\tee[S=0,Y=0],\tee[S=0,Y=1],\tee[S=1,Y=0],\tee[S=1,Y=1]\rangle$
    \State $F\gets[]$ {\small \tt // Table of fairness values}
    \For{$i\gets1$ to $N$}
    % \hadis{this needs to be updated A1,...}
        \State $A_i,F_{1}\cdots F_{m}\gets${\sc SamplingOracle}$(\langle\gee_{0}, \gee_{1},\gee_{2},\gee_{3}\rangle,\ef)$
        % \nazanin{I added $i$ to show ith model}
        \State {\bf if} $A_i < \alpha$ {\bf then} {\bf continue} {\small \tt // reject the model}
        \State $F_{i1}\cdots F_{im}\gets F_{1}\cdots F_{m}$
        % \EndIf
        % \If{$Accuracy < \alpha$} // Reject the model
        %     \State {\bf continue}
        % \EndIf}
    \EndFor
    \State $corr_\ell\gets${\bf corr}$(F)$ {\small \tt // Pearson correlations}
     \EndFor
\For{$i,j\gets 1$ to $m$}
    \State $corr[i,j]\gets${\bf avg}$(corr_1[i,j]\cdots corr_L[i,j])$
    \State $e[i,j]\gets Z(1,\frac{\alpha}{2})$ {\bf stdev}$(corr_1[i,j]\cdots corr_K[i,j])/\sqrt{L}$
\EndFor
\State \textbf{return} $(corr,e)$
\end{algorithmic}
\end{small}
\end{algorithm}

\neww{ 
Pearson correlation is a measure that reflects the strength and direction of a linear relationship between two variables and fails to adequately characterize nonlinear or nonmonotonic relationships \cite{schober2018correlation}. In this paper, identifying the interaction of fairness metrics through estimating the correlation follows the linearity assumption.
% Correlation is a measure of a monotonic association between two variables when the change in the magnitude of one variable is associated with a change in the magnitude of another variable, either in the positive or negative direction. 
This assumption may not hold when the associated mapping between two fairness metrics is nonlinear since the association between two metrics might change across the domain. However, it still provides a reliable guideline for clustering the metrics for unfairness mitigation. In particular, a small positive/negative or zero correlation designates a weak or no linear relationship, which implies there is no relationship between two metrics or the relationship is not linear. In both cases, the metrics need to be treated separately; thus, it does not change the outcome of our proposal. 
% As a result, when the correlation values are high enough, the linear assumption and association between two metrics describes the reality while low correlation values may indicate non-linear relationship.
} 
% \vspace{-4mm}
\subsection{Developing the Sampling Oracle}
Having discussed the estimation of the correlations between the fairness metrics, next we discuss the development details of the sampling oracle.
Upon calling the oracle, it should draw an iid sample classifier and evaluate it for different fairness metrics.
Considering the set of fairness metrics of interest $\mathcal{F}=\{f_1\cdots f_m\}$, the output of the sampling oracle for the $i^{th}$ sample can be viewed as a vector of values $\{f_{1i}\cdots f_{mi}\}$, where $f_{ji}$ is the fairness of sampled classifier for metric $f_j$.
Calling the oracle by the correlation estimator $N$ times forms a table of $N$ samples where each row contains fairness values (Unfairness gaps) for a sampled classifier (Table~\ref{fig:fair_table}).

% \begin{figure}[!tb]
%     \centering
%     \begin{tabular}{|c||c|c|c|c|}
%         \hline
%          Sample ID&$f_1$&$f_2$&$\cdots$&$f_m$ \\ \hline\hline
%          $h_1$&$f_{11}$&$f_{21}$&$\cdots$&$f_{m1}$ \\\hline
%          $h_2$&$f_{12}$&$f_{22}$&$\cdots$&$f_{m2}$ \\\hline
%          $\vdots$&$\vdots$&$\vdots$&$\cdots$&$\vdots$ \\\hline
%          $h_N$&$f_{1N}$&$f_{2N}$&$\cdots$&$f_{mN}$ \\\hline
%     \end{tabular}
%     \caption{Table of fairness values for the $N$ sampled models}
%     \label{fig:fair_table}
%       \vspace{-6mm}
% \end{figure}

Two requirements are important in the development of the sampling oracle. First, since our objective is to find the correlations between the fairness metrics, we would like the 
samples to provide different values for the fairness metrics. 
In other words, the samples should provide randomness over fairness values space.
Besides, since the correlation estimator calls the oracle many times before it computes the correlations, we want the oracle to be efficient.

Our strategies to satisfy the design requirement for the sampling oracle are based on a simple observation: the performance of a model for a protected group depends on the ratio of samples from that protected group and the distribution of their label values~\citep{agarwal2018reductions,omnifair}.
To better explain this, let us consider a binary classifier $h$ and two groups $g_1$ and $g_2$.
Clearly, if all samples in the training data belong to $g_1$ then the model is only trained for $g_1$, totally ignoring the other group.
As the ratio of samples from $g_2$ increases in the training data, the model trains better for this group. Specifically,
since the training error -- the {\em average} error across training samples --  is minimized during the training process, the ratio of $g_1$ to $g_2$ in the training data directly impacts the performance of the model for each of the groups. 
Besides, the other factor that impacts the prediction of the model for a protected group is the ratio of positive to negative samples.
Therefore, as an indirect method to perturb over the space of fairness values, we consider perturbation over the ratios of the samples from each protected group and label values in the training data.

% \begin{minipage}{\textwidth}
    % \begin{minipage}[!b]{0.49\textwidth}
    \begin{table}
    \centering
    \begin{tabular}{|c||c|c|c|c|}
        \hline
         Sample ID&$f_1$&$f_2$&$\cdots$&$f_m$ \\ \hline\hline
         $h_1$&$f_{11}$&$f_{21}$&$\cdots$&$f_{m1}$ \\\hline
         $h_2$&$f_{12}$&$f_{22}$&$\cdots$&$f_{m2}$ \\\hline
         $\vdots$&$\vdots$&$\vdots$&$\cdots$&$\vdots$ \\\hline
         $h_N$&$f_{1N}$&$f_{2N}$&$\cdots$&$f_{mN}$ \\\hline
    \end{tabular}
      \captionof{table}{Table of fairness values for the $N$ sampled models}
      \label{fig:fair_table}
    % \end{minipage}
    \end{table}
    %   \hfill
    % \begin{minipage}[!b]{0.49\textwidth}
    \begin{figure}
    \centering
    \includegraphics[width=0.7\textwidth]{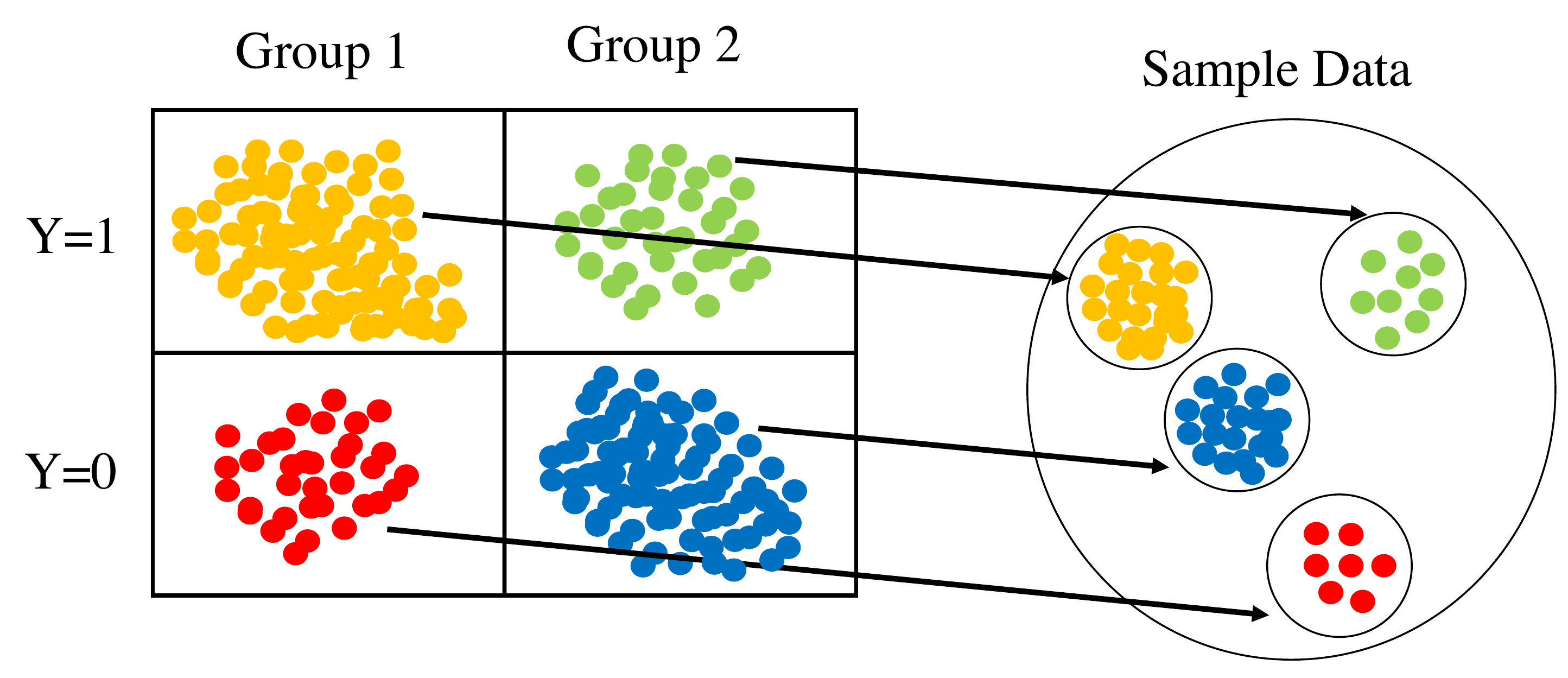}
    \captionof{figure}{Bootstrap Sampling}
    \label{fig:bootstrap}
    \end{figure}
%   \end{minipage}
%   \end{minipage}

% The next observation is that when sampling to estimate the correlations between the fairness metrics, we are not concerned with the accuracy of the model.This gives the opportunity to use sub-sampling to train the model on small subsets of the training data in order to gain efficiency. 
% \hadis{if we do not have enough from each cell then it is affecting the accuracy of the model when the fraction is small (WE NEED TO DISCUSS THIS). If we want to drop the models we need to explain here.}
% \new{Note that any non-representative sub-sampling for traning purpose (e.g. obtained from sampling $g_1$ excluding $g_2$) results in higher misclassification error. To ensure adequate accuracy level in sub-sampling to estimate the correlations between the fairness metrics, we utilize an $\alpha$ accuracy threshold to filter out the inaccurate models. This gives the opportunity to train applicable models on small subsets of the training data in order to gain efficiency.}
Using this observation, we propose a Bootstrap resampling approach to generate different subsamples from the dataset for the training purpose.
{Moreover, to ensure adequate accuracy over generated models, we consider an accept/reject strategy that rejects models with an accuracy below a certain threshold.}
Bootstrapping is a data-driven statistical inference (standard error and bias estimates, confidence intervals, and hypothesis tests) methodology that could be categorized under the broader class of resampling techniques \citep{efron1994introduction,hesterberg2011bootstrap}. The core idea of Bootstrapping is similar to random sampling with replacement without further assumptions.
%In fact, the sampling distribution could be approximated through multiple Bootstrap samples or resamples. However, the number of Bootstrap samples needs to be large enough to effectively represent the Data \citep{hesterberg2011bootstrap}. 
Let $K$ be the number of drawn Bootstraps samples. Given the training dataset $D$, we aim to construct smaller representative subsets of $D$ to train the sampled model.

% \begin{figure}[!tb]
%     \centering
%     \includegraphics[width=0.6\textwidth]{plots/boostrap.pdf}
%     \caption{Bootstrap Sampling}
%     \label{fig:bootstrap}
%     % \vspace{-6mm}
% \end{figure}

% \hadis{alg and the text are not consistent. explain the alg}

% \nazanin{We consider a binary classification problem ($Y\in\{0,1\}$) with two protected groups ($g_1$: Group1, $g_2$: Group2) to describe our sampling procedure. The primary aim is to bootstrap different samples ratios from each of the protected groups and label values as shown in Figure \ref{fig:bootstrap}. To do so, we first generate $r$ ratios drawing from a uniform random  distribution. Next, for each ratio, $w_i$, we sample the training data proportional to the number of observation in each subgroup $|\gee_k|$ and append create the sample training set (t-set). We repeat this process $T$ times, using $T$ different random states,under each ratio. The oracle uses each dataset t-set to train the sampled classifier $h_{it}$ which represents the trained classifier in iteration $t$ while using ratio $i$ to bootstrap different samples. The final step is to evaluate each sampled model's accuracy, $A_it$, as well as evaluating $m$ fairness metrics $f_{i'}$. Let $F_{it}$ represent a vector consisting of $m$ fairness metric values in iteration $t$ and using $i$th ratio for sampling the data. The algorithm~\ref{alg:oracle} will then return $A$, the matrix consisting of elements $A_{it}$, and $F$, the matrix consisting of elements $F_{it}$ as the output.} 

Consider a binary classification problem ($Y\in\{0,1\}$) with two protected groups ($g_1$: Group1, $g_2$: Group2)
to describe our sampling procedure.
In order to (indirectly) control the fairness values, we bootstrap different samples ratios from each of the protected groups and label values
as shown in Figure \ref{fig:bootstrap}. 
Let $\mathbf{w}=\{w_1, w_2,w_3,w_4\}$ be the ratios for each of the cells of the table. To generate each sample, we need to draw the vector $\mathbf{w}$ uniformly from the space of possible values for $\mathbf{w}$. Given that $\mathbf{w}$ represents the ratios from each cell, $\sum_{i=1}^4w_i=1$.
To make sure values in $\mathbf{w}$ are drawn uniformly at random, we first generate four random numbers, each drawn uniformly at random from the range $[0,1]$. Then, we normalize the weights as $w_i=w_i/\sum_{i=1}^4w_i$.
We then bootstrap $w_i\times K$ samples from the samples of $D$ that belong to cell $i$ of the table to form the bootstrapped dataset $B_j$.
Next, the oracle uses the dataset $B_j$ to train the sampled classifier $h_j$.
Having trained the classifier $h_j$, it next evaluates the model to compute the values $f_{jk}$, for each fairness metric $f_k\in\mathcal{F}$, and return the vector $\{f_{j1},\cdots, f_{jm}\}$. {Algorithm~\ref{alg:oracle} also returns $A$, the sampled model's accuracy.}

The set of samples collected from the sampling oracle then form the table of fairness values shown in Table~\ref{fig:fair_table}, which is used to estimate the correlations between the fairness metrics. 
% The Pseudocode of our proposed sampling approach is provided in the Appendix.
The Pseudocode of our proposed sampling approach is provided in Algorithm~\ref{alg:oracle}. 
\neww{
In order to generate a sample vector of fairness values, the algorithm first generates a normalized vector $w_1,\cdots,w_r$, where each value $w_i$ specifies the ratio of samples to be taken from each group $\gee_i$.
Next, to generate a portion of $w_i$ of the target size $t$-size from $\gee_i$, the algorithm generates $w_i\times t$-size random indices, each corresponding to a tuple in $\gee_i$.
After collecting the training data $t$-set, the algorithm uses it to train a model. It next evaluates the model using the test-set, computes the fairness measures for each of the metrics, and returns the results.
}

% \hadis{in alg 2 why do we have test set here? we evaluate fairness on t-set, I think accuracy should be on t-set as well}
\begin{algorithm}[!tb]
\caption{SamplingOracle}\label{alg:oracle}
\begin{small}
\begin{algorithmic}[1]
\Require {$\langle\gee_{0}, \gee_{1},\cdots,\gee_{r}\rangle,\ef$, model-type, t-size, {test-set}}
\Ensure {$F_1\cdots F_m$}
\For{$i\gets 1$ to $r$}
    \State $w_i\gets$ {\bf uniform}$(0,1)${\small \tt // random uniform in range [0,1]}
\EndFor
\State $w_1\cdots w_r \gets \frac{w_1}{\vert \mathbf{w} \vert }\cdots \frac{w_r}{\vert \mathbf{w} \vert }$ {\small \tt // normalize}
\State t-set$\gets[]$ {\small \tt // training set}
\For{$i=1$ to $r$}
    \For{$j=1$ to $w_i\times$t-size}
        \State rand-index$\gets${\bf random}$(1, \vert \gee_i \vert)$
        \State add $\gee_i[$rand-index$]$ to t-set
    \EndFor
\EndFor
% \State model $\gets${\bf train}$($t-set, model-type$)$
\State {model $ \gets$ {\bf train}(t-set, model-type)}
\State {$A \gets$ {\bf evaluate}(model,{\textit test-set})}

\For{$i=1$ to $m$}
    \State $F_i\gets${\bf audit}$($model$, f_i$,{\textit test-set})
\EndFor
\State \textbf{return} $A, F_1\cdots F_m$
\end{algorithmic}
\end{small}
\end{algorithm}

\subsection{Finding the Representative Fairness Metrics using Correlations}\label{sec:corr}

To discover the $\mathcal{R}_{\mathcal{F}}$ representative subset of fairness metrics that are highly correlated, we utilize the correlation estimation from our proposed Monte-Carlo sampling approach described in the previous sections. 

Consider a complete graph of $m$ vertices (denoting each fairness metric $f_i\in\mathcal{F}$), where the weight of an edge $(f_i,f_j)$ between the nodes $f_i$ and $f_j$ is equal to their correlations $r_{ij}$. The goal is to identify the subsets of vertices such that the within subset positive correlations and between subsets negative correlations are maximized.
This problem is proven to be NP-complete~\citep{bansal2004correlation}.
\neww{An exact solution for finding the optimal solution needs to evaluate all subsets in the power-set of the relevant metrics.
As a result, it is inefficient even for tens of metrics.
Therefore, while we underscore that one can solve the problem optimally \citep{queiroga2021integer}, we use the well-known approximation algorithm proposed in
\cite{bansal2004correlation}, which provides a constant approximation ratio for this problem.}
Considering the complete graph of correlations where the edge weights are in $\{+,-\}$, the algorithm first selects one of the nodes as the pivot, uniformly at random.
Next, all the nodes that are connected to the pivot with a $+$ edge are connected to the cluster of the pivot. Next, the algorithm removes the already clustered node from the graph and repeats the same process by selecting the next pivot until all points are clustered.

In order to adapt this algorithm for finding the representative fairness metrics, we consider a threshold $\tau$.
Then, after selecting the pivot (a fairness metric $f_i\in \mathcal{F}$), we connect each fairness metric $f_j\in \mathcal{F}$ to the cluster of $f_i$, if $r_{ij} \geq \tau$.
Moreover, to find a small subset $\mathcal{R}_\mathcal{F}$, we repeat the algorithm multiple times and return the smallest number of subsets.
For every cluster of metrics, the pivot $f_i$ is added to the set of representative metrics $\mathcal{R}_\mathcal{F}$. 
{Considering a positive $\tau>0$, we make sure that the representative metrics positively correlate with the metrics in their cluster.}

% \new{Considering a positive $\tau$, any two notions that are negatively correlated will not fall into one cluster. Also, if the combination of three (or more) notions is negatively correlated, then the pairwise combinations are negatively correlated, then they cannot fall in one cluster. This means that if we improve two simultaneously, it does not reduce the third one (i.e., fixing one will partially fix the other). 
% \hadis{we need to mention that the we are considering partially mitigating unfairness (impossibility theorem}
% \begin{proposition}
% Given $\tau \geq 0$ mutually exclusive fairness notions do not belong to the same cluster.
% \end{proposition}
% \begin{proof}
% We provide the proof by contradiction. Suppose $f_1$, $f_2$, and $f_3$ are mutually exclusive, yet they
% % Lets assume $f_1$, $f_2$, and $f_3$ 
% belong to the same cluster noted by $C_k$. 
% Since the three variables are mutually exclusive at least one pair of those should be negatively correlated CITE. \hadis{I could not find a citation.}
% Thus, $r_{12} \geq \tau \text{ and } r_{13} \geq \tau \text{ and } r_{23} \geq \tau$. Since $\tau \geq 0$ then $r_{12} \geq 0 \text{ and } r_{13} \geq 0 \text{ and } r_{23} \geq 0$. Thus, none of the pairs are mutually exclusive.
% If at least one pair of $\{f_1,f_2,f_3\}$ is mutually exclusive then at least one $r_{ij} < 0$ then $r_{ij} \geq \tau$ is \emph{false} and $f_i \text{ and } f_j \notin C_k$, which contradicts the assumption. Hence, Each pair are positively correlated otherwise they could not be in one cluster.
% \end{proof}
% }

{The selected representatives can be used by the practitioners in the model development process. There has been extensive research in mitigating the unfairness of ML models using different notions as discussed in Section~\ref{sec:related}. 
% Training a classifier taking one of proposed actions on one representative satisfy other notions connected to the selected notion (not selected) in each cluster. 
Using such techniques, the user can mitigate unfairness on the representative metrics.
Lemma~\ref{lem.3.1} shows that mitigating unfairness on a representative metric approximately mitigates unfairness on the metrics it represents.
In Section~\ref{sec:exp}, we demonstrate this phenomenon empirically using different datasets. 
% \hadis{we probably want to discuss more on omnifair and fairrover for simultaneous reduction or sequential consideration. we also can discuss the between representative correlations as well.}
}

% \hadis{Not yet sure about the following.}

{
\begin{lemma}\label{lem.3.1}
Fairness improvement on a representative metric satisfies an approximation ratio of $\frac{1}{\tau}$ over the fairness metrics it represents. 
% Given a unfairness upperbound hyperparameter $\epsilon$ the total ethical consideration using $\tau_1$ is smaller than of that with $\tau_2$ where $\tau_1 > \tau_2$.
\end{lemma}
\begin{proof}
% Let $f_1^*$ be a representative metric, using the correlation threshold $\tau_1$, and let $f_2^*$ be the representative of the given cluster using $\tau_2$. Given $\tau_1 > \tau_2$ and an instance cluster $C$, $r_{ij}^1 \geq r_{ij}^2$ where $r_{ij}^1$ and $r_{ij}^2$ are correlations of notions in an instance cluster using $\tau_1$ and $\tau_2$, respectively. 
% Suppose the goal is to mitigate the unfairness (all notions) by $\epsilon$. 
Let $f^*$ be a representative metric for which the unfairness has been reduced.
% Hence, if we reduce the unfairness of the selected representative $f_1^*$, then 
Let the unfairness reduction on $f^*$ after taking an action for unfairness mitigation be $\epsilon$. That is,
$\Delta f^*=\epsilon$. The unfairness reduction on $f^*$ partially reduces the unfairness of metrics $f_j \in C$ represented by it proportional to their correlations, i.e., $\Delta f_j = r_{ij} \Delta f^* = r_{ij} \epsilon $. 
Thus the unfairness for $f_j$ has been mitigated by an approximation factor of $\frac{\epsilon}{\Delta f_j} = \frac{1}{r_{ij}}$.
% Thus, the approximate unfairness mitigation for $f_j$ connected notions that result in an approximation factor of $\frac{1}{r_{ij}}$. 
Since $r_{ij}\geq \tau, \forall f_j \in C$, the approximation over all fairness metrics represented by $f^*$ is at least $\frac{1}{\tau}$, i.e., $\frac{1}{r_{ij}} \leq \frac{1}{\tau} $.
\end{proof}}

{
\begin{lemma}\label{lem.3.2}
Increasing the value of $\tau$ monotonically increases the number of representative metrics.
\end{lemma}
\begin{proof}
Given $\tau_1 > \tau_2$, if $r_{ij} \geq \tau_1$ then $r_{ij} \geq \tau_2$. As a result, $f_i$ and $f_j$ belong to the same cluster under both scenarios. However, if $\tau_1 \leq r_{ij} \geq \tau_2$ then $f_i$ and $f_j$ belong to one cluster under $\tau_2$ but not $\tau_1$. Note that $\tau_2 \leq r_{ij} \geq \tau_1$ cannot happen. Thus, the number of clusters under $\tau_1$ is always greater than equal of that under $\tau_2$.
\end{proof}}

{
Lemmas~\ref{lem.3.1} and~\ref{lem.3.2} show the trade-off between the number of representative metrics and their effectiveness in fairness mitigation.
On one end, when $\tau$ is a large number, unfairness mitigation on representative metrics accurately mitigates unfairness over other metrics as well. But the number of representative metrics in such cases is large.
On the other hand, when $\tau$ is small, the set of representative metrics is small. But since the correlation between the representatives and the metrics they represent are low, mitigation on them is less effective on the other metrics.
}
% \hadis{as tau increases the number of rep increases but gets more accurate and vice versa.}
% \hadis{We need to discuss how the alg uses this criteria together with tau or separately?
% Directly providing $\tau$ might not be accessible for the user, instead the user can provide the number of representative fairness metrics to be found. binary search}
% \hadis{we can surely say that within each cluster the min unfairness reduction is proportional to the min(Corr(rep,other)) times the epsilon upperbound.  Then the minimum total(across all clusters) reduction would be min(min(...)). In the above theorem I was trying to relate this fact to the choice of tau. However, it is might not be correct since a larger tau results in a larger number of clusters that might have the same reduction amount totally.  }

{
We consider $\tau$ as a user-specified hyper-parameter.
However, directly providing $\tau$ might not be accessible for the user. Instead, the user can provide the number of representative fairness metrics to be found.
% Alternatively, the user can specify the number of representative metrics. 
In such cases, we apply a binary search on the value of $\tau$, to find the corresponding value that returns the user-specified number of representative metrics.
}

\section{Experiments}\label{sec:exp}
% \vspace{-2mm}
   \subsection{Datasets}\label{sec:data}
Our empirical results are based on the benchmark datasets in fair-ML literature\footnote{ https://aif360.readthedocs.io/en/latest/} toolkit \citep{aif360}:

\textbf{COMPAS}\footnote{ProPublica, \url{https://bit.ly/35pzGFj}}: published by ProPublica \citep{propublica}, this dataset contains information of juvenile felonies
such as {\em marriage status, race, age, prior convictions}, etc. 
We normalized data so that it has zero mean and unit variance.
We consider {\tt \small race} as the sensitive attribute and filtered dataset to black and white defendants. The dataset contains 5,875 records after filtering.
We use {\em two-year violent recidivism} record as the true label of recidivism: $Y=1$ if the recidivism is greater than zero and $Y=0$ 
otherwise. We consider \emph{race} as the sensitive attribute.

\textbf{Adult}\footnote{CI repository, \url{https://bit.ly/2GTWz9Z}}: contains 45,222 individuals' income extracted from the 1994 census data with attributes such as {\em age, occupation, education, race, sex, marital-status, native-country, hours-per-week} etc. We use {\em income} (a binary attribute with values $\geq \$50k$ and $\leq \$50k$) as the true label $Y$. The attribute \emph{sex} is considered as the sensitive attribute.

\textbf{German Credit Data} 
\footnote{UCI repository, \url{https://bit.ly/36x9t8o}}: includes 1000 individuals' credit records containing attributes such as marital status, sex, credit history, employment, and housing status. 
We consider both \emph{sex} and \emph{age} as the sensitive attributes, and {\em credit rating} (0 for bad customers and 1 for good customers) as the true label, $Y$, for each individual.

\textbf{Bank marketing}\footnote{UCI repository, \url{https://archive.ics.uci.edu/ml/datasets/Bank+Marketing}}:
is published by \citep{moro2014data} the data is related to direct marketing campaigns number of phone calls of a Portuguese banking institution. The classification goal is to predict if a client will subscribe to a term deposit (variable $Y$). The Dataset contains 41188 and 20 attributes that were collected from May 2008 to November 2010. We consider \emph{age} as the sensitive attribute.

% Protected attributes: Age 

% \textbf{Law School}

% Protected attributes: Race

% https://github.com/microsoft/tempeh

% \textbf{Medical Expenditure Panel Survey Data (MEPS)}

% MEPS is a set of large-scale surveys of families and individuals, their medical providers, and employers across the United States. MEPS is the most complete source of data on the cost and use of health care and health insurance coverage.

% Protected attributes: Race  
% \vspace{-3mm}
\subsection{Performance Evaluation}
In order to estimate between-fairness correlations using our proposed Monte-Carlo method, we use $N=1000$ sampled models for each round and repeat the estimation process for $L=30$ {times on each dataset with at least an $\alpha=0.5$ accuracy level (threshold to filter out the inaccurate models).}
Our proposed approaches are evaluated using a set of commonly-used classifiers; Logistic Regression ({\bf Logit}), Random Forest ({\bf RF}), K-nearest Neighbor ({\bf KNN}) 
%\hadis{what is the arch details of NN} \nazanin{3 layers: one input layer of size20, one dense layer of size 20, and the final output layer of size 1 which use sigmoid activation function (binary cross entropy is the loss function)}
, Support Vector Machines ({\bf SVM}) with linear kernel, and Neural Networks ({\bf NN}) with one dense layer. \neww{Our proposed framework is agnostic to the type of ML model and can incorporate other classifiers as well.}
% Our findings are transferable to other classifiers. \hadis{well how?}
% \new{The algorithm constructs an overall 3000 sample model estimations on each data set with at least an $\alpha$ accuracy level (threshold to filter out the inaccurate models).}
% We begin by evaluating our proposed correlation estimation approach using a wide variety of datasets and classifiers. Then, we focus on identifying the representative subsets of fairness metrics using the approach explained in the previous sections.

% \hadis{discuss accuracy and how dropping less accurate models does not significantly change the estimations (conducting a statistical testing offline).Filtering out models with below some threshold and generate new ones (accept and reject). Is not it computationally inefficient?
% }

\begin{figure*}[!tb]
\centering
% \vspace{-10mm}
\subfloat[COMPAS]{\includegraphics[width=0.5\textwidth]{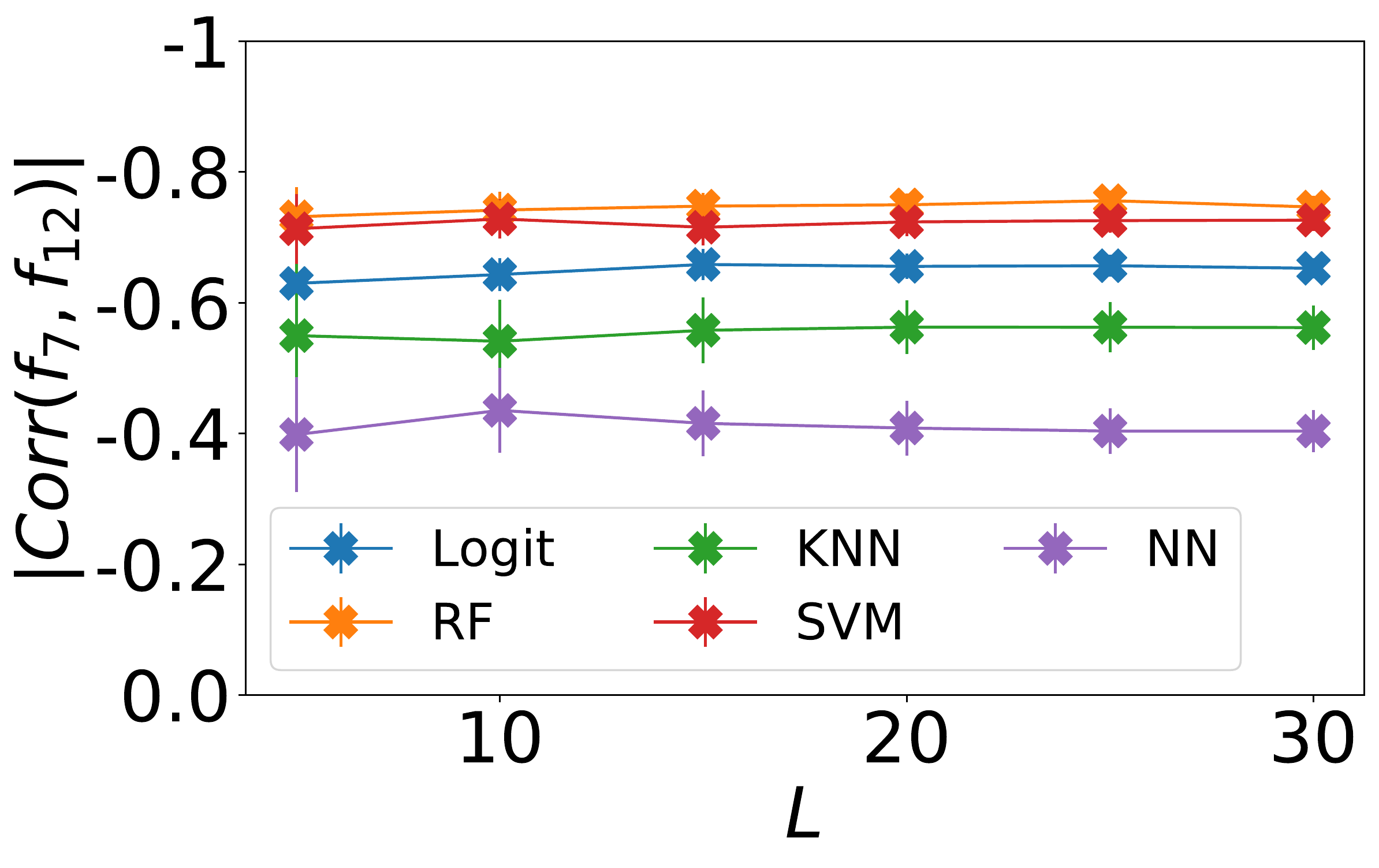}}
\subfloat[Adult]{\includegraphics[width=0.5\textwidth]{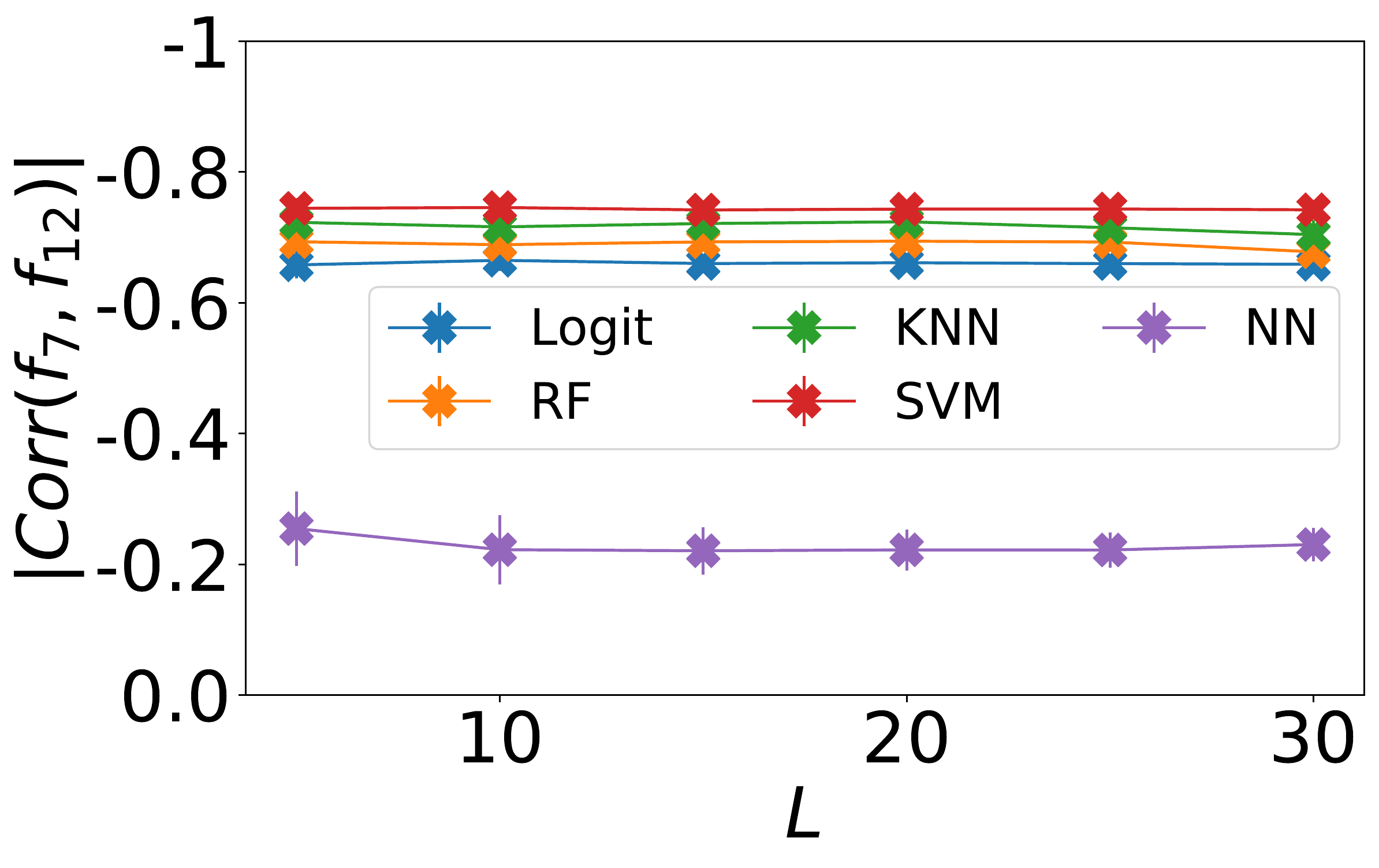}}

\subfloat[Credit]{\includegraphics[width=0.5\textwidth]{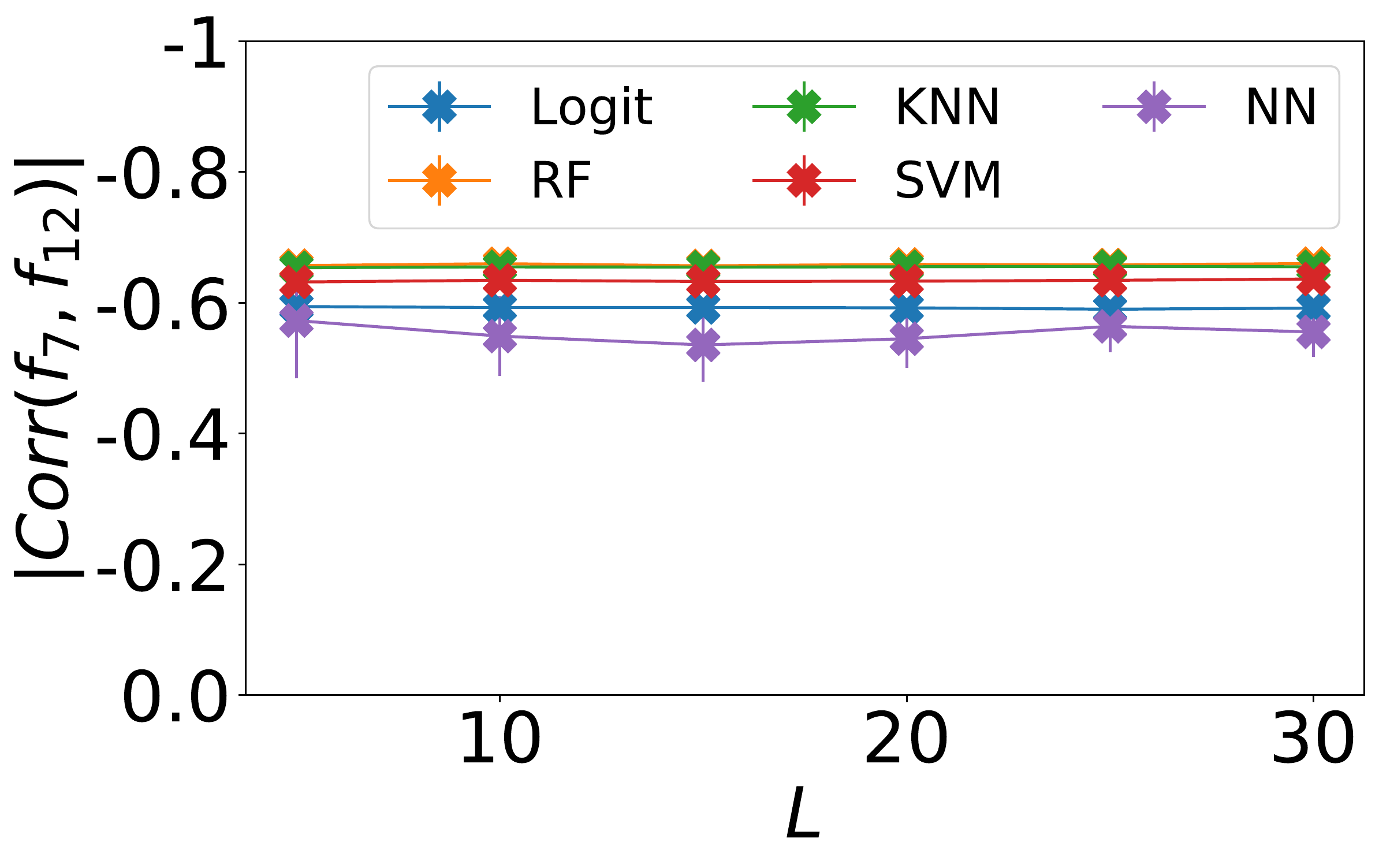}}
\subfloat[Bank]{\includegraphics[width=0.5\textwidth]{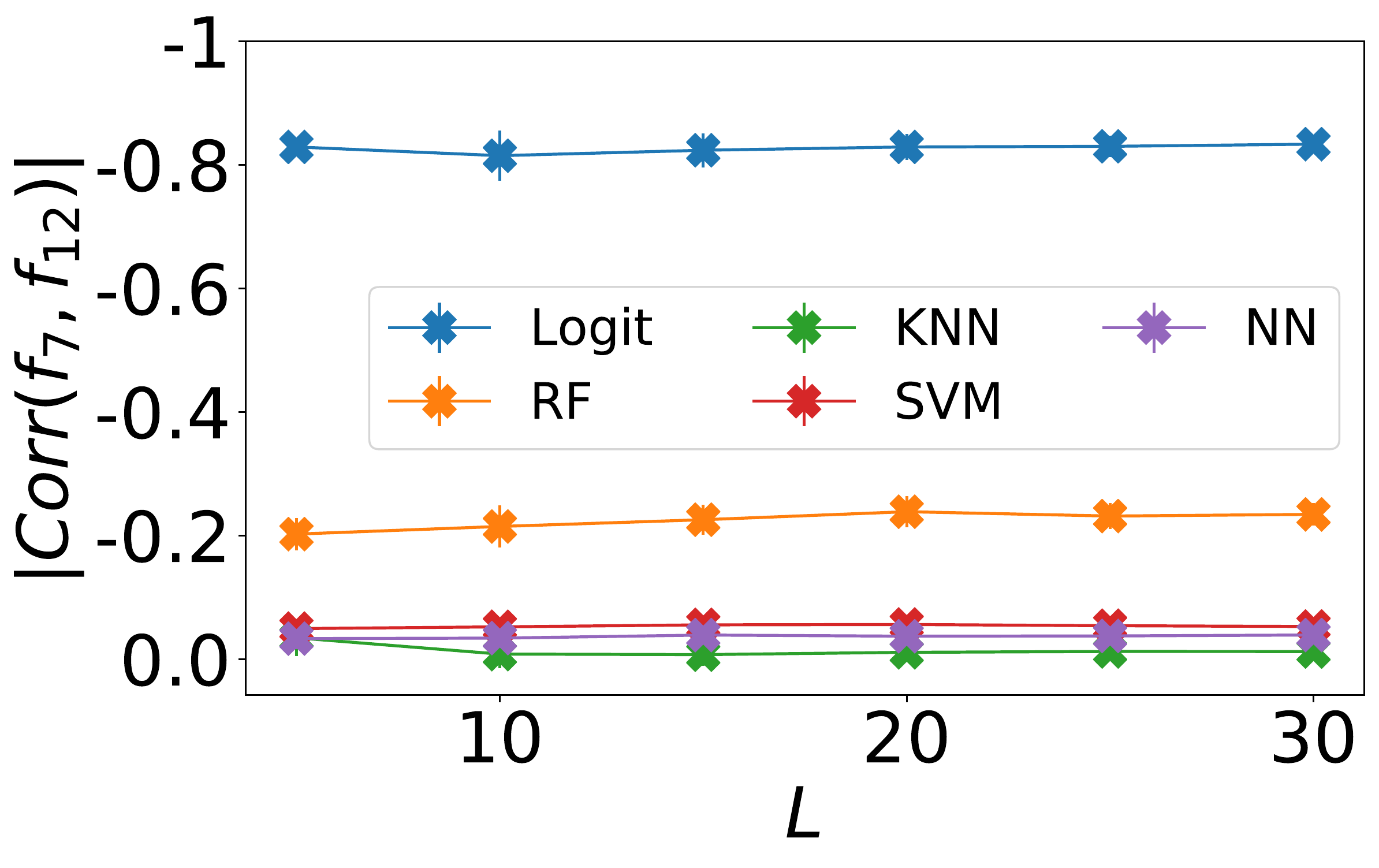}}

\caption{Correlation Estimate for $f_7$ and $f_{12}$ for different datasets}
\label{fig:correstimate}
\hspace{\fill}
\end{figure*}

\noindent{\bf Correlation estimation quality:}
We begin our experiments by evaluating the performance of our correlation estimation method.
Recall that we designed a Monte-Carlo approach for correlation estimation.
In every iteration, the algorithm uses the sampling oracle and samples classifiers to evaluate correlations between pairs of fairness notions, for which we use 1000 samples. To investigate the impact of the number of iterations on the estimation variance and confidence error, we vary the number of iterations from $L=2$ to 30. %\hadis{why the plot label is sample size?}. 
Since the number of estimation pairs are quadratic to $ \vert \mathcal{F} \vert $, we decided to (arbitrarily) pick a pair of notions and provide the results for it. 
To be consistent across the experiments, we fixed the pair $f_7$ and $f_{12}$ for all datasets/models/settings.
We confirm that the results and findings for other pairs of notions are consistent with what is presented for $f_7$ and $f_{12}$.
% \footnote{Complimentary results are provided in the appendix.}
Figure \ref{fig:correstimate} provides the results for $f_7$ and $f_{12}$ for COMPAS (a), Adult (b), Credit (c), Bank (d) datasets.
% based on repeated  estimation that every time calls 
% In our first experiment illustrated in Figure \ref{fig:correstimate}, we demonstrate the correlation estimates of two fairness metrics of $f_7$ and $f_{12}$ for COMPAS (a), Adult (b), Credit (c), Bank (d) datasets. 
Looking at the figure, one can confirm the stable estimation and small confidence error bars, which demonstrate the high accuracy of our estimation.
Also, as the number of iterations increases, the estimation variance and confidence error significantly decrease.
% \hadis{Nazanin did you check the x axis of fig 3?}
% \nazanin{I checked the file: 10,20,30 shows the number of repetition based on different ratios (basically $L$ in algorithm)}
% \nazanin{I also checked the final models that we used. I think the final results show repeating 30 times (using 30 different ratios.csv file) and creating 1000 models each time, and then taking avg and std over 30 repetition. So, the overall number of models used is actually 30000 instead of 3000.}
% The results indicate the robustness of the proposed sub-sampling technique for the correlation estimation. In particular, the estimates of correlation does not change dramatically as we increase the number of samples for estimation. \hadis{we need to add [10, 20, 30] subsets}

\begin{figure*}[!tb]
\centering
% \vspace{-3mm}
\subfloat[Correlation values for $f_7$ and $f_{12}$ for COMPAS]{\includegraphics[width=0.33\textwidth]{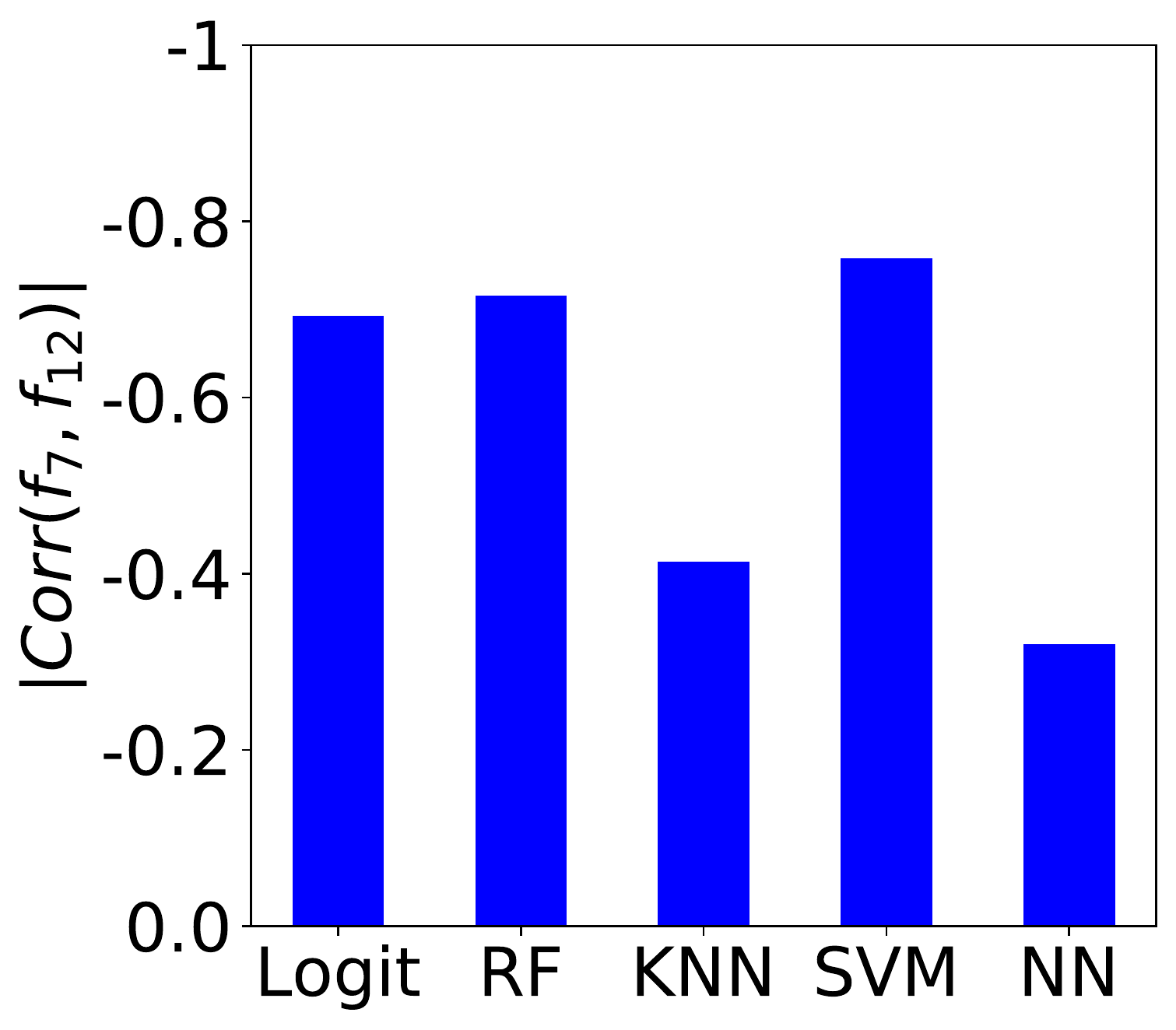}}\hfill
\subfloat[Correlation values COMPAS dataset with Logit]{\includegraphics[width=0.33\textwidth]{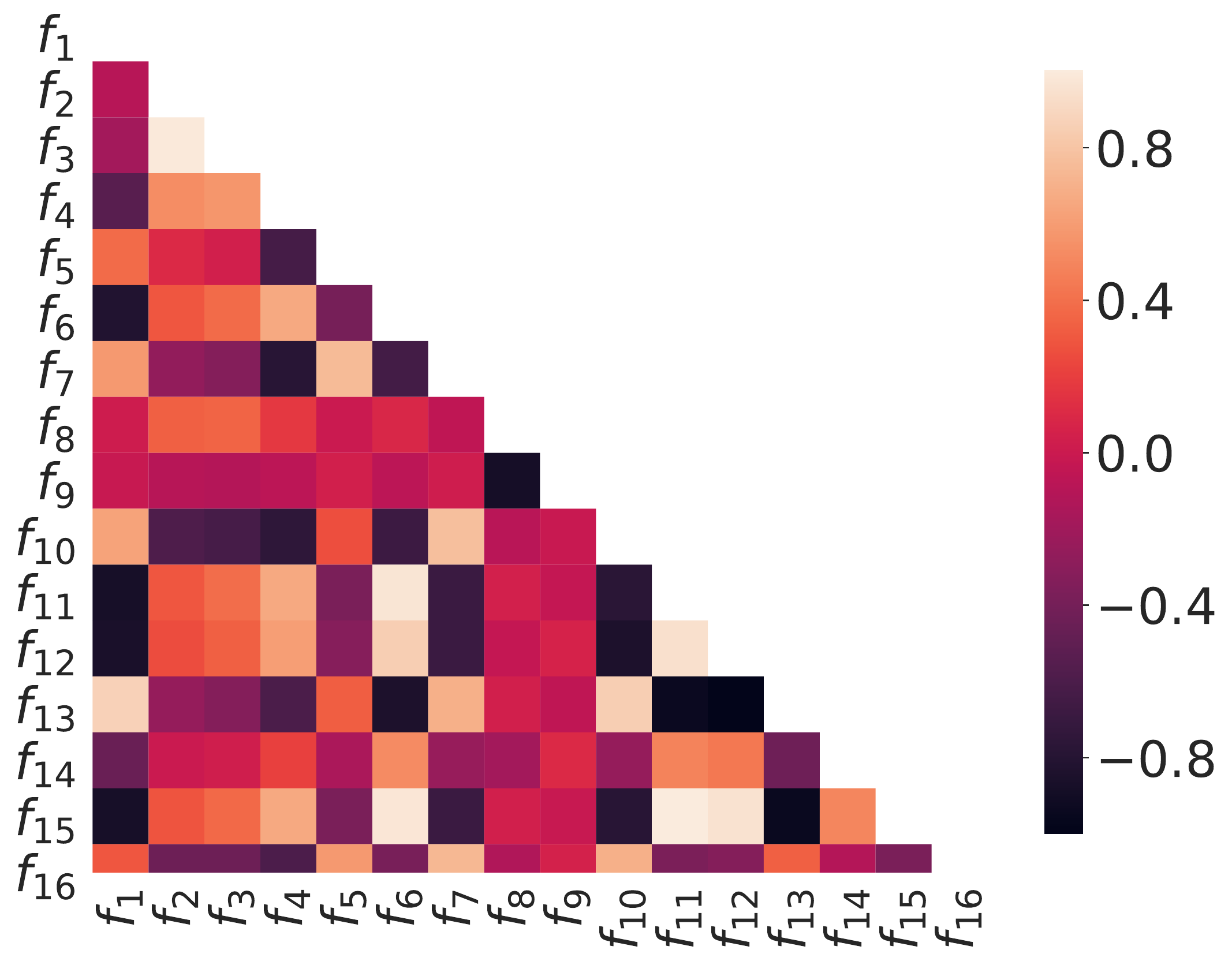}}\hfill
\subfloat[Correlation values COMPAS dataset with NN]{\includegraphics[width=0.33\textwidth]{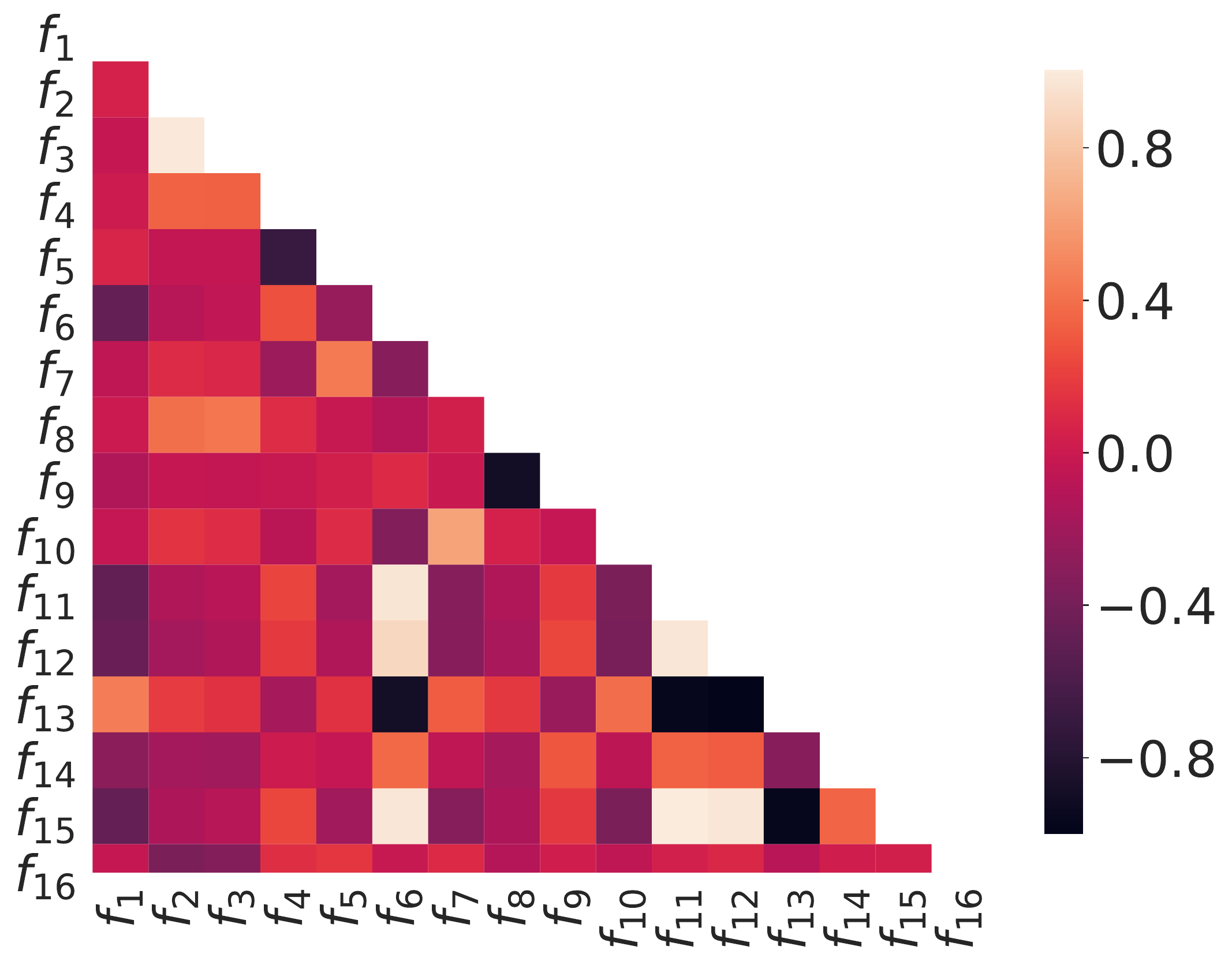}}
% \vspace{-3mm}
\caption{Using COMPAS data set to illustrate that Correlations are model dependent.}\label{fig:heat1}
% \vspace{-4mm}
\hspace{\fill}
% \vspace{-5mm}
\end{figure*}

\begin{figure*}[!tb]
\centering
% \vspace{-3mm}
\subfloat[Correlation values for $f_7$ and $f_{12}$ for Logit]{\includegraphics[width=0.33\textwidth]{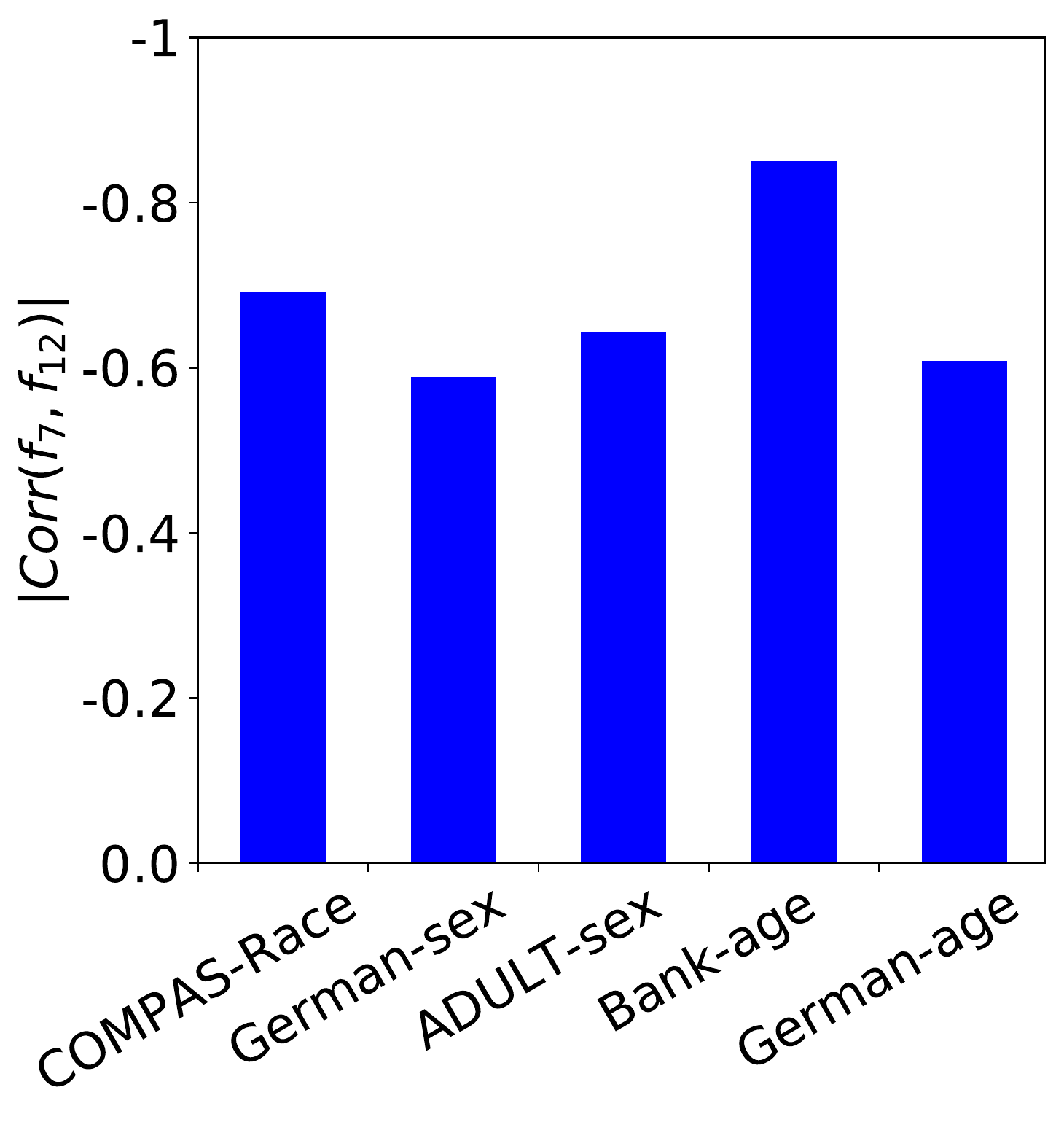}}\hfill
\subfloat[Correlation values Credit dataset with Logit]{\includegraphics[width=0.33\textwidth]{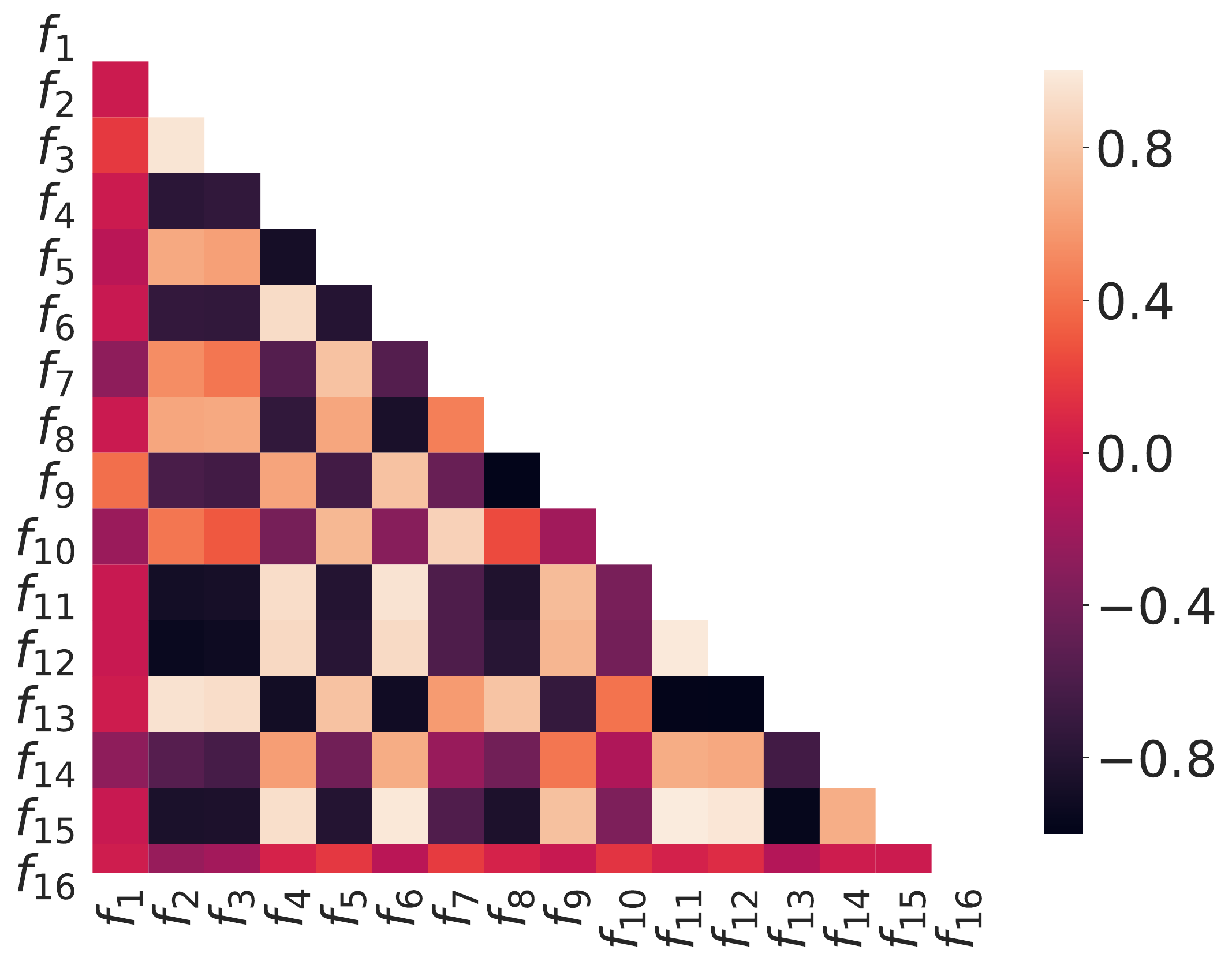}}
\hfill
\subfloat[Correlation values Adult dataset with Logit]{\includegraphics[width=0.33\textwidth]{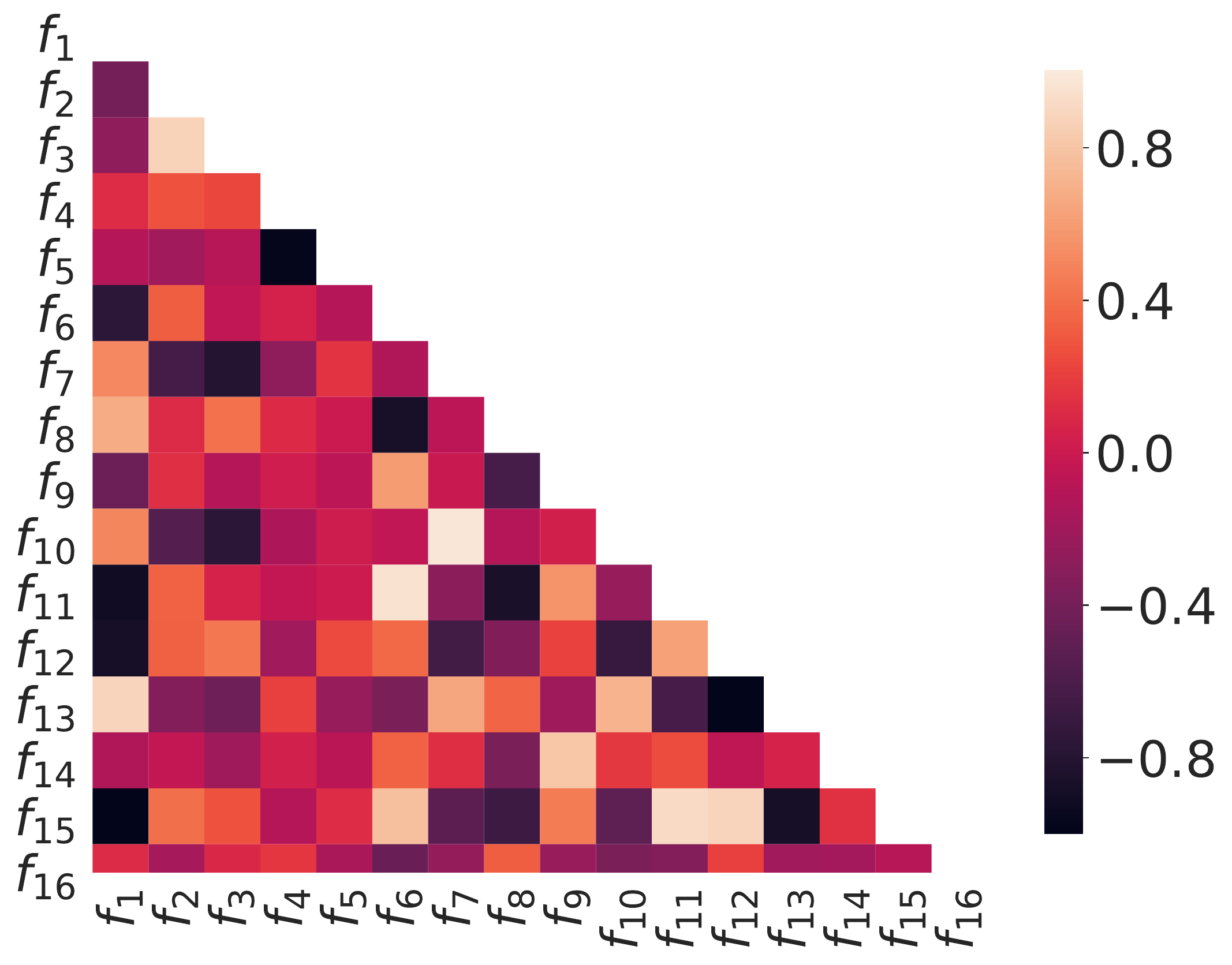}}
% \vspace{-3mm}
\caption{Using Logistic Regression to illustrate that Correlations are data dependent.}\label{fig:heat2}
% \vspace{-4mm}
\hspace{\fill}
% \vspace{-5mm}
\end{figure*}

% The interpretation should also take into account the confidence interval of the observed coefficient as an estimate of what the correlation could plausibly be in the population from which the data were sampled.
\noindent{\bf Impact of data/model on correlation values:} 
In this paper, we propose a framework to identify representative fairness metrics for a given context (data and model).
The underlying assumption behind this proposal is that correlations are data and model-dependent.
Having evaluated our correlation estimation quality, in this experiment, we verify that the data/model-dependent assumptions for the correlations are valid.
To do so, we first fix the dataset to see if the correlations are model-dependent (Figure \ref{fig:heat1}) and then fix the model to see if the correlations are data-dependent (Figure \ref{fig:heat2}). 
First, we confirm that the results for other datasets/models are consistent with what is presented here.
% Looking at Figure \ref{fig:heat1}, it is clear that correlations are model dependent. In particular, generally speaking, the complex boundary of the non-linear models (e.g. NN) reduce the correlation between fairness metrics, compared to linear models (e.g. Logit). 
{Looking at Figure \ref{fig:heat1}, it is clear that correlations are model-dependent. In particular, {\bf NN} which is capable of constructing more complex boundaries, resulting in more flexible models with a wide range of fairness values for different metrics. As a result, the correlations between fairness metrics for {\bf NN}  was in general, less than the other models. %\hadis{do we want to add anything here? model capacity}
% generally speaking, models with complex boundaries (e.g. NN) reduce the correlation between fairness metrics, compared to less complex models (e.g. Logit).
}
Similarly, Figure~\ref{fig:heat2} verifies that correlations are data-dependent. This is because different datasets represent different underlying distributions with different properties impacting the fairness values.

% Figure \ref{fig:heat1} and Figure \ref{fig:heat2} represents the correlation estimates magnitude (absolute value) for -- and -- fir COMPAS dataset using different models. \hadis{discuss}

\begin{figure*}[!tb]
\centering
% \vspace{-7mm}
\subfloat[COMPAS]{\includegraphics[width=0.5\textwidth]{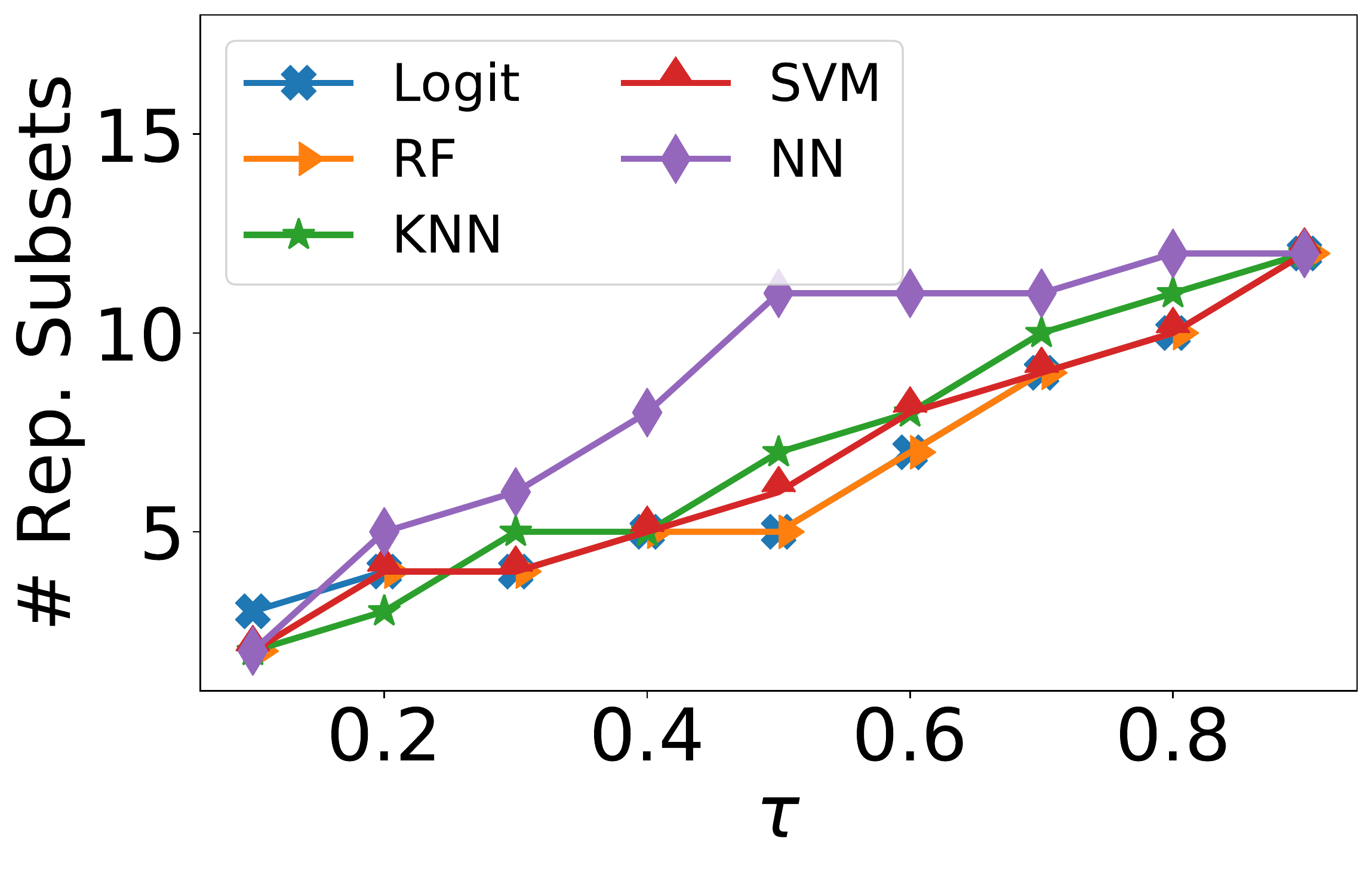}}
\subfloat[Adult]{\includegraphics[width=0.5\textwidth]{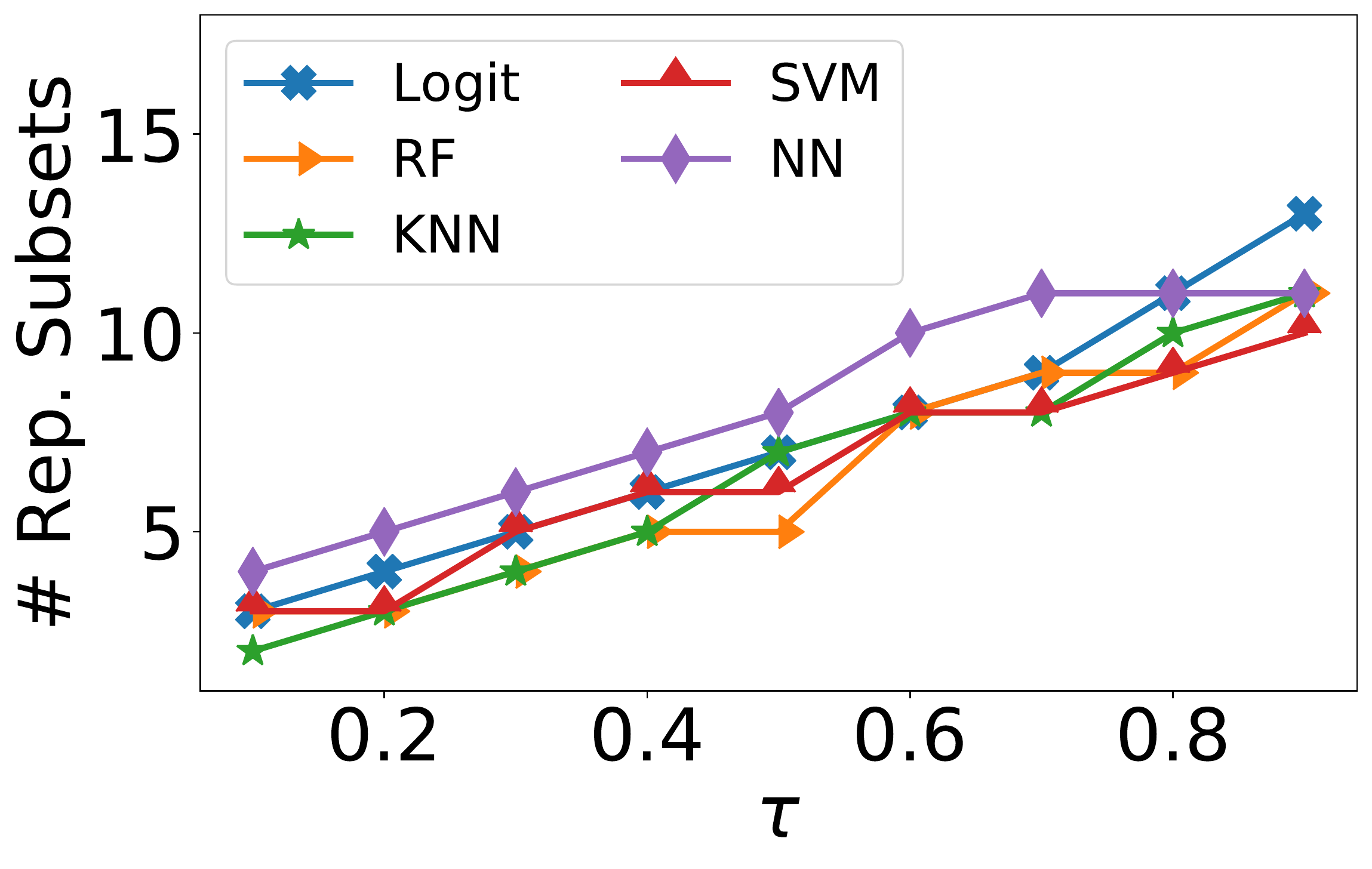}}

\subfloat[Credit]{\includegraphics[width=0.5\textwidth]{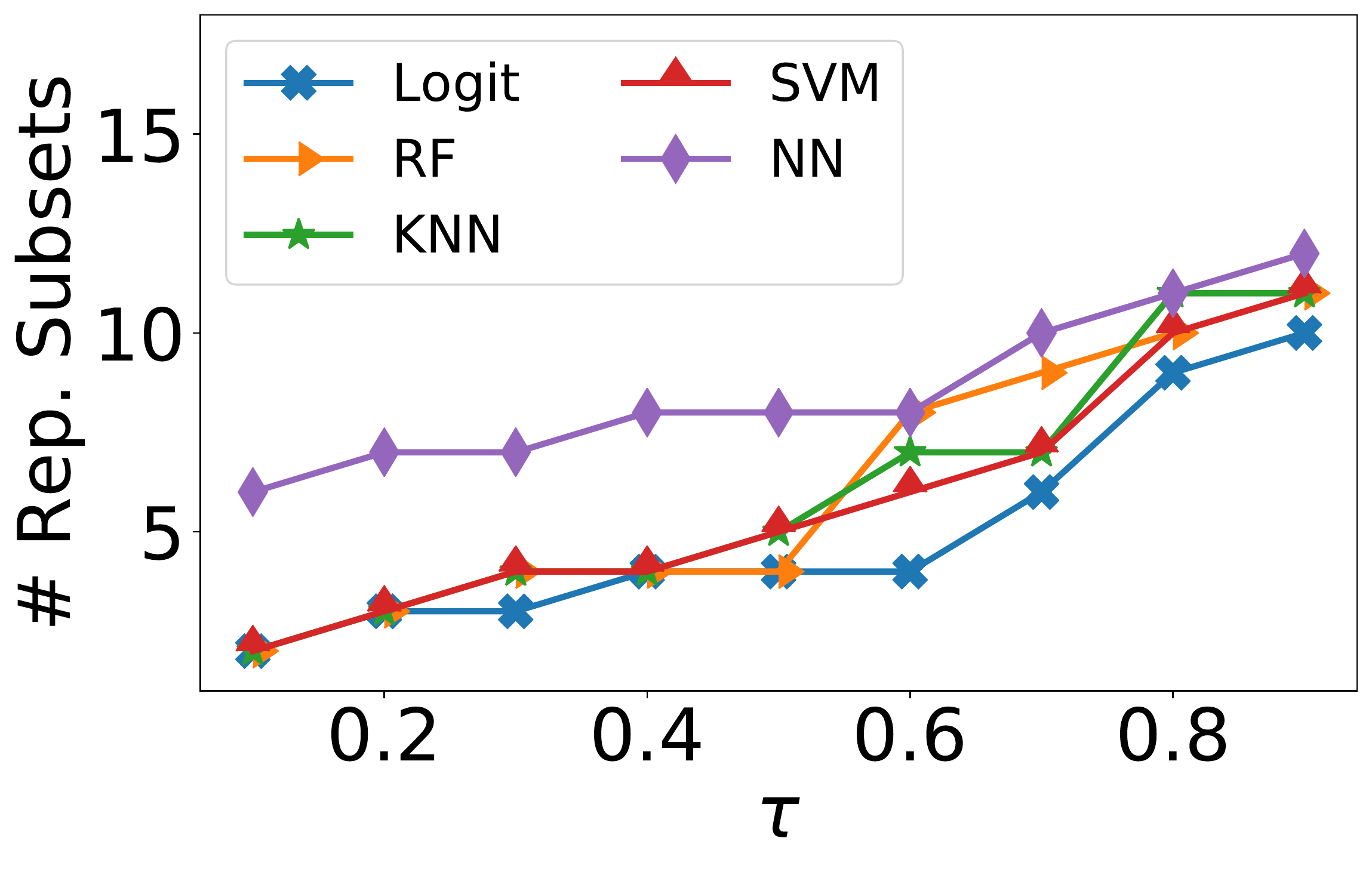}}
\subfloat[Bank]{\includegraphics[width=0.5\textwidth]{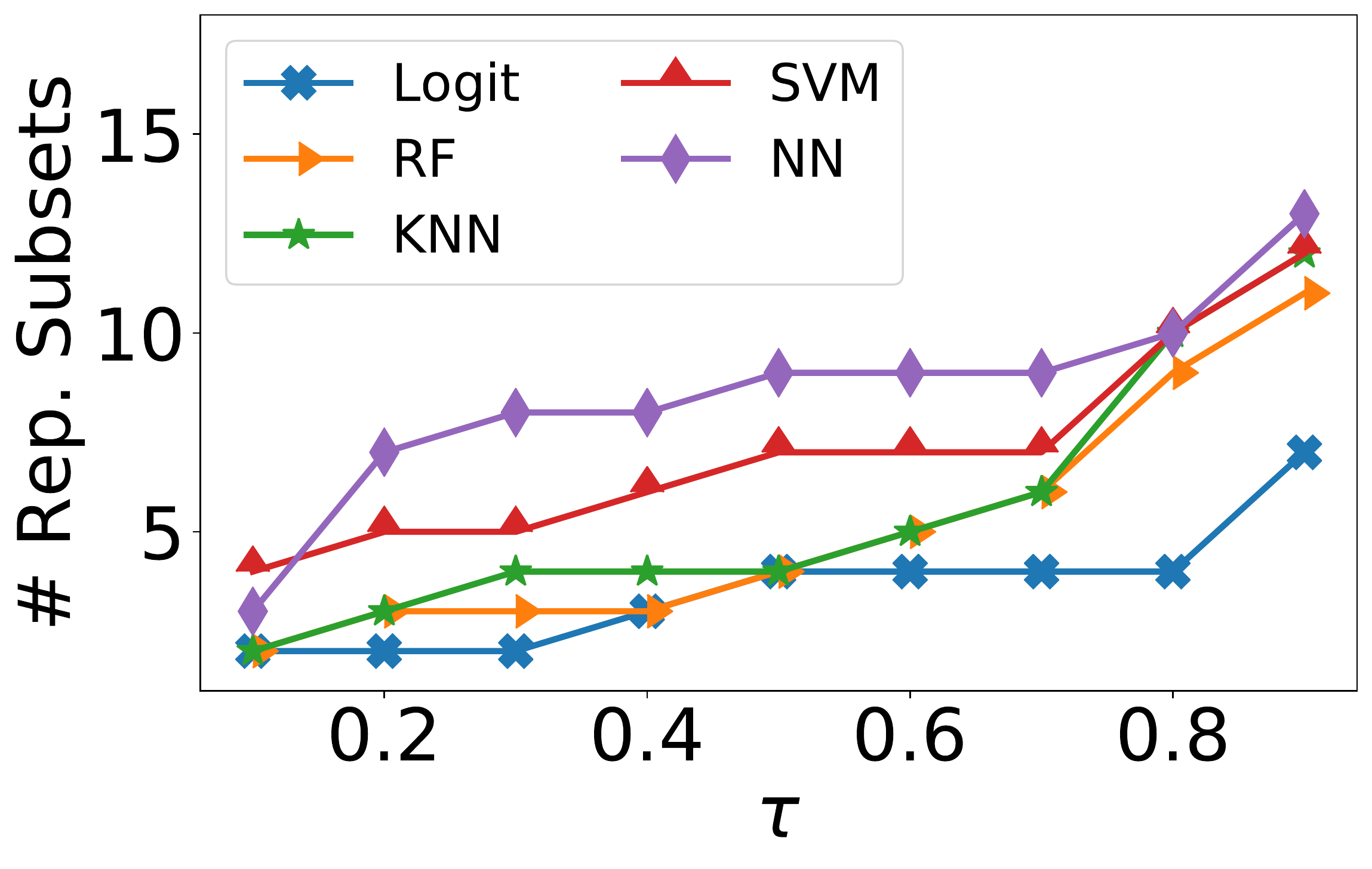}}

\caption{Number of representative subsets given a dataset using different models}
\label{fig:threshold}
\hspace{\fill}
% \vspace{-7mm}
\end{figure*}

\noindent{\bf Number of representative metrics:} 
Next, we evaluate the impact of the parameter $\tau$, used for identifying the representative metrics, on the number of representatives $\mathcal{R}_{\mathcal{F}}$. 
Figure \ref{fig:threshold} presents the results for various values of the threshold $\tau$ for each ML model for COMPAS, Adult, Credit, Bank datasets. 
The thresholds values are selected as $\tau \in \{0.1,0.2,\cdots,0.9\}$.
% are used to categorize the strength of the relationship between fairness metrics in the representative identification algorithm. 
% \hadis{we need to probably revise the non-linearity discussion.}
{Confirming Lemma~\ref{lem.3.2}, we observe that as $\tau$ increases the number of subsets increases.
Besides, as discussed earlier, the number of representatives are model-dependent, and it is relatively larger for {\bf NN}. %due to its complex boundaries and capability in identifying local patterns in the data.
} 
% For non-linear classifiers such as NN, the number of subsets is relatively larger due to the sensitivity of the non-linear decision boundaries to the subset of samples in the training set. 
In such a situation, the fairness metrics would be less correlated. In general, fairness metrics of linear decision boundaries are more correlated. Although a similar overall pattern can be observed from one dataset to another, the number of the subset of representatives is different. The results indicate that the proposed approach for the estimation of correlation is model-dependent.

\begin{figure*}[!tb]
\centering
% \vspace{-7mm}
\subfloat[Logit]{\includegraphics[width=0.5\textwidth]{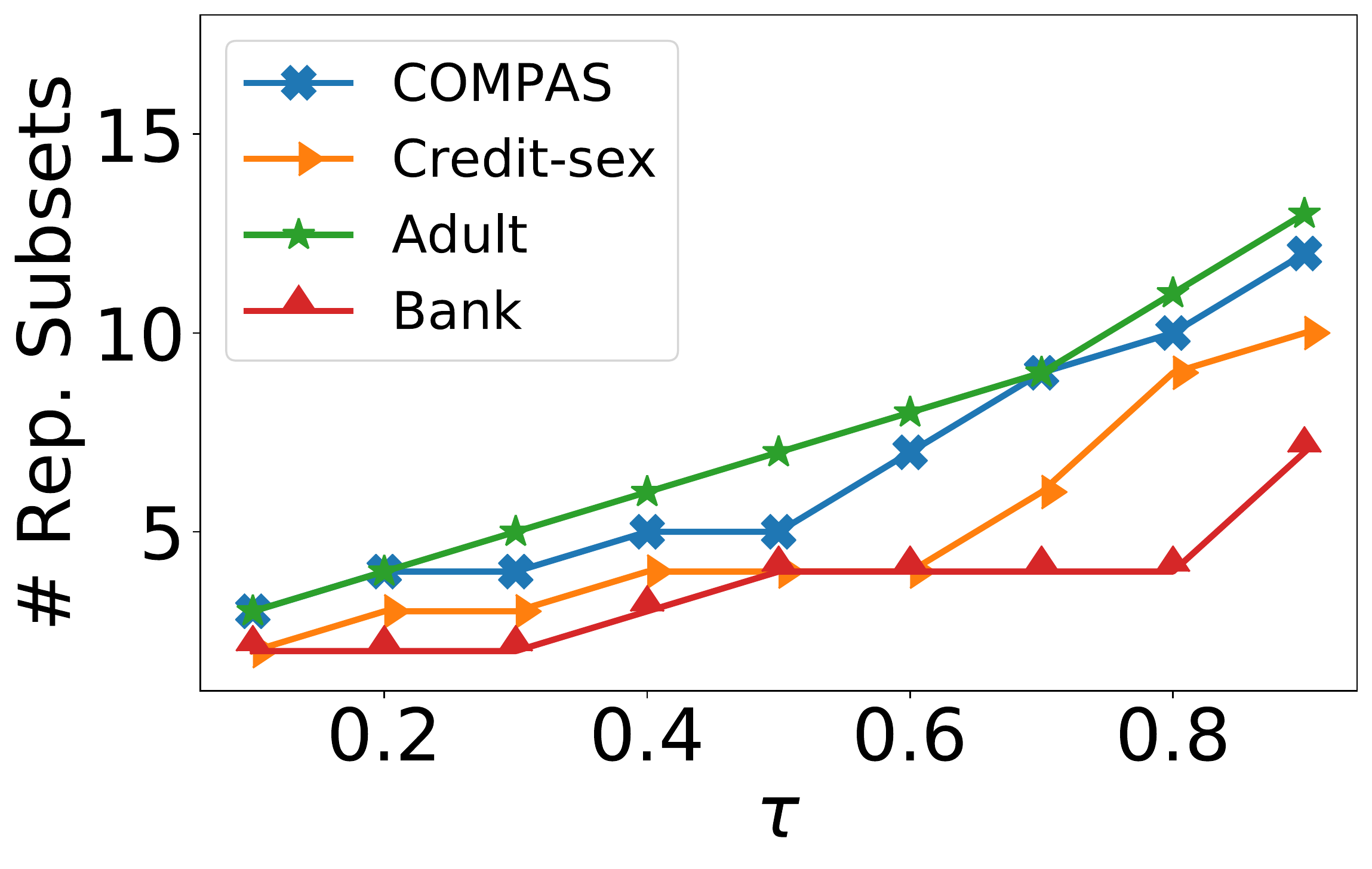}}
\subfloat[KNN]{\includegraphics[width=0.5\textwidth]{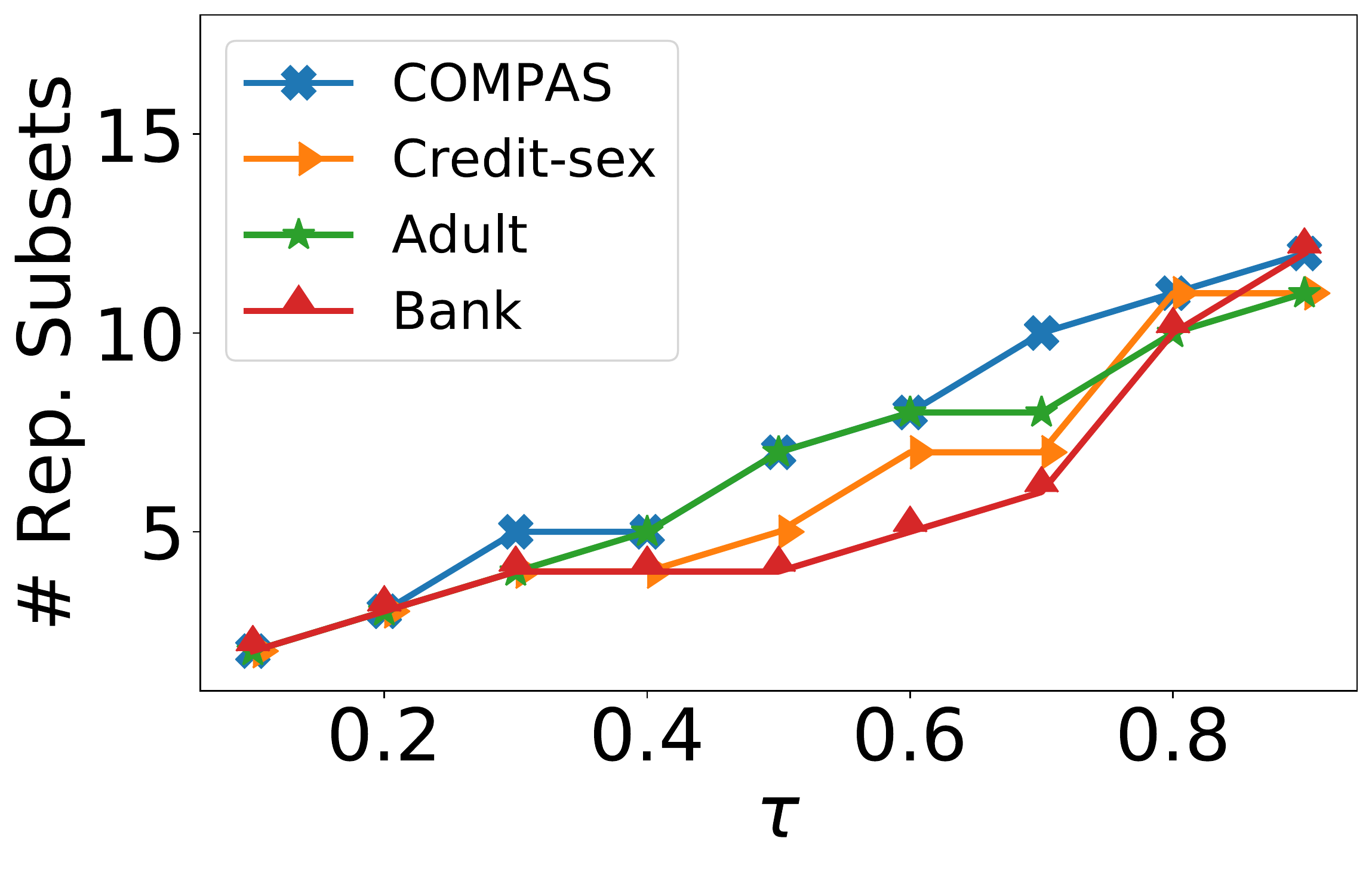}}

\subfloat[SVM]{\includegraphics[width=0.5\textwidth]{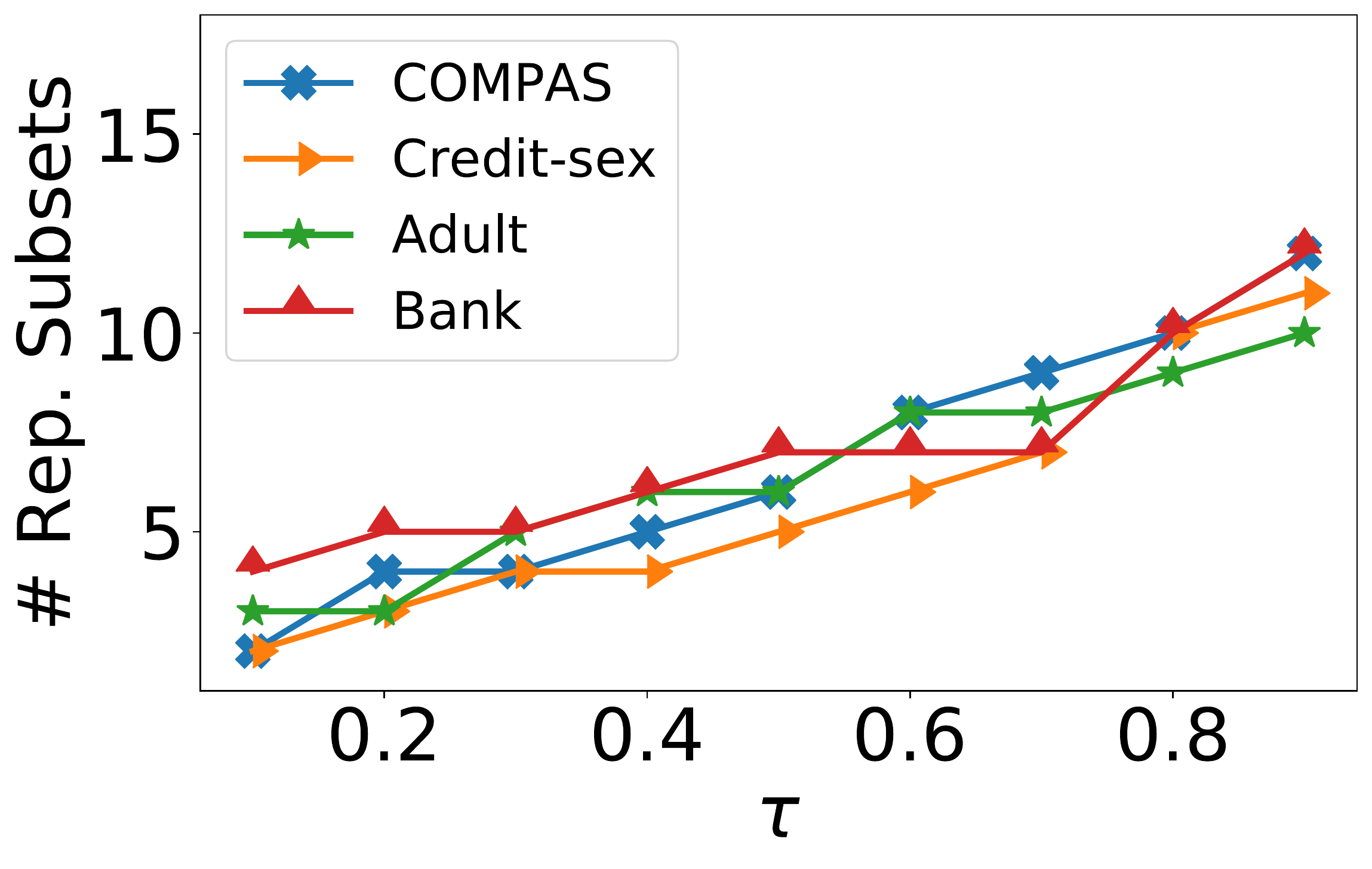}}
\subfloat[NN]{\includegraphics[width=0.5\textwidth]{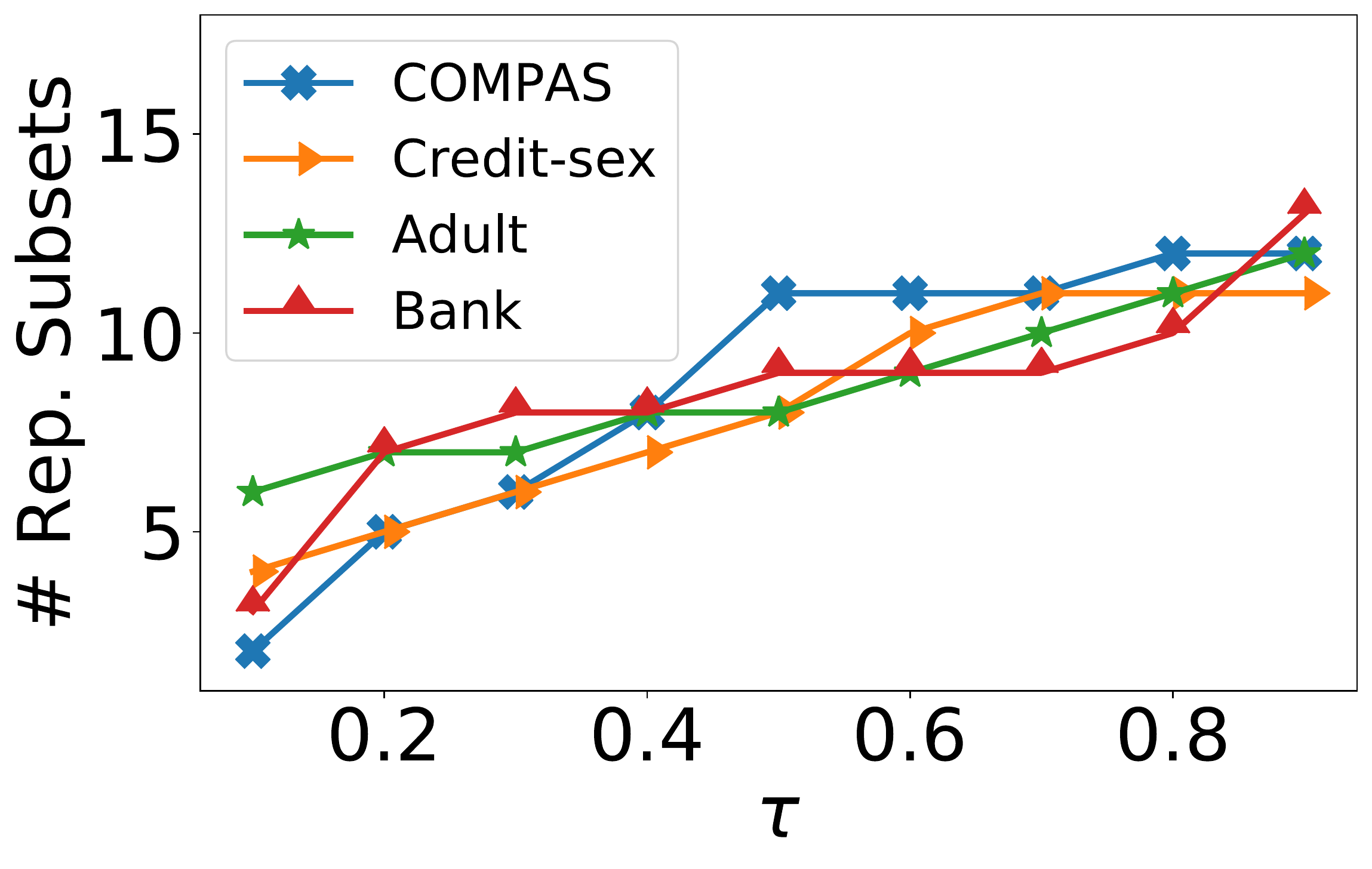}}
% \vspace{-4mm}
\caption{Number of representative subsets given a model using different dataset}
\label{fig:threshold-model}
\hspace{\fill}
% \vspace{-7mm}
\end{figure*}

In our next experiments, Figure \ref{fig:threshold-model} represents the number of representative subsets of fairness metrics, $\mathcal{R}_{\mathcal{F}}$, for different datasets fixing the ML model. We demonstrate that given a model as $\tau$ increases the size of $\mathcal{R}_{\mathcal{F}}$ increases. The nonlinear models, as expected, require more subsets. The results indicate that the proposed approach for the estimation of correlation is data-dependent. 

%\hadis{Nazanin revise this paragraph}

\begin{figure*}[!tb]
\centering
\subfloat[Credit-Logit]{\includegraphics[width=0.40\textwidth]{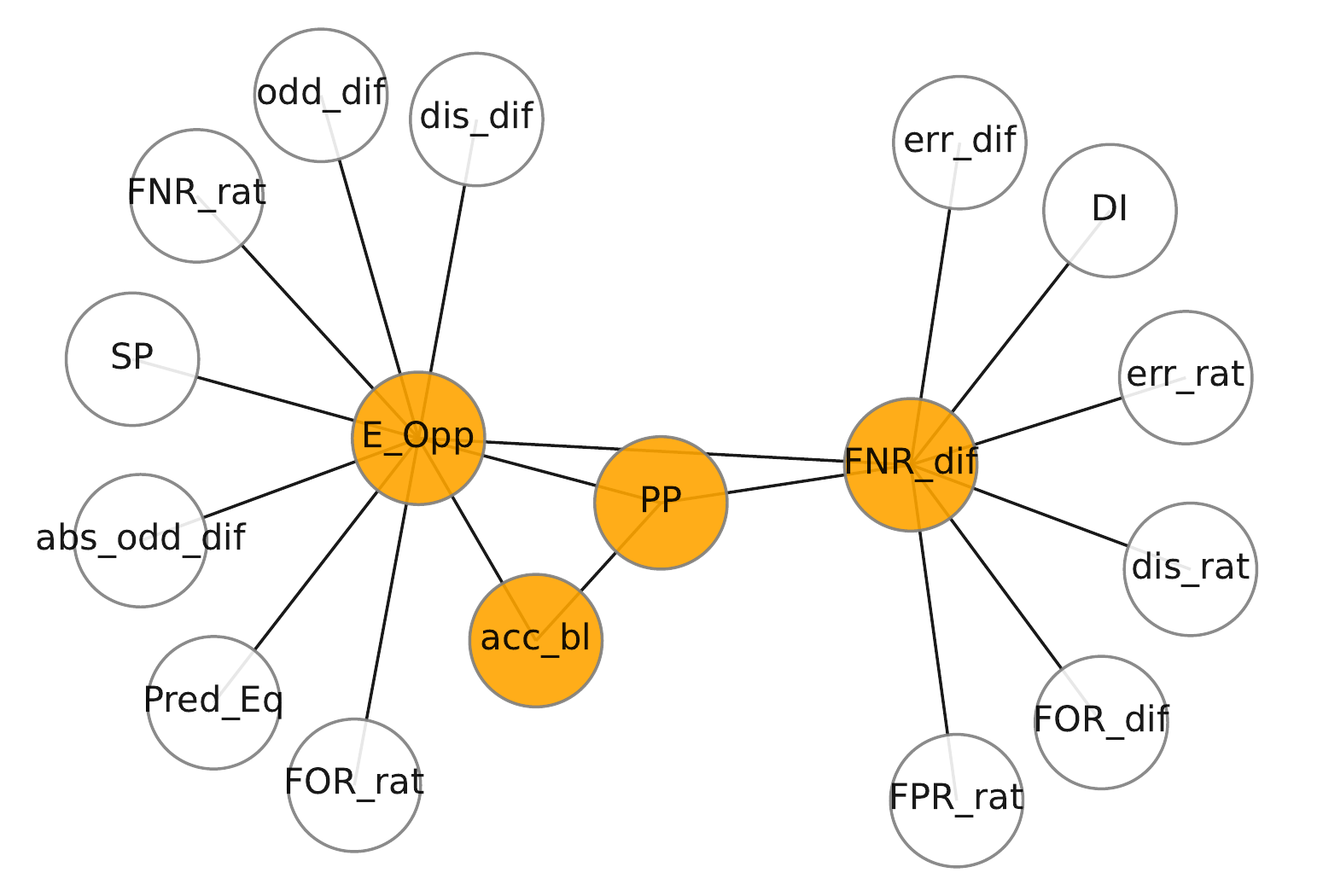}}
\subfloat[Correlation values]{\includegraphics[width=0.33\textwidth]{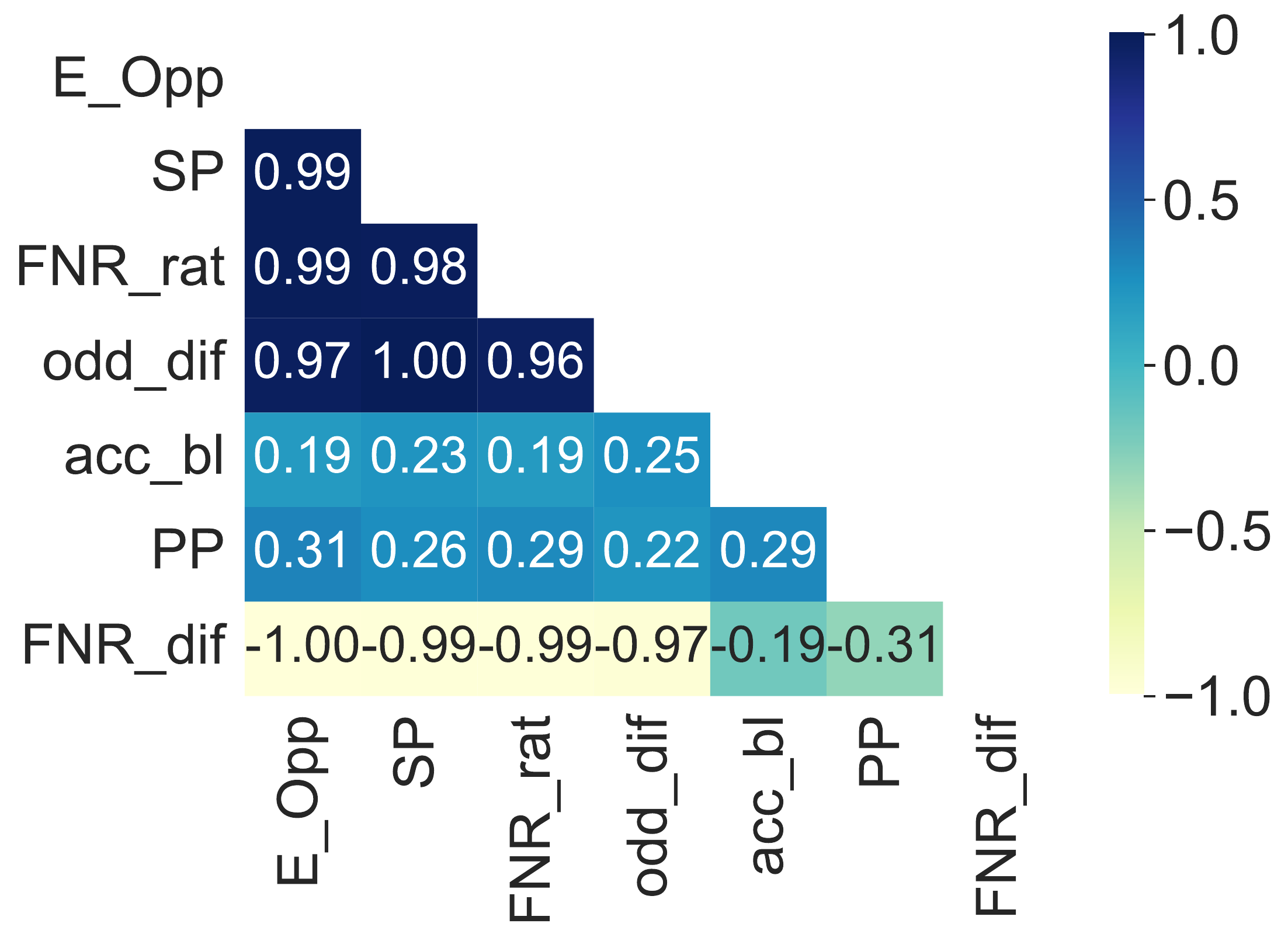}}
\subfloat[Unfairness Mitigation]{\includegraphics[width=0.31\textwidth]{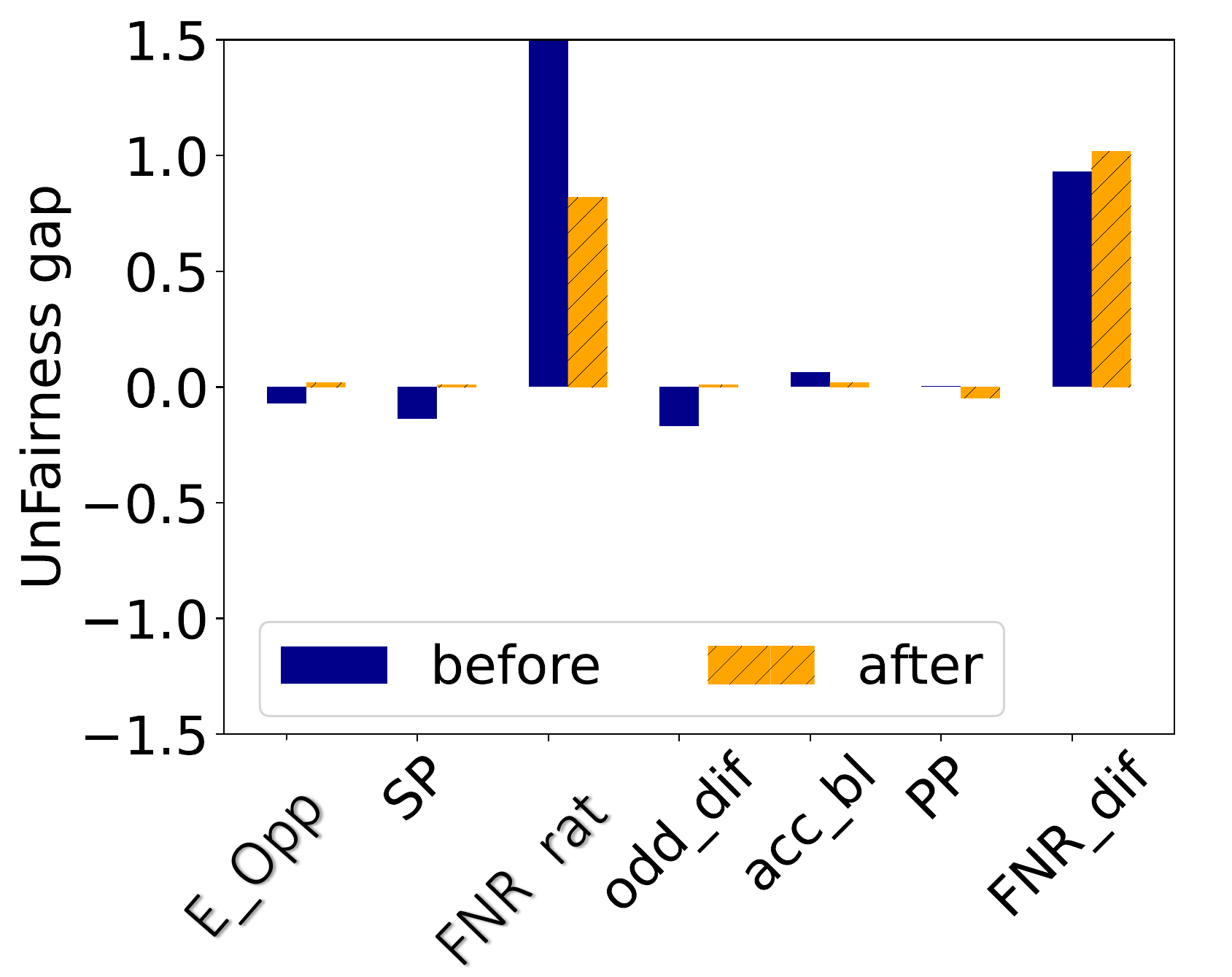}}
%Mitigate_German-sex-modelLogit.pdf

\subfloat[Credit-RF]{\includegraphics[width=0.40\textwidth]{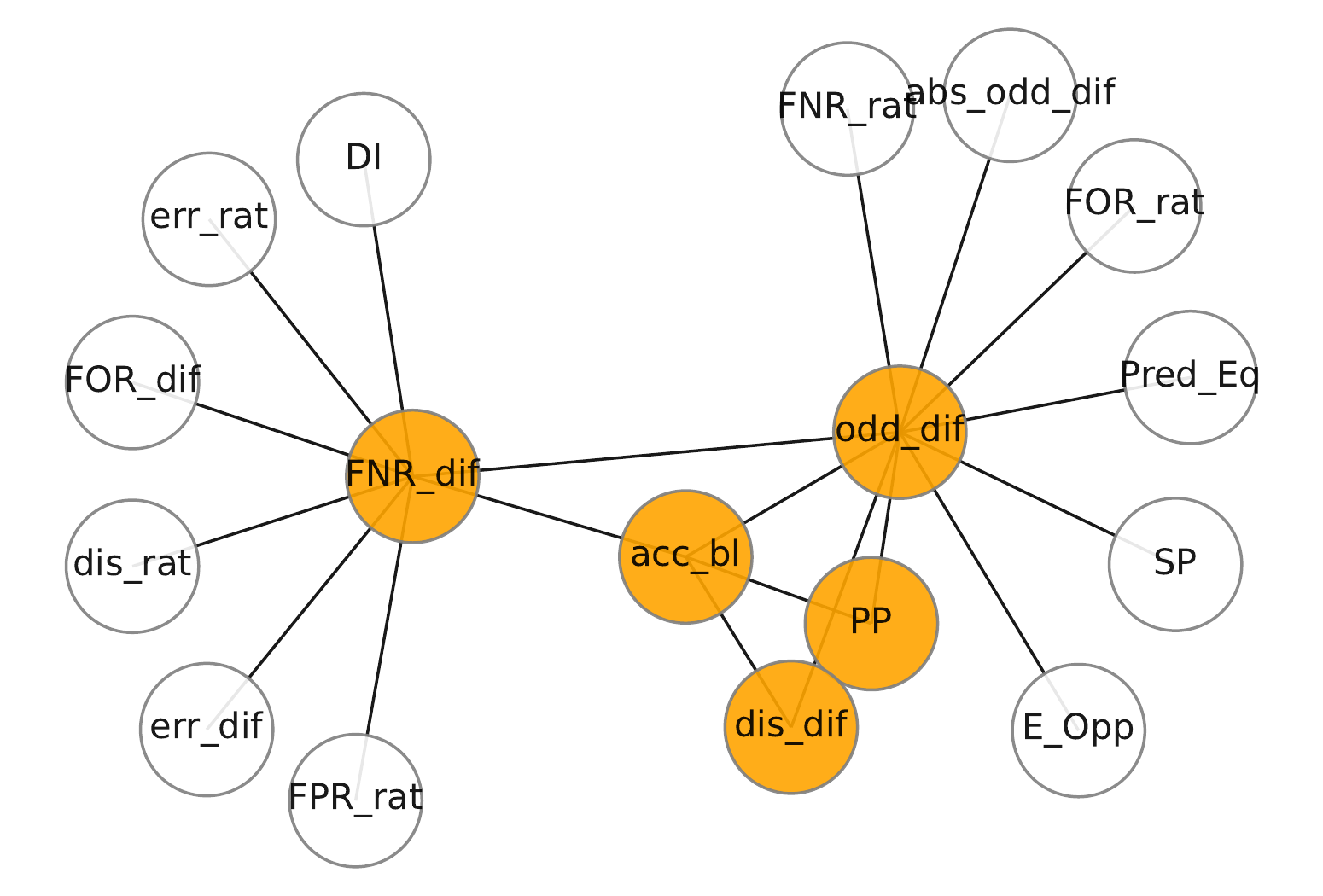}}
\subfloat[Correlation values]{\includegraphics[width=0.33\textwidth]{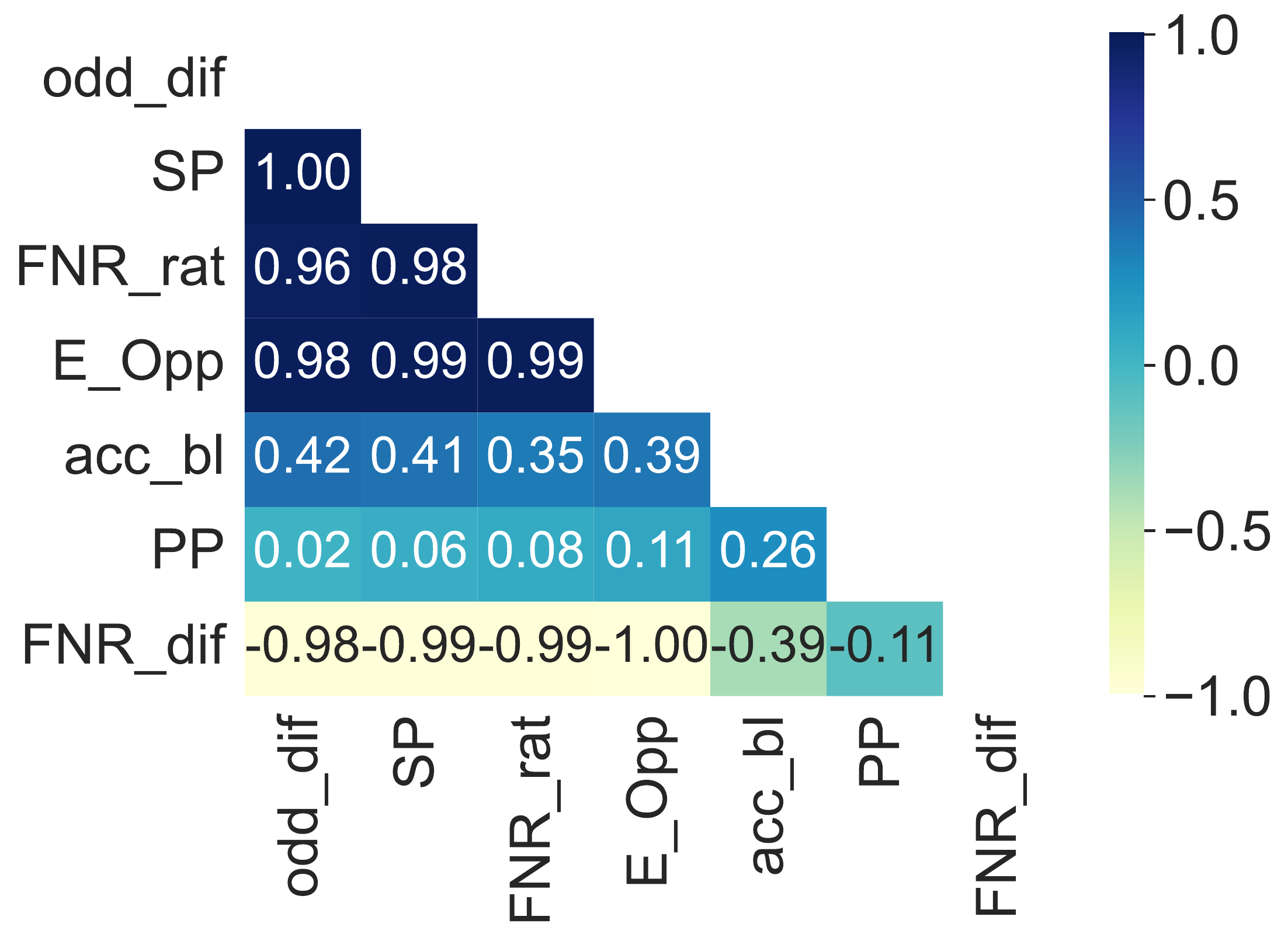}}
\subfloat[Unfairness Mitigation ]{\includegraphics[width=0.31\textwidth]{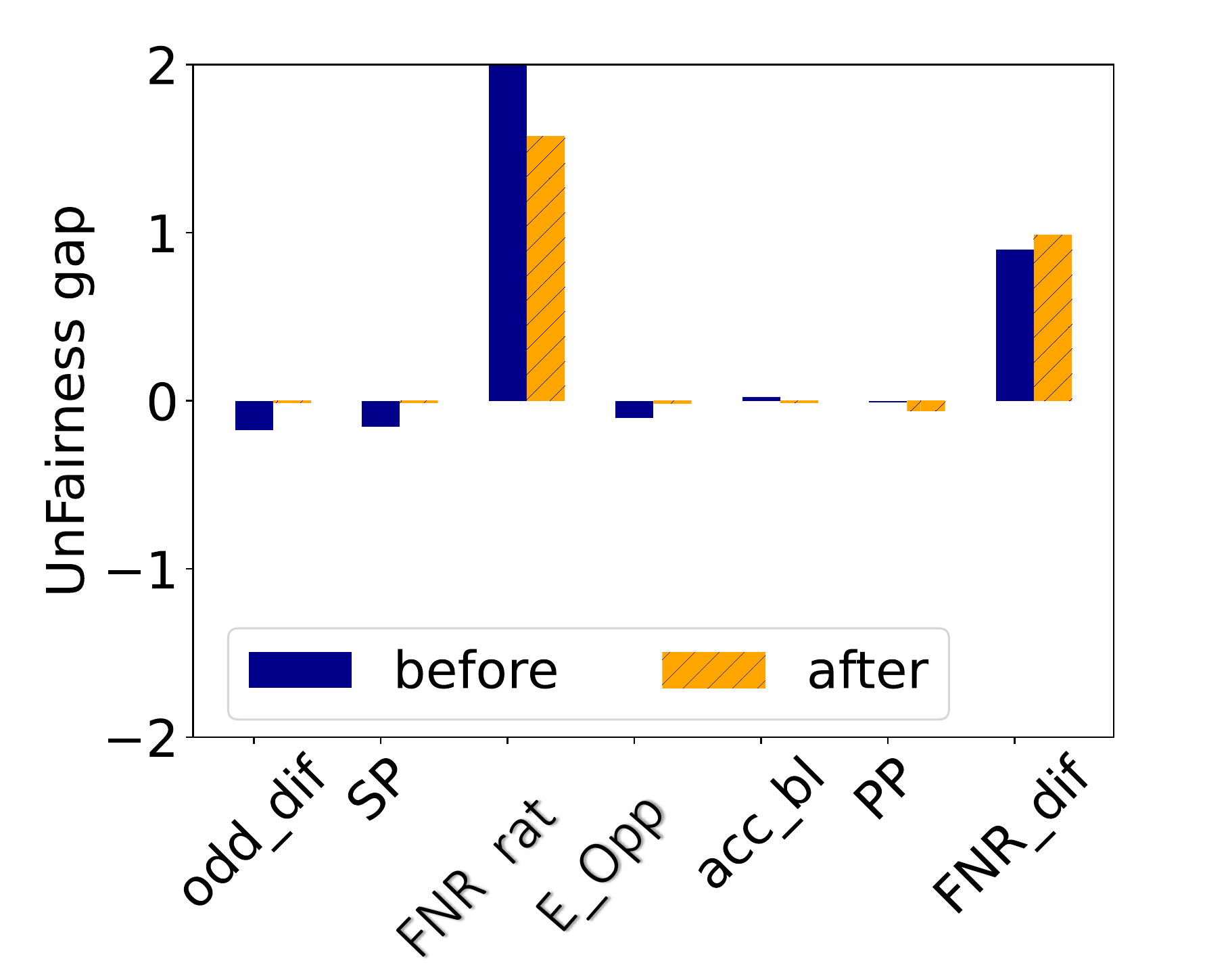}}
%Mitigate_German-sex-modelRF.pdf

\subfloat[COMPAS-RF]{\includegraphics[width=0.40\textwidth]{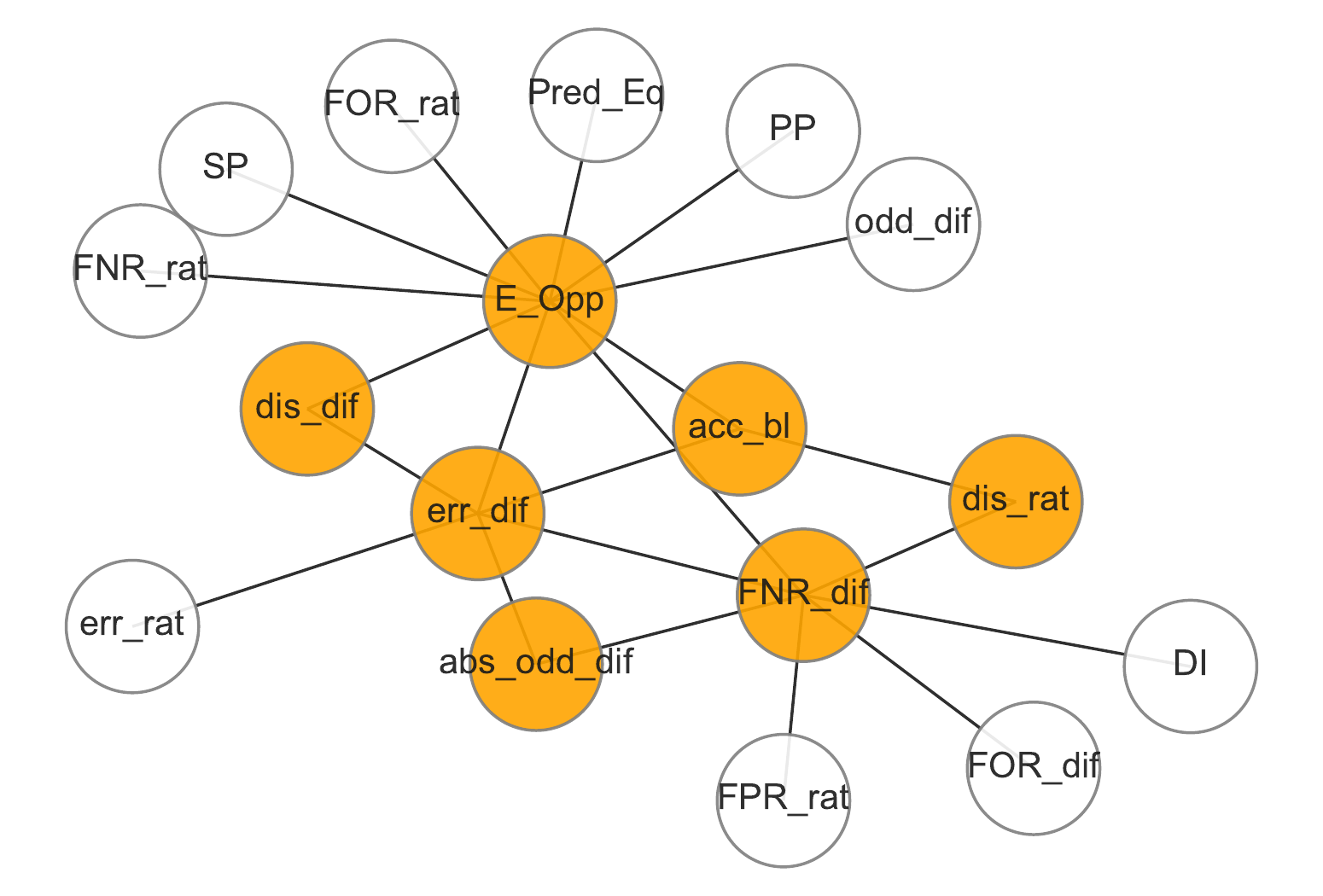}}
\subfloat[Correlation values]{\includegraphics[width=0.33\textwidth]{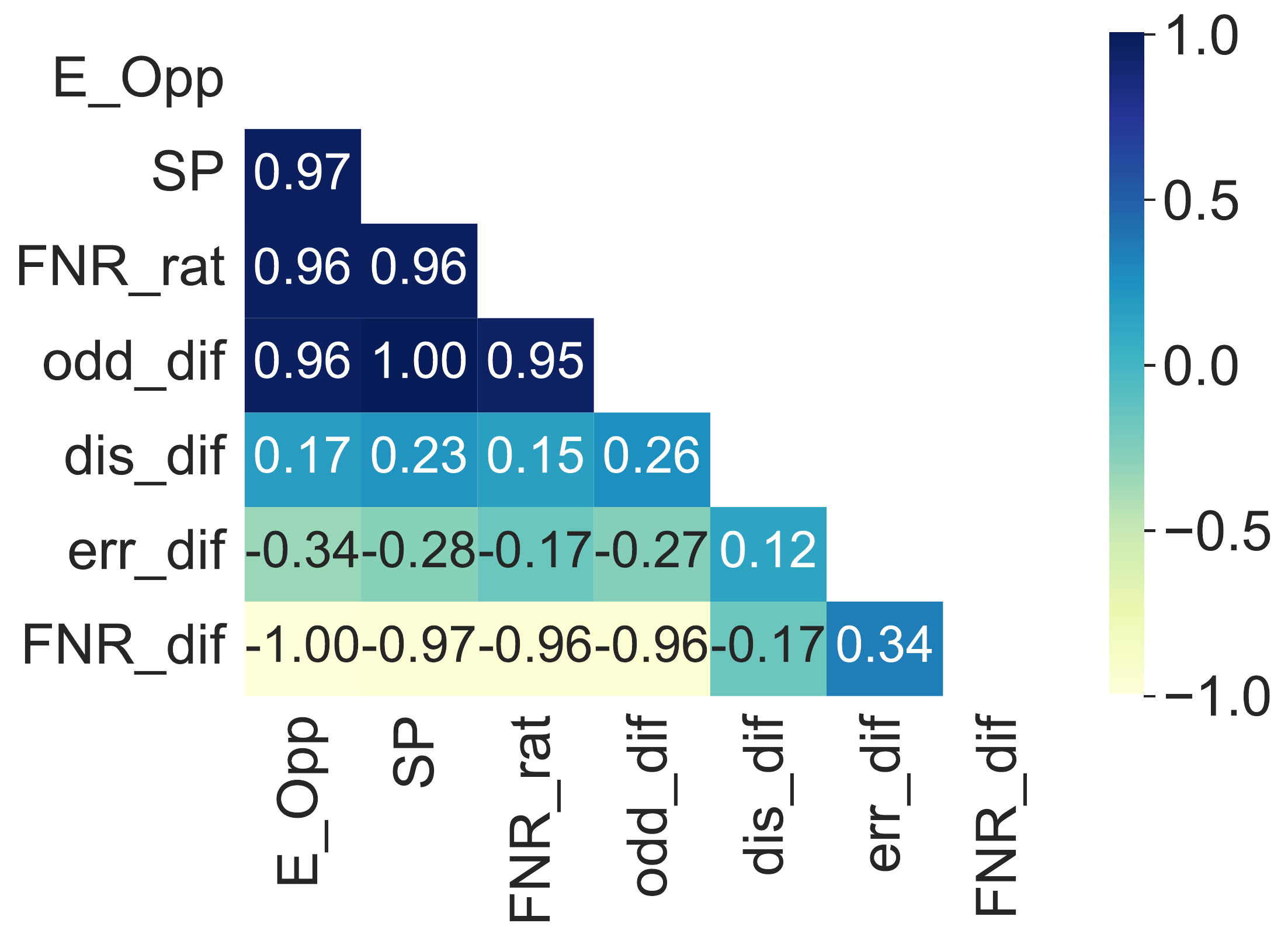}}
\subfloat[Unfairness Mitigation]{\includegraphics[width=0.31\textwidth]{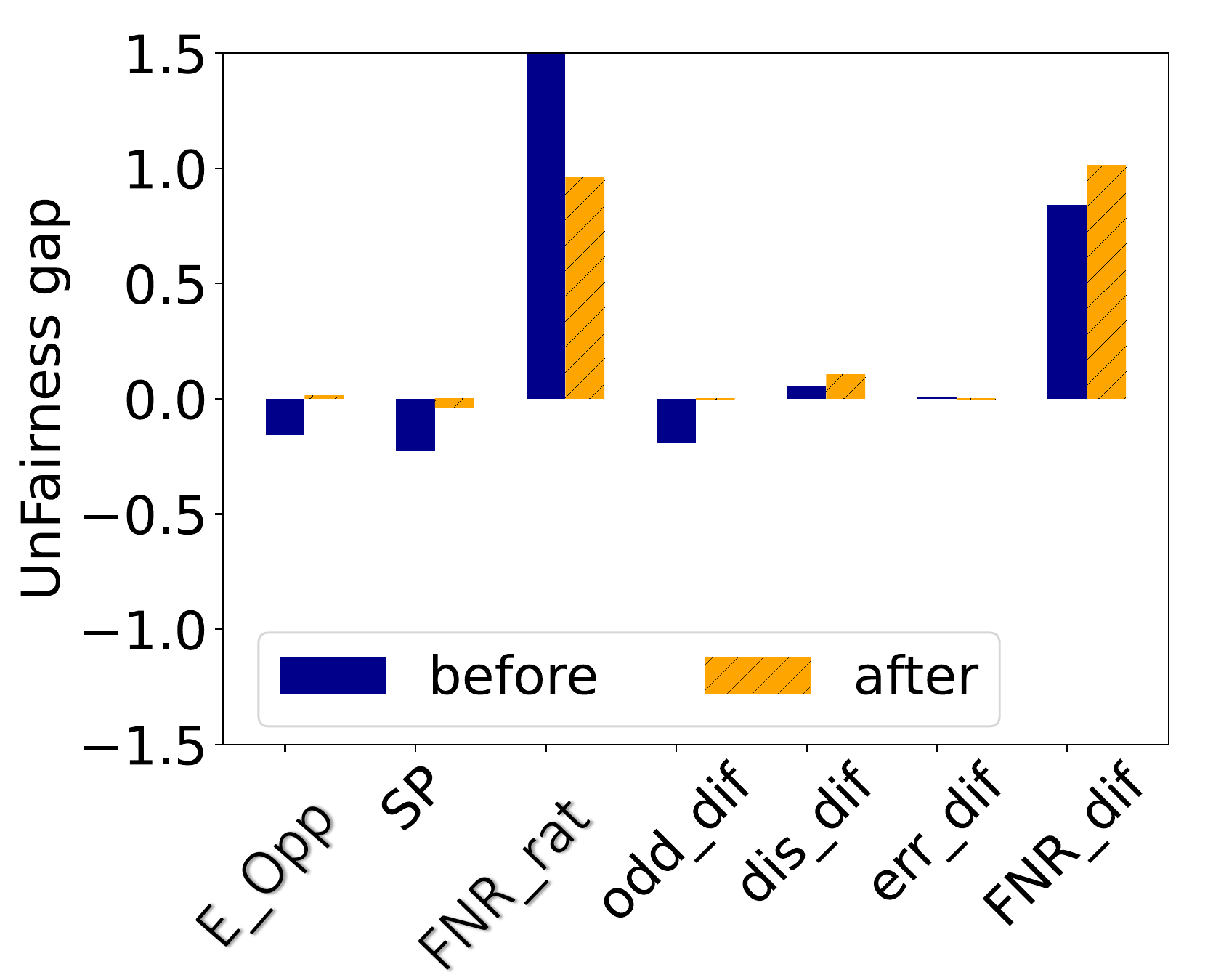}}
%Mitigate_COMPAS-race-modelRF.pdf

\subfloat[COMPAS-SVM]{\includegraphics[width=0.40\textwidth]{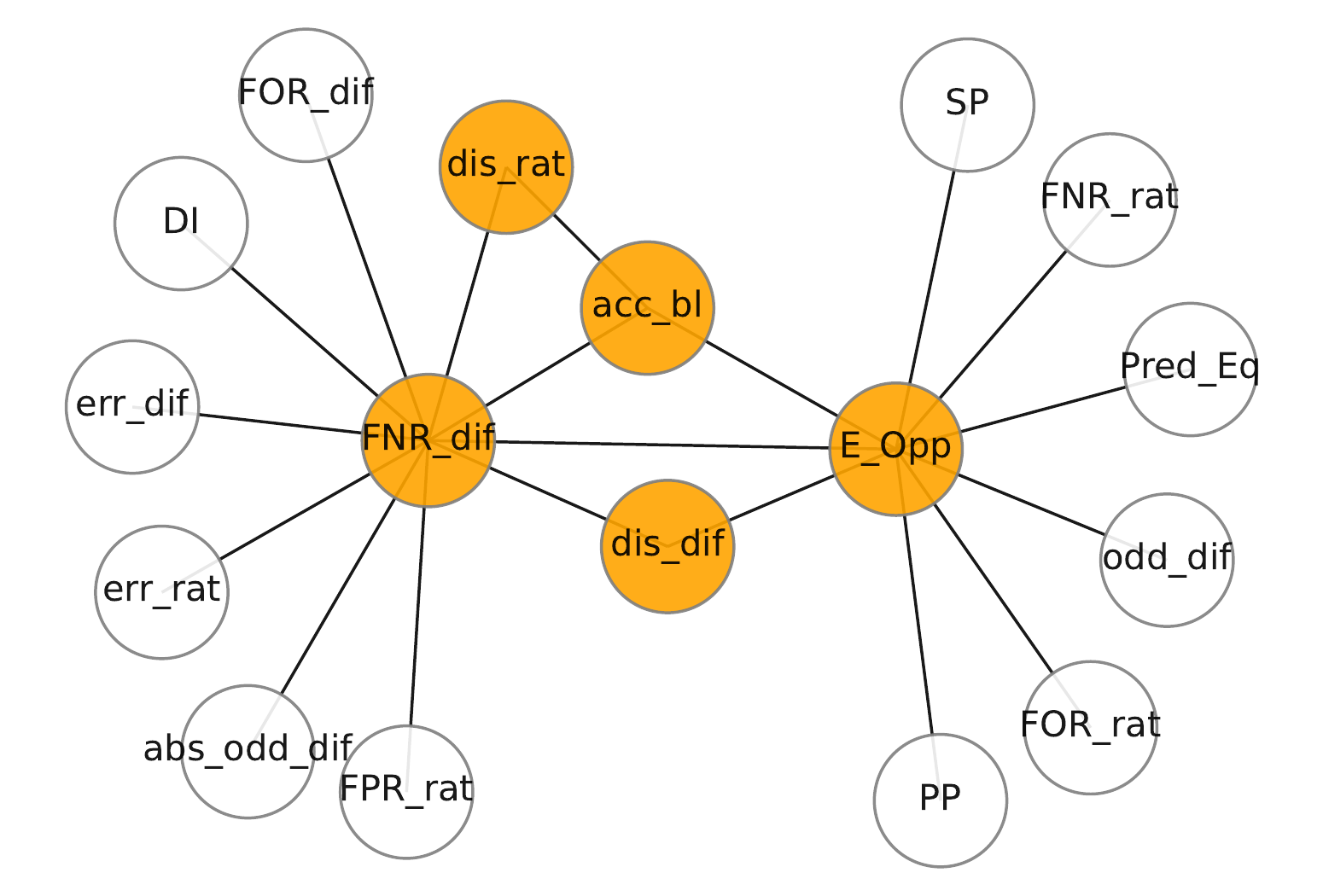}}
\subfloat[Correlation values]{\includegraphics[width=0.33\textwidth]{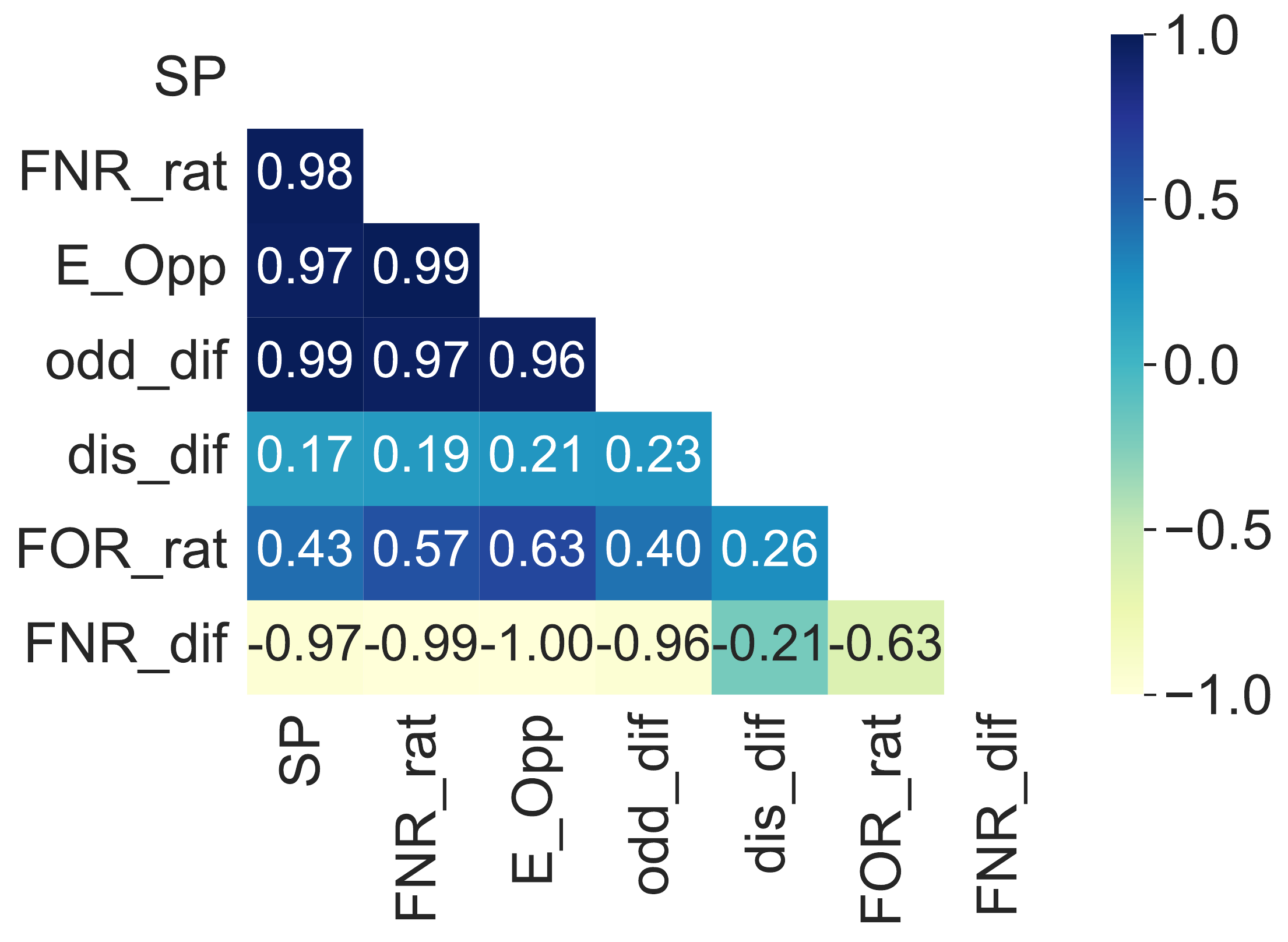}}
\subfloat[Unfairness Mitigation ]{\includegraphics[width=0.31\textwidth]{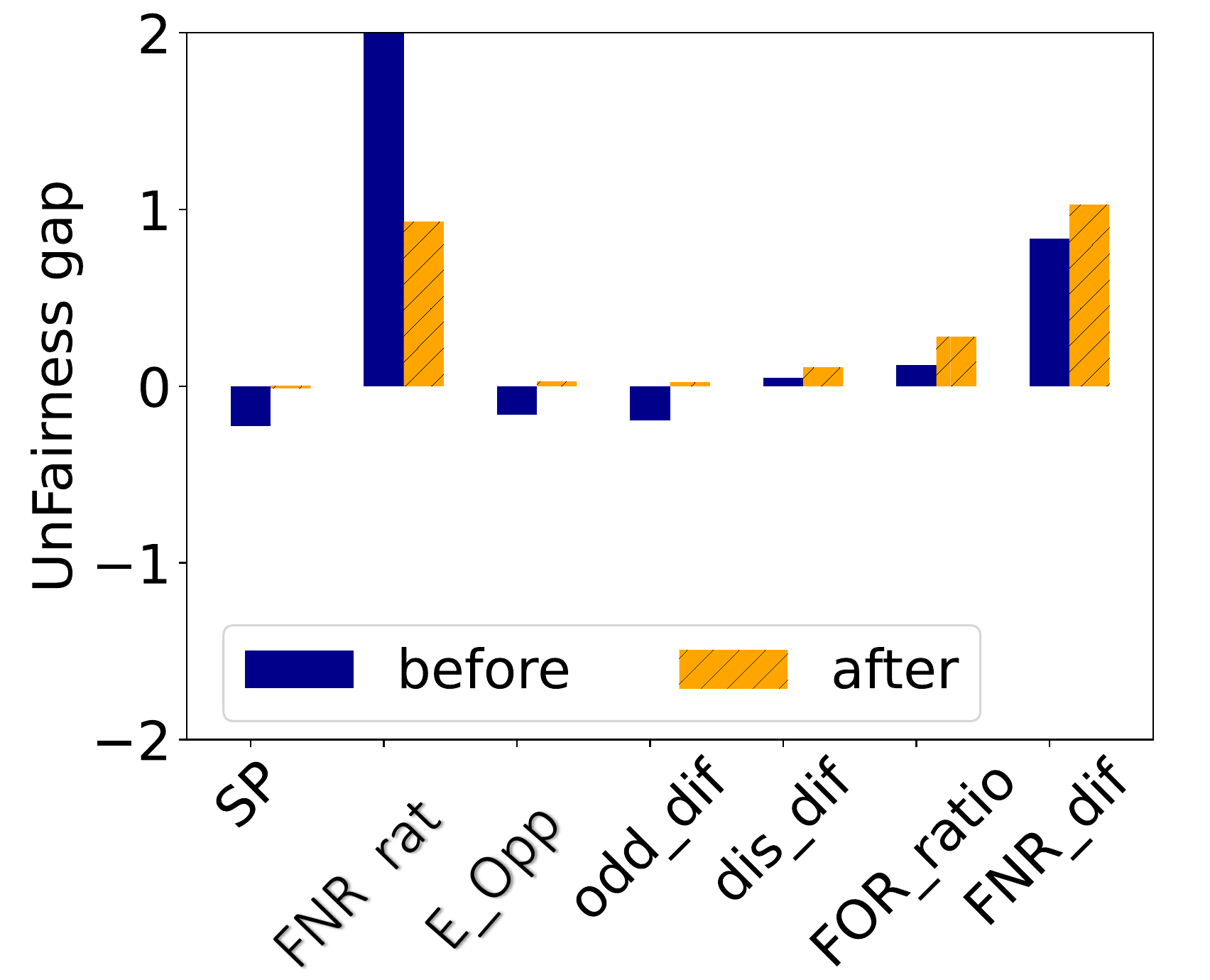}}
%Mitigate_COMPAS-race-modelSVM.pdf

\caption{Graph illustration of identified subset of fairness representatives, correlation heatmap, and bias mitigation (Exponentiated Gradient Reduction) impact on Intra-cluster and within-cluster metrics } 

\label{fig:graphs-cluster}
% \vspace{-5mm}
\end{figure*}

\neww{Figure \ref{fig:graphs-cluster} illustrates examples of clustering graphs of fairness metrics for COMPAS and Credit datasets. Each orange node indicates the representative metrics of each subset, and white nodes having an edge to the representative node show the subsets of highly correlated metrics. We used $\tau=0.5$ to generate the results of this plot. Comparison of Figures (a) and (d) in Figure~\ref{fig:graphs-cluster} confirms that the number of representatives obtained with a linear {\bf Logit} model is smaller compared with {\bf RF}, which is a non-linear model for Credit dataset. Similarly, comparing Figures (g) and (j) in Figure~\ref{fig:graphs-cluster} shows that using a linear model such as {\bf SVM} leads to a smaller number of representatives compared with {\bf RF} for the COMPAS dataset.}

%The graph confirms that the number of subsets is smaller using {\bf Logit} model than that using {\bf RF}, as previously discussed \hadis{we did not discuss RF previously we discussed NN and now we do not have NN results anymore}. Similarly, Figure \ref{fig:graphs-credit} shows the representative metrics for Credit dataset using the {\bf Logit} and {\bf NN} models. \hadis{check the figure labels}
%\emph{Discovery Ratio}, \emph{Predictive Parity}, \emph{Equality of Opportunity}, and \emph{Average Odd Difference} are selected as representative metrics for Credit dataset when we use {\bf Logit} classifier. 
% \hadis{you need to first compare logit and RF for credit and then say similarly for compas with RF and another linear model SVM with linear kernel.}}

% \hadis{computation time report and plot of subsets versus time similar to fig 3}

% \hadis{check figure references}

% \hadis{Nazanin add the discussion for fixed size other datasets for computation time}

\begin{figure*}[!tb]
\centering
\subfloat[COMPAS-NN]{\includegraphics[width=0.5\textwidth]{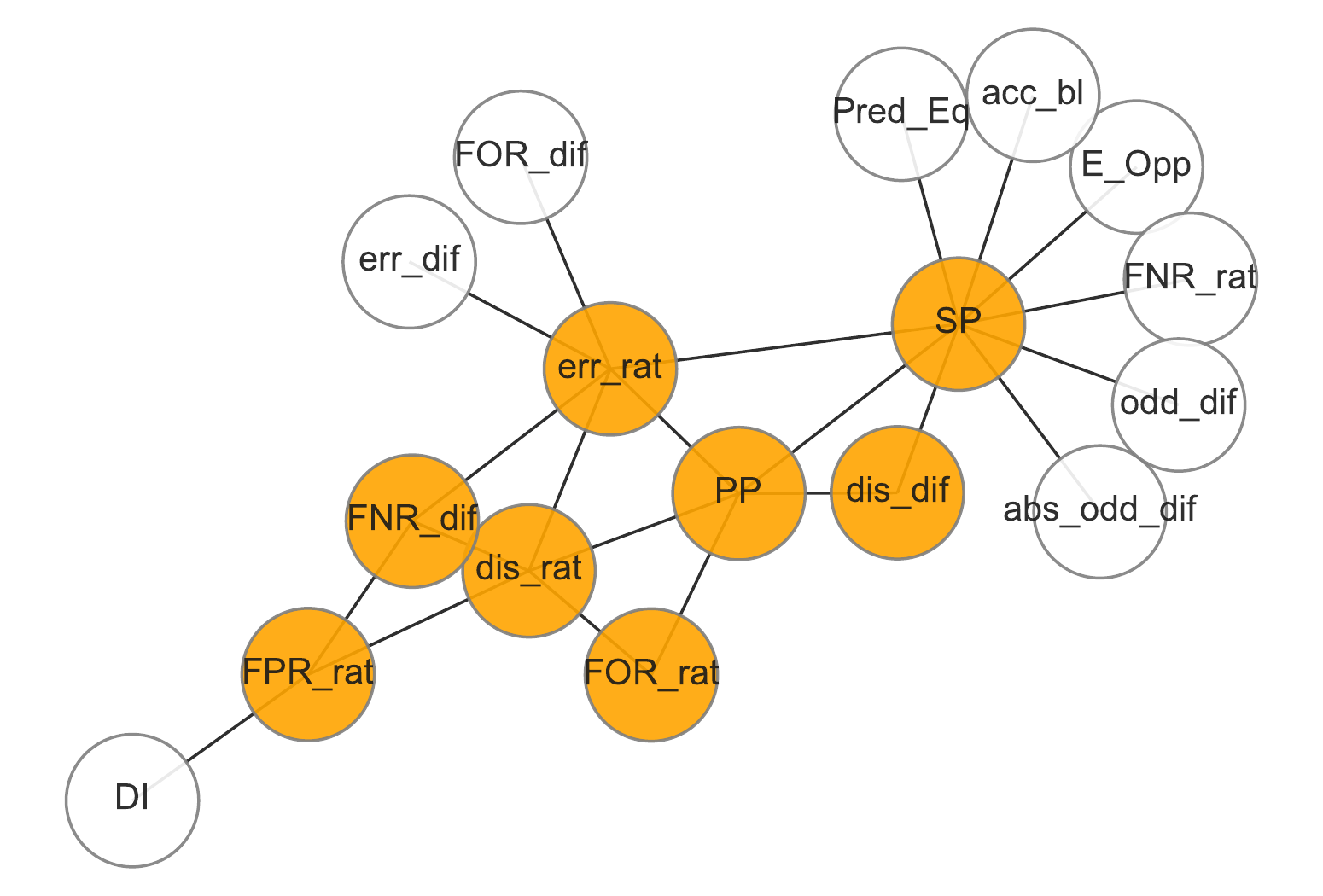}}
\subfloat[Adult-SVM]{\includegraphics[width=0.5\textwidth]{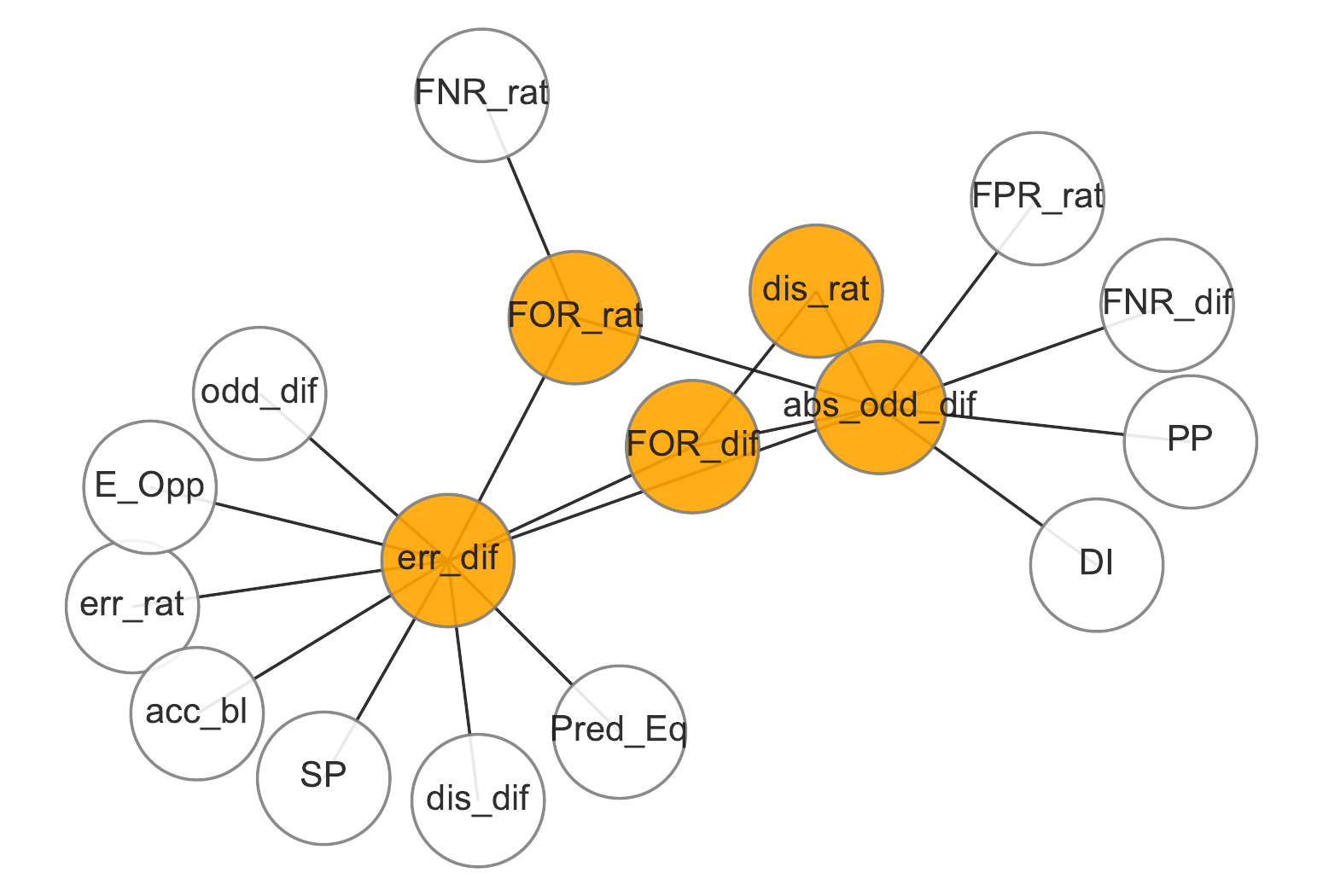}}

\subfloat[Bank-KNN]
{\includegraphics[width=0.5\textwidth]{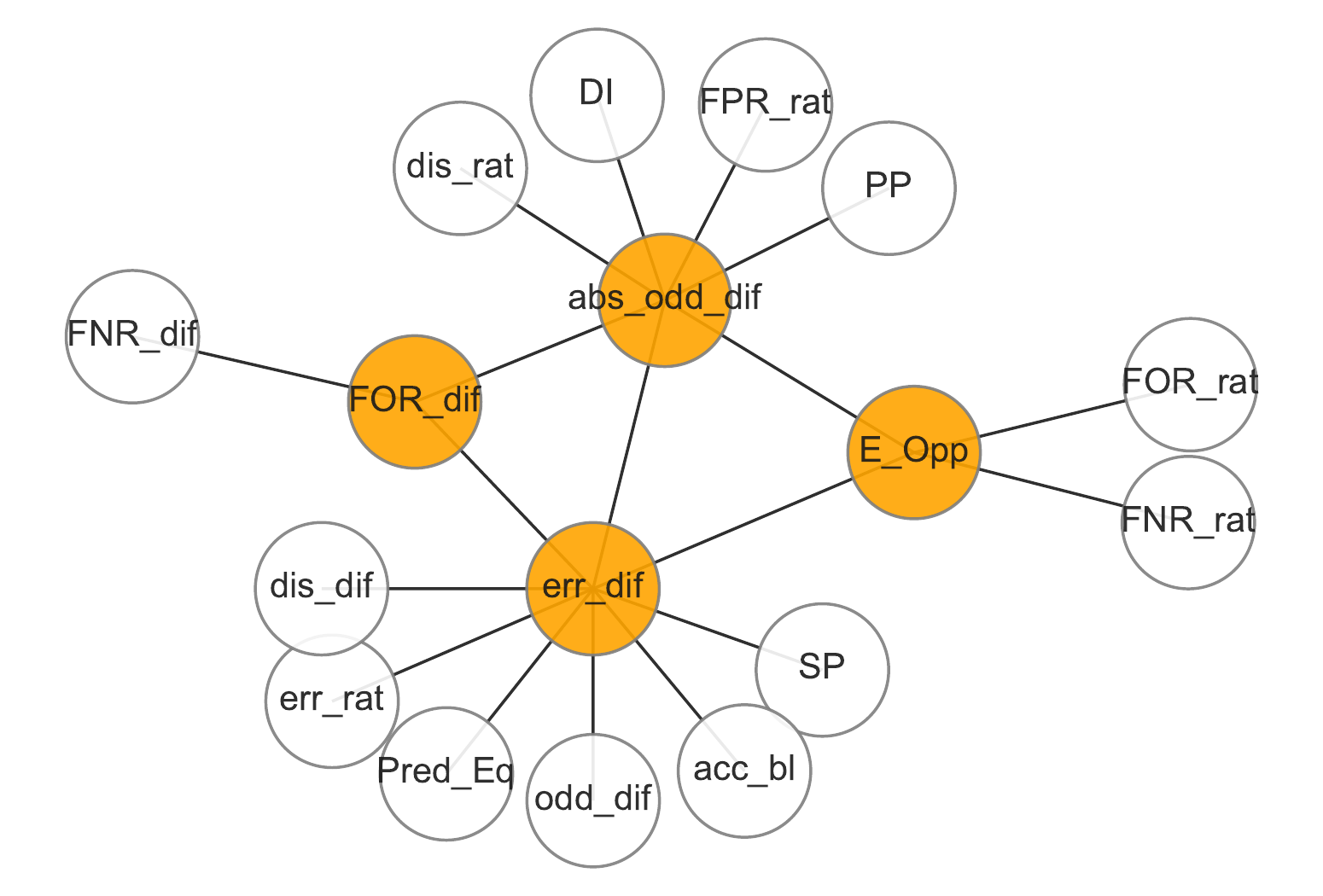}}
\caption{Graph illustration of identified subset of fairness representatives} 
\label{fig:other_graphs}
\end{figure*}

\neww{The primary goal of our proposed framework is to guide practitioners in performing the mitigation task while being mindful of the correlations between fairness metrics and, consequently, the implications of mitigating one on others. 
Figures~\ref{fig:graphs-cluster} (c, f, i, l) present the impact of unfairness mitigation on both within and between cluster fairness metrics based on our estimated correlation values. We use Exponentionated Gradient Reduction~\citep{agarwal2018reductions,aif360} as an in-processing approach to mitigate the unfairness of one given representative measure (orange nodes) in different problems and models. We analyzed the impact of mitigation on the highly correlated metrics (within-cluster metrics) and the ones that are either orthogonal or negatively correlated (between-cluster metrics, i.e., other representatives). For example, mitigating unfairness based on the Equal opportunity metric (E-Opp) and Equalized odds difference (odd-dif) in the Credit dataset, shown in Figure~\ref{fig:graphs-cluster} (a-c) and (d-f), respectively, results in diminishing unfairness across highly correlated metrics such as Statistical Parity (SP), False negative rate ratio (FNR-rat). However, since other representative metrics are orthogonal or negatively correlated with the considered representative, unfairness mitigation may impact them adversely.}
%\hadis{make an example which notion gets worse}. 
\neww{For example, Predictive Parity (PP) and Equalized odd difference (odd-dif) are not correlated (correlation value of 0.02) based on Figure~\ref{fig:graphs-cluster} (d-f), hence, mitigating unfairness based on odd-dif does adversely impact the unfairness for PP.}
In particular, according to the correlation heatmap values, Figure~\ref{fig:graphs-cluster} (b), if two measures have a strong negative correlation, such as False negative rate difference (FNR-dif) and Equal opportunity, performing mitigation task on Equal opportunity will exacerbate the resulting unfairness for the other one (e.g., FNR-dif). Similarly, Figure~\ref{fig:graphs-cluster} (g-i) and (j-l) indicate bias mitigation results based on the Equal opportunity metric (E-Opp) and Statistical Parity (SP) for the COMPAS dataset. The results are consistent across different models and datasets, indicating that reducing the unfairness for the representative metric induces unfairness reduction among other highly correlated metrics while adversely impacting the unfairness result for the representatives, which have a strong negative correlation. Figure~\ref{fig:other_graphs} shows some other clustering graph results obtained from applying the proposed framework on other datasets using different models.

\begin{figure}[!tb]
\centering
\includegraphics[width=0.4\textwidth]{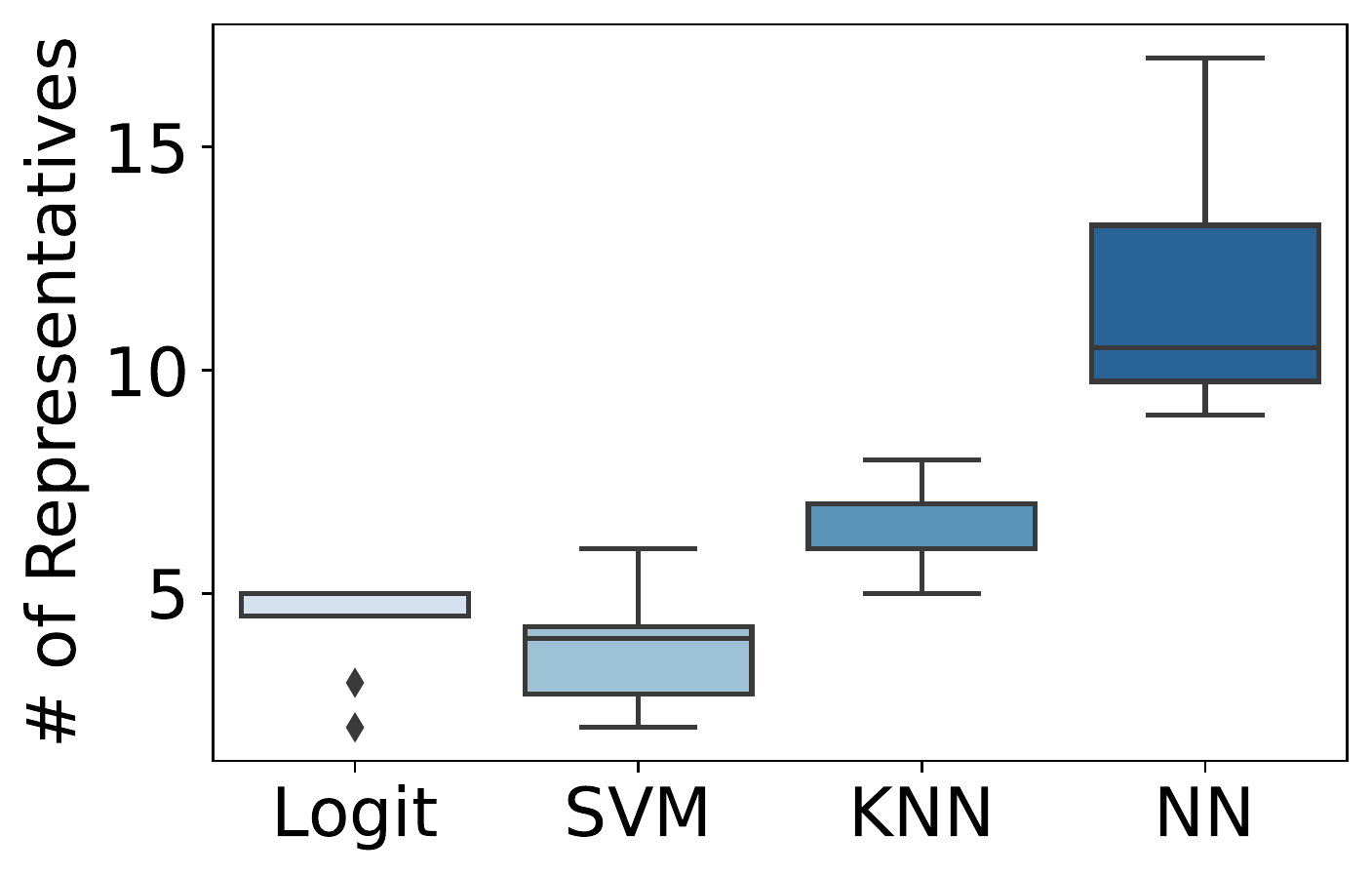}
\caption{Number of identified fairness representatives by different models}\label{fig:Nrep}
\end{figure}

\neww{In Figure~\ref{fig:Nrep}, we studied the impact of the complexity hyperparameters of different ML models on the number of clusters obtained from each model considering the COMPAS and Credit dataset. We performed our analysis on each model under different settings (datasets and hyperparameters). \neww{We respectively modified the $k$, number of neighbors, for {\bf KNN}, the kernel type and penalty term for {\bf SVM}, regularization hyperparamter for {\bf Logit}, and the number of hidden layers and nodes for {\bf NN}.} 
% \hadis{explain what hyperparameters and levels you used for each model} 
Our results indicate that more complex models such as {\bf NN} with several hidden layers and nodes tend to result in a higher number of cluster representatives on average compared with their linear counterparts (e.g., {\bf Logit}).}%\hadis{this  last sentence does not make sense}}
%as {\bf KNN} with small $k$ and
% KNN: $K \in \{1,10,20,50\}$
% Logit: $C \in {0.01,0.1,1,10}$
% SVM: $Kernel:{rbf,linear}, C={1,10}$
% NN: layers and nodes 
%\hadis{in svm what is the kernel? linear?}

\begin{figure}[!tb]
\centering
\includegraphics[width=0.4\textwidth]{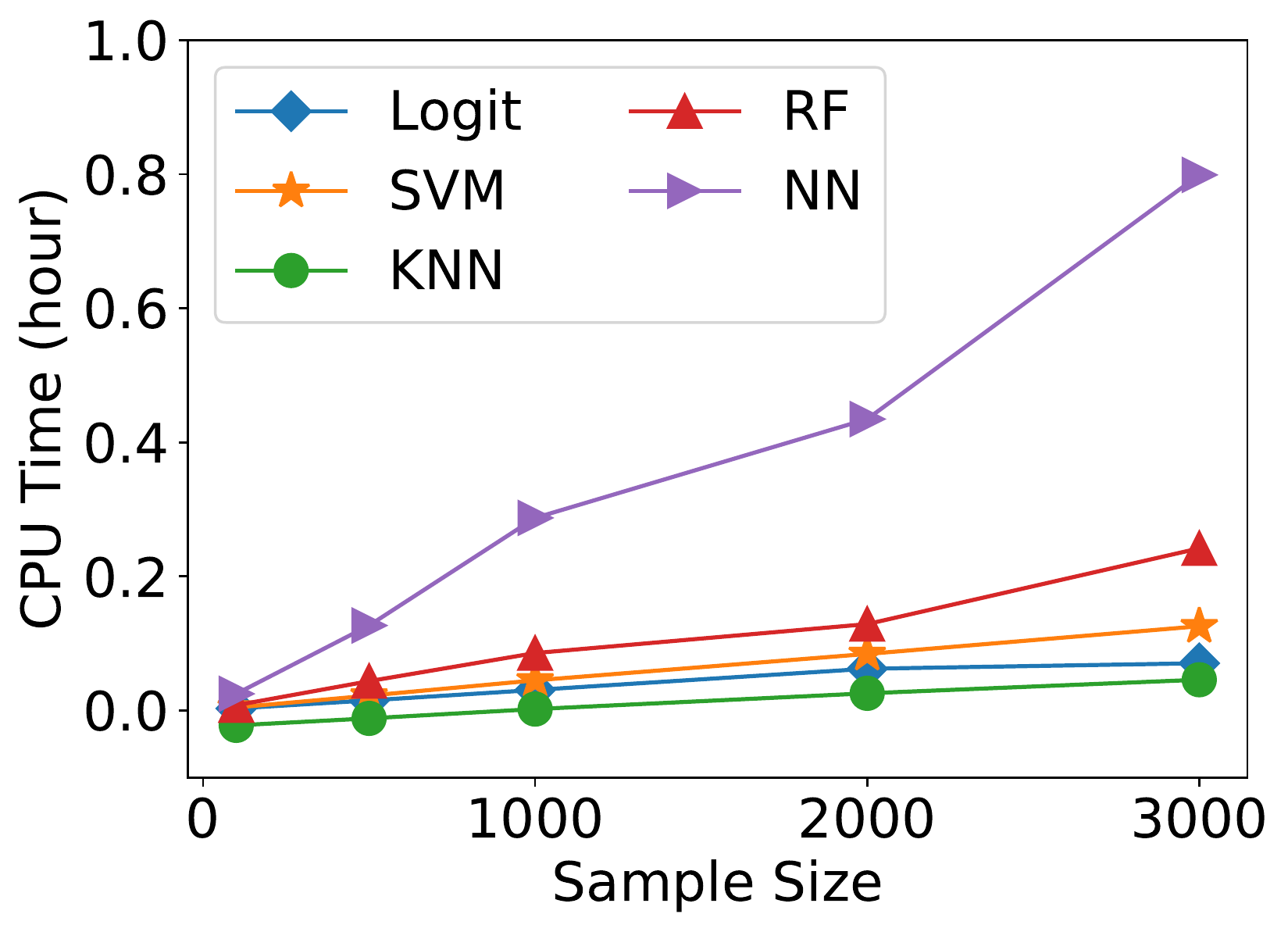}
\caption{End-end time of proposed approach across different models and sample size on COMPAS data set}\label{fig:time}
\end{figure}

Figure~\ref{fig:time} shows the end-end time for our proposed framework, including data sub-sampling, model training, and fairness evaluation steps on the COMPAS dataset with about 5875 observations. The computation time is proportional to the number of constructed sample models (sample size) for correlation estimation (sample size=1000 in our experiments). Note that using a fixed sample size for sub-sample the training data, our framework is scalable to any larger dataset and will have a similar CPU running time ($<1$ hr).


\section{Related work}\label{sec:related}
%\textbf{Fairness}
Algorithmic fairness has been studied extensively in recent years \citep{corbett2017algorithmic,kleinberg2018algorithmic}. Various fairness metrics have been defined in the literature to address the inequalities of algorithmic decision-making from different perspectives. \cite{barocas2017fairness} and \cite{verma2018fairness} define different fairness notions in detail.
The majority of works focus on the fairness consideration in different stages of predictive modeling including pre-processing \citep{feldman2015certifying,kamiran2012data,calmon2017optimized}, in-processing \citep{calders2010three,zafar2015fairness,asudeh2019designing}, and post-processing \citep{pleiss2017fairness,feldman2015certifying,stoyanovich2018online,hardt2016equality} to mitigate the outcome bias. Furthermore, the proposed interventions are tied to a specific fairness notion; statistical parity \citep{calders2010three}, equality of opportunity \citep{hardt2016equality}, disparate impact \citep{feldman2015certifying}, etc. 

A few recent works discuss the challenge of choosing the appropriate fairness metric for bias mitigation considerations.
\cite{makhlouf2021applicability} surveys notions of fairness and discusses the subjectivity of different notions for a set of real-world scenarios. 
% \cite{fazelpour2020algorithmic}
% demonstrate a connection between the recent literature on fair machine learning and the ideal approach in political
% philosophy, and show that some recently uncovered shortcomings
% in proposed algorithms reflect broader troubles faced by the ideal
% approach.
% show this analysis through for different formulations of fairness and conclude with a critical discussion of real-world
% impacts and directions for new research
The challenges about a growing pool of fairness metrics for unfairness mitigation and some aspects of the relationships between fairness metrics are highlighted in \cite{castelnovo2021zoo} with
respect to the distinctions individual vs. group and observational vs. causality-based. As a result, the authors highly promote quantitative research for fairness metric assessment for bias mitigation. Building on previous works \citep{kleinberg2016inherent,chouldechova2017fair}, \cite{garg2020fairness} provides a comparative using mathematical representations to discuss the trade-off between some of the common notions of fairness. \neww{In \cite{friedler2019comparative} Friedler et al. studied the effectiveness of fairness mitigation strategies proposed in the literature and showed the correlation between fairness metrics. To the best of our knowledge, none of the existing work proposed an automated framework to identify a representative subset of fairness metrics given a dataset and a model type. We addressed this gap in the literature and proposed an effective framework to discover the representative fairness metrics and facilitate the fairness metric selection and unfairness mitigation process for data scientists.} 

% \hadis{better positioning with respect to the existing work: describe 11 26 18 precisely}
% The precise differences, implications and “orthogonality” between these notions
% have not yet been fully analyzed in the literature. In this work, we try to make some order out of this zoo of definitions.

% \cite{holstein2019improving}
% identify areas of alignment and disconnect between the challenges faced by teams in practice and the
% solutions proposed in the fair ML research literature. Based
% on these findings, they highlight directions for future ML and
% HCI research that will better address practitioners’ needs.

\section{Final Remarks}\label{sec:con}
The abundance, trade-offs, and details of fairness metrics are major challenges towards responsible practices of machine learning for the ordinary data scientists.
\neww{
In particular, since completely mitigating unfairness from the perspective of all fairness measures might be impossible, it is of interest to instead reduce unfairnesses. While existing fair ML systems enable specifying multiple fairness constraints, they fail in practice when the number of fairness constrains is not small.
As a result, in order to balance unfairness among different metrics, it is critical to identify their correlations, and to select a small subset of representative ones to focus on.
}
% \newww{Since there is no single mitigation approach that works for all notions and models, in order to balance unfairness among different metrics for a given context, it is really important to identify the correlation between them. The metrics can be positively or negatively correlated, or they can be orthogonal, hence, mitigating one might help or hurt the other. As a result, the goal of this work is not to mitigate representatives using one universal technique or to mitigate one representative to mitigate all others. The outcome of our proposal provides an insight into the set of representative metrics that, if addressed separately, would maximize the total unfairness from different perspectives (various notions).}
To alleviate the overwhelming task of selecting a subset of fairness measures to consider for a context (a data set and a model type), we proposed a framework that, given a set of fairness notions of interest, estimates the correlations between them and identifies a subset of notions that represent others.
\neww{
Our goal in this paper is not to design new techniques for fair ML but to specify a subset of metrics that represent others. 
The outcome of our proposal provides an insight into the set of representative metrics that, if the model unfairness is balanced between those, the total unfairness from different perspectives (various notions) is balanced.
}

We conducted comprehensive experiments using benchmark data sets and 
% Our comprehensive experiments on benchmark data sets and 
different classification models to evaluate the our approach.
First, our experiments verified the effectiveness of our approach in finding correlations, returning robust and low-variance estimations.
Next, we observed in the experiments that correlation values between different fairness metrics are model-type and data-dependent. Finally, our approach could specify a subset of representative metrics.

While the general approach proposed in this model can be applied to other ML tasks, our focus in this paper was on classification. Extending our findings for other tasks such as regression is an interesting direction for future work.
Another interesting direction for future work is to study the dynamic systems and the impact of data distribution drift on the correlations and the set of representative measures.

% \section{Acknowledgment}
% This study project was funded in part by Institute of Education Sciences 

\section{Application in Higher Education}

The Education Longitudinal Study (ELS:2002) dataset, is a longitudinal study of 10th graders in 2002 and 12th graders in 2004 that was designed to collect data on the students who were tracked throughout their secondary and postsecondary education. In this study, we concentrate on a selected subset of the available attributes and discard observations with a large number of missing attributes.  We consider the variable \emph{highest level of degree} as an indicator of students success and  construct a binary classification problem by classifying students with a college degree (BS degree and higher) as the favorable outcome (label=1) and others as the unfavorable outcome (label=0). The sensitive attribute in this case is considered to be the attribute race.
% \nazanin{\textbf{Education Longitudinal Study (ELS:2002)} is a nationally representative, longitudinal study of 10th graders in 2002 and 12th graders in 2004 which attempted to collect data on the students followed throughout secondary and postsecondary years. We focus on a small subset of available attributes in this study, and  remove the observations that have many
% missing attributes. We consider the variable \emph{highest level of degree} as an indicator of students success and construct a binary classification problem by labeling students with a college degree (BS degree and higher) as the favorable outcome (label=1), and others as the unfavorable outcome (label=0). The attribute race is considered as the sensitive attribute in this problem.} 

\begin{figure*}[!tb]
\centering
\subfloat[ELS-Logit]{\includegraphics[width=0.5\textwidth]{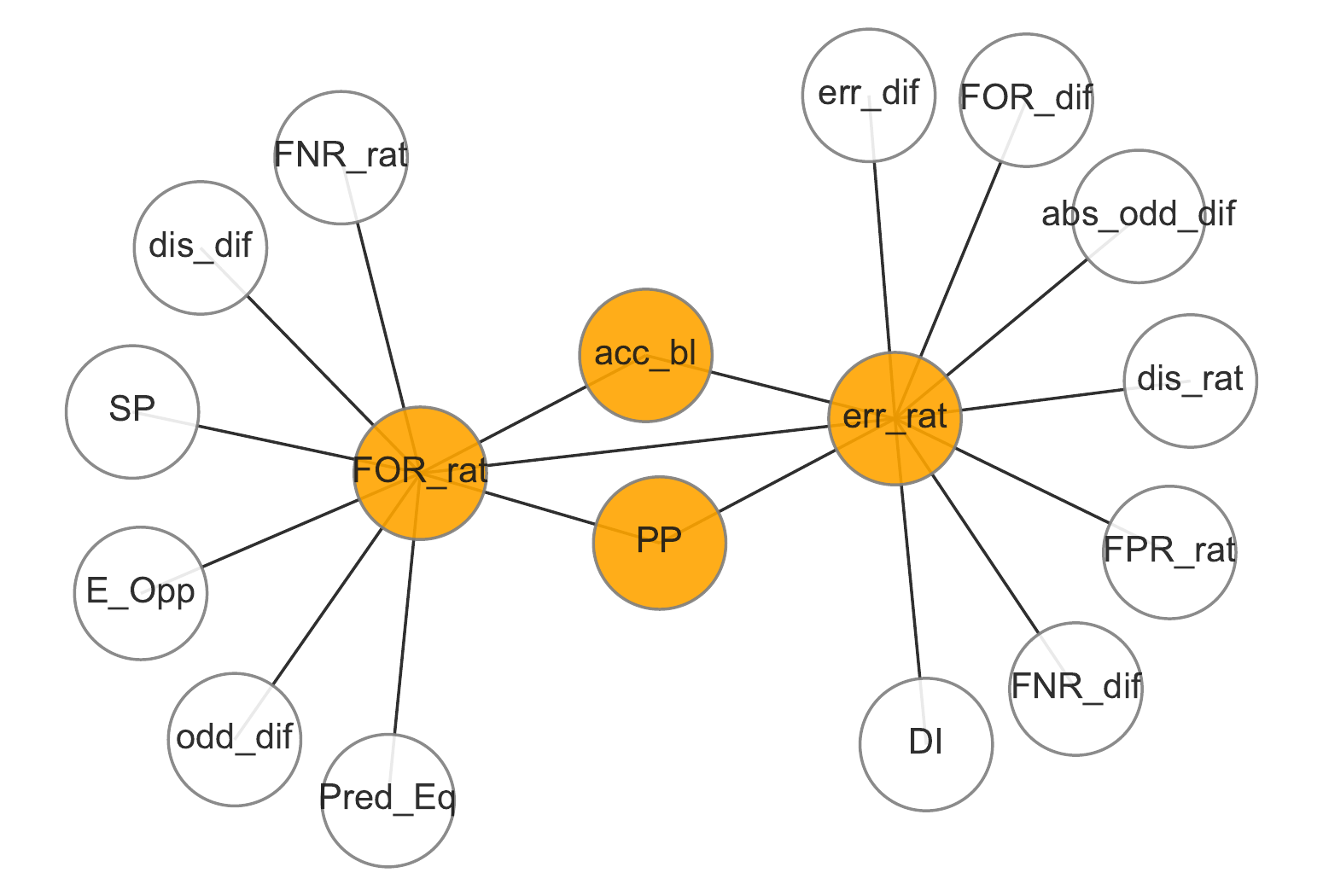}}
\subfloat[ELS-SVM]{\includegraphics[width=0.5\textwidth]{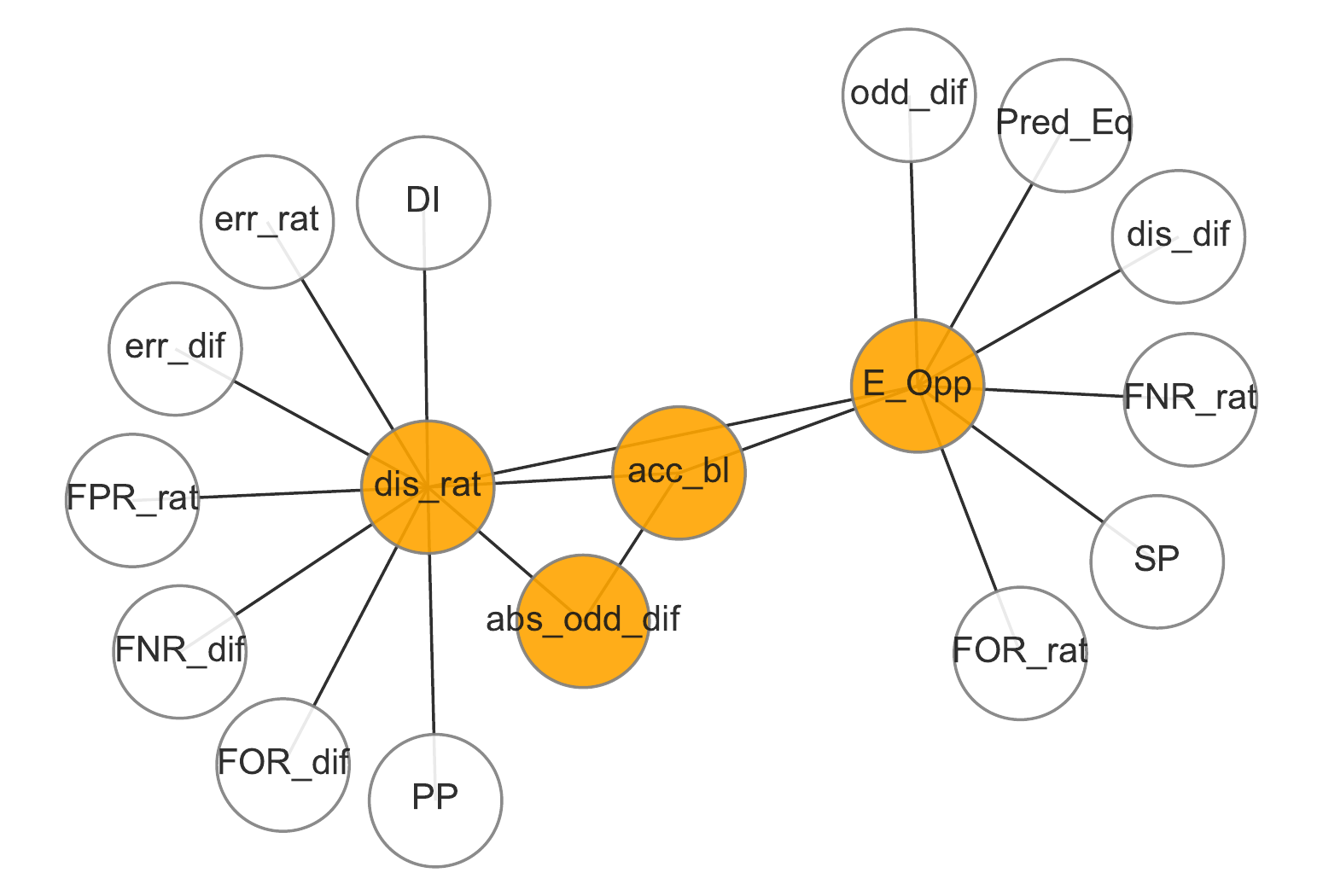}}

\subfloat[ELS-RF]
{\includegraphics[width=0.5\textwidth]{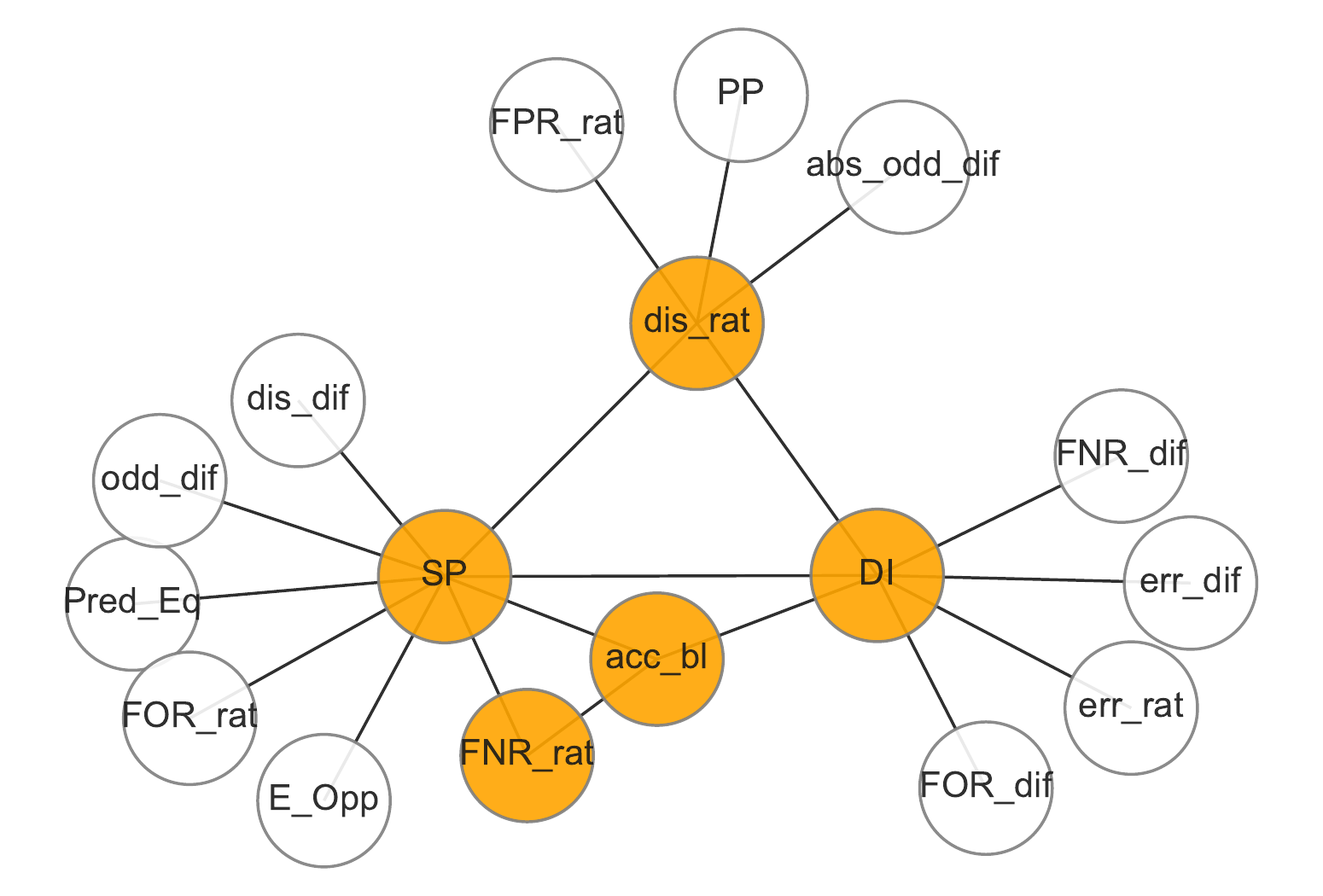}}

\subfloat[ELS-KNN]
{\includegraphics[width=0.5\textwidth]{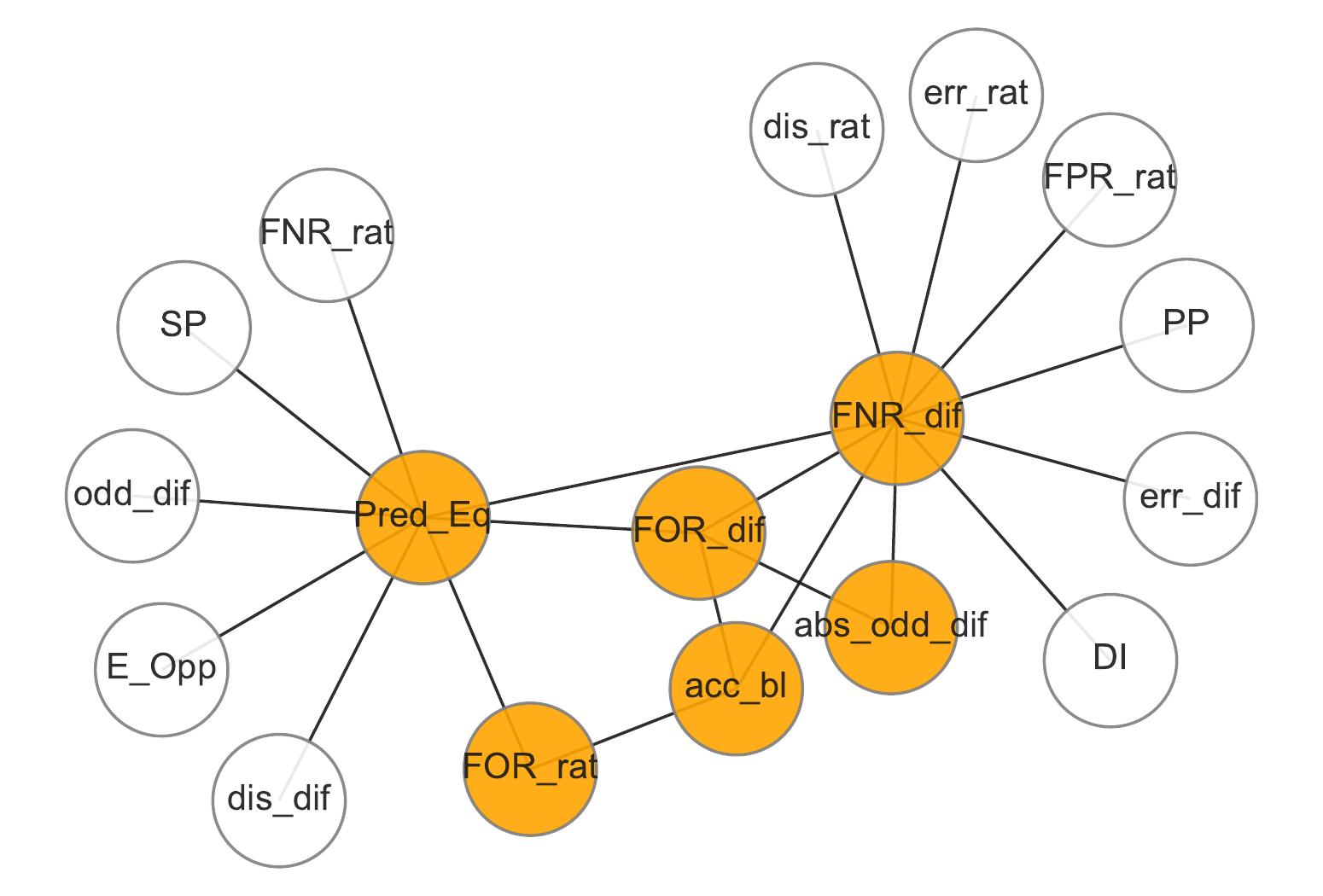}}
\subfloat[ELS-NN]
{\includegraphics[width=0.5\textwidth]{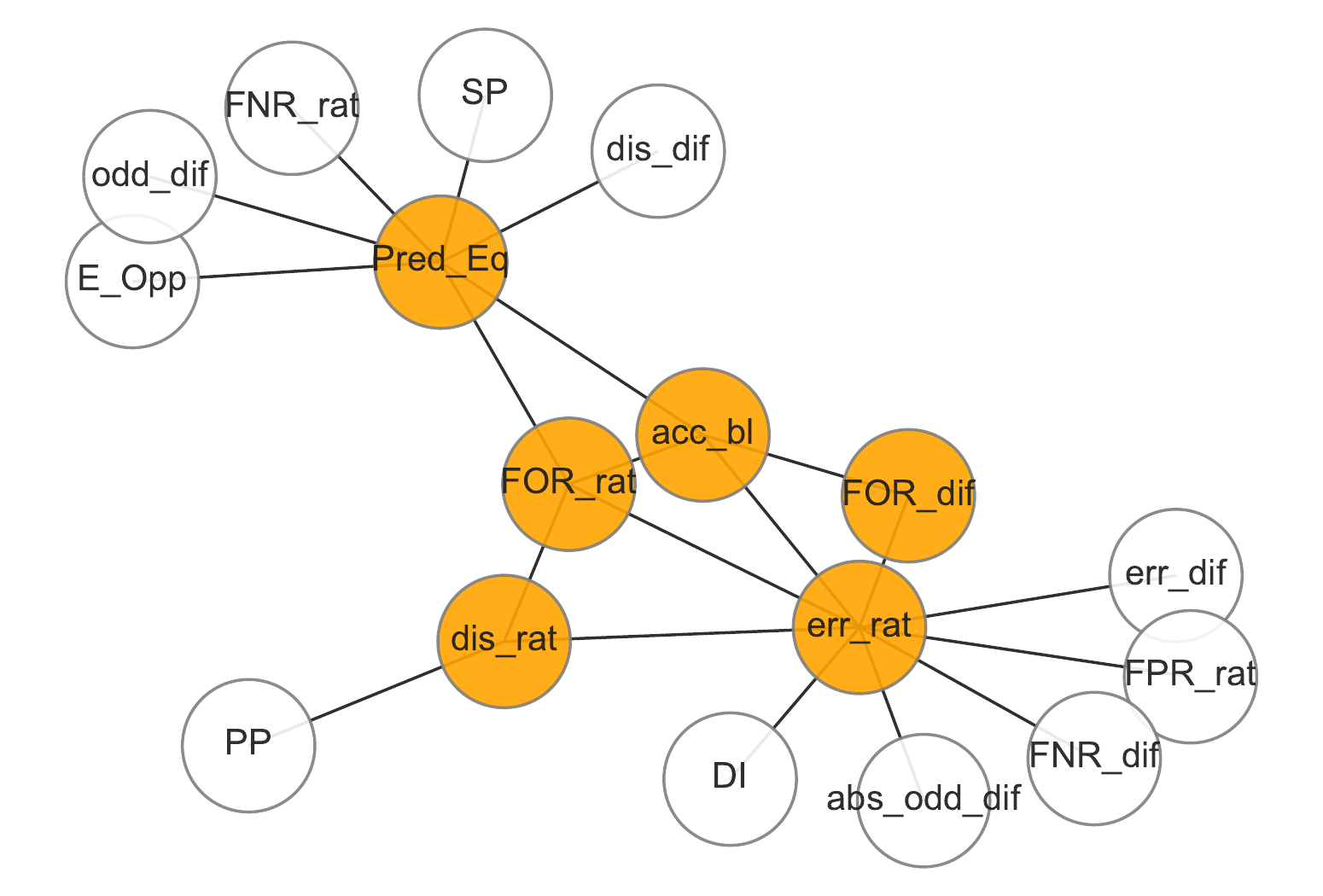}}

\caption{Graph illustration of identified subset of fairness representatives for ELS dataset} 
\label{fig:ELS-graph}
\end{figure*}

\section{Compliance with Ethical Standards}
\stitle{Funding:} Hadis Anahideh and Nazanin Nezami were funded in part by Institute of Education Sciences (R305D220055). Abolfazl Asudeh was supported in part by NSF grant 2107290.
\stitle{Ethical approval:} This article does not contain any studies with human participants or animals performed by any of the authors.

% \section{Limitations}

% \nazanin{The application of our framework would help decision makers while performing bias mitigation tasks. Being aware of the interaction between important notions within a context prevents the introduction of new source of bias while fixing the problem based on a specific fairness metric. However, we note that a single bias mitigation approach does not works for all prediction models and notions in practice. For each model and notion, we might need to consider a different bias mitigation approach. As a result, we first need to identify an approach to fix the unfairness issue for the representative notion of interest, and then utilize the correlation clustering analysis to identify the impact of the corresponding mitigation on the other metrics.}

\bibliography{ref,  ref-abol}

% \input{intro}
% \input{background}
% \input{technical}
% \input{results}
% \input{related_work}
% \input{conclusion}
%\input{Limitation}
%\clearpage
% \newpage
%\bibliographystyle{informs2014}
%\bibliography{sn-bibliography}

\end{document}